\documentclass{article}

\usepackage[nonatbib,final]{neurips_2024}

\usepackage[utf8]{inputenc} %
\usepackage[T1]{fontenc}    %
\usepackage{hyperref}       %
\usepackage{url}            %
\usepackage{booktabs}       %
\usepackage{multirow}
\usepackage{amsfonts}       %
\usepackage{nicefrac}       %
\usepackage{microtype}      %
\usepackage{xcolor}         %
\usepackage{minitoc}

\usepackage{graphicx}
\usepackage{algorithm}
\usepackage{algorithmic}
\usepackage{subcaption}
\usepackage{tikz}
\usepackage{hyperref}
\usepackage{tcolorbox}
\usepackage{textcomp}
\usepackage{color}
\usepackage{enumitem}
\usepackage{wrapfig}
\usepackage{array}
\usepackage{amsthm}
\usepackage{amssymb}
\usepackage{amsmath}
\usepackage{bbold}

\theoremstyle{plain}
\newtheorem{prop}{Proposition}
\newtheorem{thm}{Theorem}
\newtheorem{coro}{Corollary}

\newtheorem{example}{Example}

\newtheorem*{prop*}{Proposition}
\newtheorem*{thm*}{Theorem}
\newtheorem*{coro*}{Corollary}
\newtheorem*{lem*}{Lemma}
\newtheorem*{example*}{Example}

\theoremstyle{definition}
\newtheorem{defn}{Definition}

\theoremstyle{remark}
\newtheorem{rem}{Remark}

\newcommand{\Frond}{\mathcal F}

\newcommand{\Irond}{\mathcal I}

\newcommand{\vertiii}[1]{{\left\vert\kern-0.25ex\left\vert\kern-0.25ex\left\vert #1 
    \right\vert\kern-0.25ex\right\vert\kern-0.25ex\right\vert}}

\renewcommand{\geq}{\geqslant}
\renewcommand{\leq}{\leqslant}

\usepackage{placeins}
\usepackage{bbm}
\usepackage{bm}

\floatname{algorithm}{Procedure}
\renewcommand{\algorithmicrequire}{\textbf{Input:}}
\renewcommand{\algorithmicensure}{\textbf{Output:}}

\title{When is an Embedder More Promising \\ than Another? 
}
\def\sys{$\overline{\mathcal{I}_S}$}
\def\is{IS}

\author{ 
    {${}^{\dagger}$\bf Maxime \textsc{Darrin}$^{1, 2, 3, 4}$} \quad {${}^{\dagger}$\bf Philippe \textsc{Formont}$^{1, 2, 4, 5}$ } \quad \bf Ismail \textsc{Ben Ayed}$^{1, 5}$ \\     {\bf Jackie Chi Kit \textsc{Cheung}$^{2, 3}$}   \quad {\bf Pablo \textsc{Piantanida}$^{1, 2, 4, 6}$} \\ {$^{1}$International Laboratory on Learning Systems, $^{2}$Mila - Quebec AI Institute, $^{3}$McGill University} \\ {  $^{4}$Université Paris-Saclay, $^{5}$ÉTS Montréal, $^{6}$CNRS, CentraleSupélec}, ${}^{\dagger}$equal contribution \\ { \url{maxime.darrin@mila.quebec}, \url{philippe.formont@mila.quebec}}
}

\begin{document}
\doparttoc %
\faketableofcontents %

\maketitle

\begin{abstract}
Embedders play a central role in machine learning, projecting any object into numerical representations that can, in turn, be leveraged to perform various downstream tasks. The evaluation of embedding models typically depends on domain-specific empirical approaches utilizing downstream tasks, primarily because of the lack of a standardized framework for comparison. However, acquiring adequately large and representative datasets for conducting these assessments is not always viable and can prove to be prohibitively expensive and time-consuming.  In this paper, we present a unified approach to evaluate embedders. First, we establish theoretical foundations for comparing embedding models, drawing upon the concepts of sufficiency and informativeness. We then leverage these concepts to devise a tractable comparison criterion (information sufficiency), leading to a task-agnostic and self-supervised ranking procedure. We demonstrate experimentally that our approach aligns closely with the capability of embedding models to facilitate various downstream tasks in both natural language processing and molecular biology. This effectively offers practitioners a valuable tool for prioritizing model trials.\footnote{The code used to perform all experiments is available at \url{https://github.com/ills-montreal/emir}}
\end{abstract}

\section{Introduction}

Embeddings are a prominent tool in machine learning and are used in multiple fields, such as natural language processing~\cite{li2023angleoptimized, pimentel2023usefulness}, computer vision~\cite{schuhmann2021laion, kubota2024impressionclip, bhalla2024interpreting, khandelwal2022simple} or bioinformatics~\cite{liu2022pretraining, ahmad2022chemberta2, chithrananda2020chemberta, wang2022chemicalreactionaware}.
These models embed objects such as images, texts, or molecules into numerical representations that can be used to perform numerous downstream tasks by preserving key features of the object~\cite{murphy2013machine, DBLP:journals/corr/VilnisM14}.

Depending on the data modalities, intended purpose, and available resources, embedders showcase a wide variety of architectures, training settings (unsupervised, supervised, self-supervised, etc.), objectives (masked language modeling, contrastive learning, etc.) \cite{chang2020pretraining, 7296633, pan2022extreme, wang2022chemicalreactionaware, pmlr-v202-feng23c}, and datasets \cite{li2023towards, reimers-2020-multilingual-sentence-bert, devlin2019bert, feng2022languageagnostic, assran2023selfsupervised, yang2022learning}. And more recently, foundation models have become a natural starting point to create embedders~\cite{che2024enhancing, touvron2023llama, jiang2023mistral, SFRAIResearch2024}.

This diversity and variety of options makes selecting the most promising embedders for a data distribution challenging~\cite{muennighoff2023mteb}. 
Most work evaluates embedders focusing on the performance they enable on a finite set of downstream tasks~\cite{reimers2019sentencebert, borgeaud2022improving, saharia2022photorealistic, santos-etal-2020-word, perone2018evaluation, choi2021evaluation}.
Nevertheless, this evaluation process encounters two primary limitations. Firstly, it is \textbf{not scalable} concerning the number of embedders and tasks, as it requires fitting a downstream model for each task. Hence, prioritizing the evaluation of the most promising models becomes essential to mitigate computational costs. Secondly, \textbf{acquiring high-quality labels} can be a \textbf{time-consuming} and notably \textbf{expensive} endeavor in various applications. To overcome these limitations, in this paper, we explore task-agnostic evaluation metrics for embedders relying solely on pairwise comparisons between embedders, i.e., without the need for labeled data in downstream tasks.

More specifically, our contributions can be summarized as follows:
\begin{enumerate}[noitemsep,topsep=0pt,parsep=0pt,partopsep=0pt]
    \item \textbf{An innovative theoretical framework for comparing embedding models:} We cast the problem of ranking embedders into the noisy communication channels ordering (\autoref{sec:sufficiency_and_info_main}) and statistical experiments comparison settings (\autoref{sec:comparing_stat_expes_main}). We exploit the notions of sufficiency and informativeness and relax them, leveraging the concept of deficiency introduced by Le Cam~\cite{le1964sufficiency} (\autoref{sec:lecam_deficiency_main}), which is reframed to account for concepts and features. 
    These concepts provide us with tools to establish an embedder ranking.
   
    \item \textbf{A practical relaxation:} Estimating deficiency presents significant challenges. We propose the concept of information sufficiency ({\is}), which quantifies the information required to simulate one embedder from another (\autoref{sec:is_estimation}). We estimate the information efficiency to get a task-agnostic and label-free comparison tool for embedders evaluation.
    
    \item \textbf{Extensive experimental validation:} The expected IS correlates with the ability of embedders to enable a wide range of downstream tasks. In NLP (\autoref{Experimental-NLP}) and molecular modeling (\autoref{sec:drugdiscovery}), our method respectively achieves Spearman ranking correlations of $0.90$ ($56$ tasks) and $0.94$ ($31$ tasks); providing an efficient model trial prioritization tool for practitioners.
\end{enumerate}

\subsection{Related works}

\paragraph{Embedding evaluation.} Embedding evaluation is mainly performed based on a limited set of downstream tasks~\cite{chen2013expressive, santos-etal-2020-word, perone2018evaluation, choi2021evaluation}, for which the embeddings are used as inputs to smaller models. Therefore, embedders evaluation is field- and task-specific. In NLP, \cite{gao2022simcse, reimers2019sentencebert} they rely on a limited set of tasks; more recently, the Massive Text Embedding Benchmark (MTEB)~\cite{muennighoff2023mteb} followed this task-oriented trend and offered standardized test bed for embedders encompassing various downstream tasks in NP. Devising statistical tests to compare models and learning algorithms has a long history~\cite{demvsar2006statistical}. However, most works propose statistical tests relying on the performance of the downstream tasks of interests~\cite{lacoste2012bayesian, benavoli2017time}. Other works study the expressiveness of embedders and connect it to performance on downstream tasks \cite{tsitsulin2023unsupervised, chuang2019role}, but mostly focus on geometrical properties of the high dimensional representation in self-supervised learning settings~\cite{NEURIPS2022_70596d70, garrido2023rankme,  pmlr-v162-he22c}.

\noindent\textbf{Probing.} While probing methods do not aim at comparing embedders, they evaluate their representations to discover what these models have learned. They train small models on the internal representations of large models to perform specific downstream tasks. These procedures allow researchers to assess what information is present and recoverable from these embeddings~\cite {belinkov-mit-22, adi-iclr-17, rogers-tacl-21, pimentel-acl-20}. Other work proposed measuring mutual information (MI) between internal representations and labels. It has been used to evaluate the difficulty of a dataset as the predictiveness of the labels using the features ~\cite{ethayarajh2022understanding}. For instance, \cite{sui2023evaluating} evaluates the utility of representations in astrophysics to predict physical properties. Following this trend,~\cite{kim2022mutual} leverages the point-wise MI between Gaussian distributions to evaluate text-to-images and image-to-text generative models. However, none of these methods have focused on comparing embedders in the general case to the best of our knowledge.

\section{Theoretical Foundations for Comparing Embedding Models}
\label{sec:theory}

\subsection{Background and notation}
\label{sec:background_main}

We assume that all considered spaces are standard Borel~\cite{crauel2002random}  Each such space $\mathsf{U}$ is equipped with its Borel $\sigma$-algebra $\mathcal{B}(\mathsf{U})$. The set of all probability measures on $\mathsf{U}$ is denoted by $\mathcal{P}(\mathsf{U})$  The total variation distance between $P$ and $Q$ is denoted by $\|P - Q\|_{\textrm{TV}} $. Given a joint probability measure $P_{XY}$  induced by two random variables $X\in \mathsf{X}$ and $U\in\mathsf{U}$, the Mutual Information~\cite{Cover91} is denoted by  $I(X;U)$. A Markov (or transition probability) kernel between $\mathsf{X}$ and $\mathsf{U}$ is a mapping $P_{U|X} : \mathcal{B}(\mathsf{U}) \times \mathsf{X} \to [0,1]$. The space of all such $P_{U|X}$ is denoted by $\mathcal{K}(\mathsf{U}|\mathsf{X})$ and  $(M\!\circ\! P_{U|X})(V|x)$ indicates the composition of Markov kernels $M\in \mathcal{K}(\mathsf{V}|\mathsf{U})$ and $P_{U|X} \in \mathcal{K}(\mathsf{U}|\mathsf{X})$.  For further details, refer to \autoref{appendix_def}.

\subsection{Sufficiency and informativeness ordering of embedding models}
\label{sec:sufficiency_and_info_main}

\begin{wrapfigure}{l}{0.56\linewidth}
    \centering
    \includegraphics[width=\linewidth]{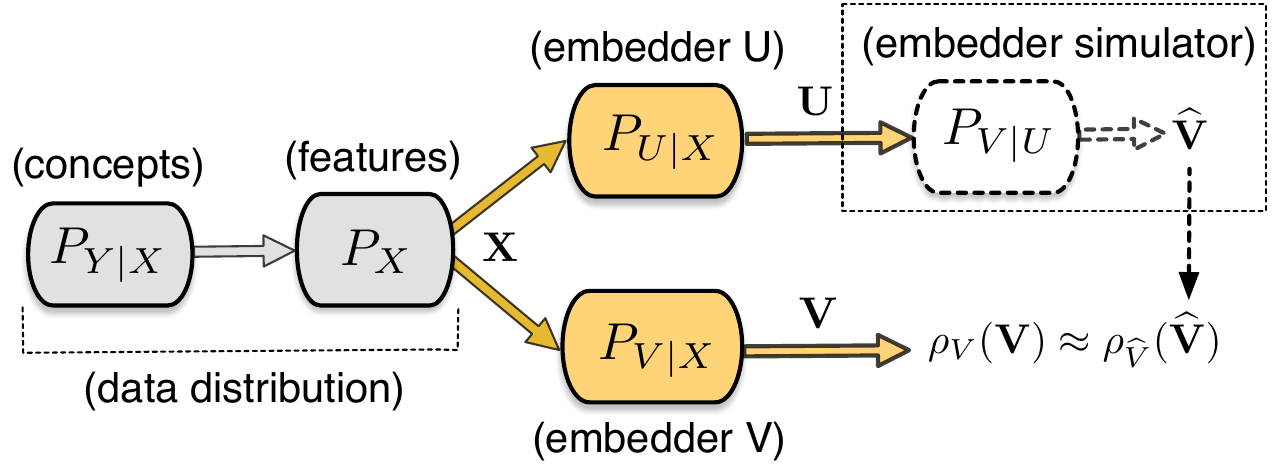}
    \caption{Communicating a concept $y\in \mathsf{Y} $ over two embedding models with prediction $\rho_V(V)$.}
    \label{fig:fig_channels}
\end{wrapfigure}

We aim to compare embedding models without relying on labeled data for downstream tasks. Let us consider two embedding models represented by their Markov kernels (or transition probabilities) $P_{U|X} \in \mathcal{K}(\mathsf{U}|\mathsf{X})$ and $P_{V|X} \in \mathcal{K}(\mathsf{V}|\mathsf{X})$, any target set $\mathsf{Y}$ of (discrete or continuous) concepts and feature space $\mathsf{X}$ with joint probability measure $P_{YX} \in \mathcal{P}(\mathsf{Y}\times \mathsf{X})$ induced by random variables $(Y,X)\in \mathsf{Y}\times \mathsf{X}$, as illustrated in \autoref{fig:fig_channels}. First, we study the question: \begin{quote} \textbf{What sufficient conditions must be met by the embedding model $U$ relative to $V$ to guarantee that $I(Y;U)\geq I(Y;V)$ for all  distributions $P_{YX}$ ?}
\end{quote}

From an information-theoretic perspective~\cite{Cover91}, the quality of an embedding model can be likened to the capacity of a noisy communication channel with an uncoded input (e.g., a text, a molecule...), where a downstream task of interest is performed at the output (the embedding) of the channel. Let $Y\in \mathsf{Y}$ represent the message (the source) to be communicated over both channels; $X$ represents the transmitted signal; and $P_{U|X}$ and $P_{V|X}$ the communication channels with outputs $U$ and $V$, respectively. This process is illustrated in \autoref{fig:fig_channels}. It naturally satisfies the Markov chain $Y \leftrightarrow X \leftrightarrow (U,V) $. A desirable property is that the embedding models $U$ and $V$ retain as much pertinent information as feasible to predict $Y$.

We shall be interested in the underlying information relationships between those embedding models that can be interpreted as channel $U$ being "more informative" for communicating $Y$ than channel $V$. The first attempt to introduce an ordering between communication channels appears in Shannon~\cite{SHANNON1958390}. Körner and Marton later introduced~\cite{Korner1977} the concepts of "less noisy" (or more informative) and "degraded" (or sufficiency) orderings between channels.

\begin{defn}[Sufficiency and informativeness orderings~\cite{Korner1977}]
    \label{def-moreinfo}
    Let $P_{U|X}$ and $P_{V|X}$ be two Markov kernels (embedding models).
    \begin{itemize}[topsep=0pt,parsep=0pt,partopsep=0pt]
        \item \textbf{Sufficiency $U \succcurlyeq_S V$.} The embedding model $P_{U|X}$ is said to be "sufficient" for the embedding model $P_{U|X}$ (or $V$ to be degraded w.r.t. $U$) if and only if there exists another Markov kernel $M \in \mathcal{T}(\mathsf{V}|\mathsf{U})$  such that $\mathbb{E} \| M\!\circ\!P_{U|X} - P_{V|X} \|_{\text{TV}}=0$, i.e. $V$ can be simulated from $U$ using $M$ without information loss).
        \item \textbf{More informative $U \succcurlyeq_I V$.} The embedding model $P_{U|X}$ is said to be "more informative" (or less noisy) than $P_{V|X}$ if and only if the  embedding models always satisfy the inequality
        \begin{equation*}
              I(Y ; U) \geq I(Y;V), \quad \forall P_{YX} \in \mathcal{P}(\mathsf{Y}\times \mathsf{X}).
        \end{equation*}
    \end{itemize}
\end{defn}

\vspace{1mm}
\begin{prop}[Relationships of sufficiency and information]\label{Relationships} The following relationships hold:
\begin{itemize}[topsep=0pt,parsep=0pt,partopsep=0pt]
    \item[(i)] \textbf{Sufficiency $\Rightarrow$ informativeness.} If the embedding model $P_{U|X}$ is  sufficient for the embedding model $P_{V|X}$, i.e. $U \succcurlyeq_S V$, then $U \succcurlyeq_I V$. However, \textbf{Informativeness $\nRightarrow$ sufficiency.} %
    \item[(ii)]  \textbf{Informativeness $\Rightarrow$ higher capacity  to distinguish  concepts.} If the embedding model $P_{U|X}$ is more informative  than embedding model $P_{V|X}$, i.e. $U  \succcurlyeq_I V$, then
    \[
        \textrm{KL}\big( P_{U|Y}(\cdot|y_0)\| P_{U|Y} (\cdot|y_1) \big) \geq     \textrm{KL}\big( P_{V|Y} (\cdot|y_0)\| P_{V|Y} (\cdot|y_1) \big),
    \]
    for any pair of concepts $(y_0,y_1)\in \mathsf{Y}\times \mathsf{Y}$ and all probability distributions  $ P_{YX}$.
\end{itemize}
\label{Prop1}
\end{prop}
\vspace{1mm}
\begin{rem}
    An immediate consequence of claim (i) is that the sufficient condition between embedding models implies that the embedding model $U$ is more informative than the embedding model $V$ relative to all target concepts in $\mathsf{Y}$ over all possible data distributions: $ I(Y ; U) \geq I(Y;V)$, for all probability distributions $ P_{YX}$. 
    
    Although  $U$ being more informative than $V$ does not necessarily imply $U \succcurlyeq_S V$~\cite{Korner1977, lindley1956measure}; (ii) states that being more informative ensures a higher statistical discrimination capacity between any pairs of target concepts (for further discussion, see \autoref{sec_distinguish}).
\end{rem}

Motivated by the concepts of sufficiency and informativeness between embedding models, we can inquire about their statistical consequences for a learner conducting an inference task on these embeddings. More precisely, given a finite set of concepts $\mathsf{Y}$, \textbf{if $U \succcurlyeq_S  V$, is the Bayes risk expected to be smaller when the inference is based on $U$ than when it is based on $V$? }

\subsection{Comparing statistical experiments with embedding Models}
\label{sec:comparing_stat_expes_main}

The pursuit of comparing statistical experiments originated from the seminal paper by Bohnenblust, Shapley, and Sherman~\cite{Bohnenblust1949ReconnaissanceIG}, followed by subsequent contributions by Blackwell~\cite{blackwell1951comparison,af8b16cd-acdb-3d4e-8fd0-40e54a9cc18a}. They formally established the relationships between sufficiency (\autoref{def-moreinfo}) and inference procedures.

In our framework, a statistical experiment~\cite{blackwell1951comparison} consists of a mathematical abstraction (see~\autoref{appendix_def} for further details) intended to represent a downstream task where a learner aims at inferring a concept $y\in \mathsf{Y}$ from the embeddings $U$ or $V$. Deciding what embedder should be used to perform a given task is too general. In this work, we do not take into account the computational cost or the size of an embedder and solely focus on the following question:

\begin{quote}
\textbf{What are the necessary and sufficient conditions that ensure that employing the embedding $U$ for any task $P_{YX}$ leads to lower risk compared to using the embedding $V$?}
\end{quote}

Drawing parallels with the theoretical framework established for comparing statistical experiments, a relationship can be derived between the concept of sufficiency and the expected risk for a specific task (see~\autoref{appendix_propo_one-task} for further discussion).

We concentrate on the scenario where $\mathsf{Y}$ consists of a finite number of concepts (e.g., classification tasks), as it is a significant case in its own right~\cite{torgersen1970comparison} and provide fundamental insights for the present work. The next Proposition states an important \textbf{relation between the concept of sufficiency and the expected Bayes risk on any classification task. }
\begin{prop}[Comparison of embedding models through Bayes risks]\label{prop-bayes}
Given two embedding models $P_{U|X}\in \mathcal{K}(\mathsf{U}|\mathsf{X})$ and $P_{V|X}\in \mathcal{K}(\mathsf{V}|\mathsf{X})$,  the following statements are equivalent:
\begin{enumerate}
    \item[(i)] The embedding model $P_{U|X}$ is sufficient relative to $P_{V|X}$, i.e. $U \succcurlyeq_S V$.
    \item[(ii)] For all conditional probability measures $P_{Y|X}$ on finite alphabet $\mathsf{Y}$, the  Bayes risks  satisfy
    \[
        \inf_{\rho_U:\mathsf{U}\rightarrow \mathcal{P}(\mathsf{Y})}  \Pr\big(\hat{Y}_U\neq Y \big) \leq   \inf_{\rho_V:\mathsf{V}\rightarrow \mathcal{P}(\mathsf{Y})} \Pr\big(\hat{Y}_V \neq Y\big),
    \]
    where $\hat{Y}_U$ and $\hat{Y}_V$ are distributed according to $\rho_U(U)$ and $\rho_V(V)$, respectively.
\end{enumerate}
\end{prop}

\begin{rem}
    In other words, if we can fully simulate an embedder $V$ from another embedder $U$, the expected risk across all potential classification tasks cannot be greater when using $U$ compared to $V$. The proof of this Proposition is given in \autoref{Appex-prop-bayes}. It is worth mentioning that various versions of this result are available in the literature~\cite{torgersen1970comparison}. However, our extension here, in a simpler setting, incorporates concepts and features into the experiment comparison framework.
\end{rem}

\subsection{Challenges in ranking embedding models and their deficiency}
\label{sec:lecam_deficiency_main}

According to the notion of "sufficiency", we can distinguish the three following possibilities:

\begin{itemize}[noitemsep,topsep=0pt,parsep=0pt,partopsep=0pt]
    \item Equivalence: $U \succcurlyeq_S V$ and $V  \succcurlyeq_S U$ denoted $U \approx V$; $U$ and $V$ can simulate each other.
    \item Comparability: $U \succcurlyeq_S V$ but $V  \not\succcurlyeq_S U$ only $V$ can be simulated from $U$.
    \item Non-comparability: $U \not\succcurlyeq_S V$ and $V  \not\succcurlyeq_S U$, neither $U$ nor $V$ can simulate each other.
\end{itemize}
Our results up to now only account for the two first possibilities. However, two embedders are generally not comparable (\autoref{examples}). This issue was addressed by Le Cam~\cite{le1964sufficiency}, who introduced the notion of "deficiency".

\begin{defn}
    \label{def:deficiency}
    The deficiency $ \delta(P_{U|X} \rightarrow P_{V|X} ) $ of $P_{V|X}$ relative to $P_{U|X}$ is defined as~\cite{le1964sufficiency}
    \[
        \delta(P_{U|X} \rightarrow P_{V|X} ) \triangleq  \inf_{M\in\mathcal{K}(\mathsf{V}|\mathsf{U})}  \mathbb{E} \| M\!\circ\!P_{U|X} - P_{V|X} \|_{\text{TV}},
    \]
    where the infimum is taken over all Markov kernels (or transition probabilities) $M \in \mathcal{K}(\mathsf{V}|\mathsf{U}) $, mapping stochastically $\mathsf{U}$ and $\mathsf{V}$, and $ \delta$ measures error between the simulated and true embedders.
\end{defn}

$\delta$ \textbf{indicates how well one model can be reconstructed from the other}, it induces a natural relaxation of the sufficiency where the reconstruction does not have to be perfect\footnote{If $U \succcurlyeq_S, V$, then $\delta(P_{V|X} \rightarrow P_{U|X} ) = 0$, while if $U \not\succcurlyeq_S V$, then $\delta(P_{V|X} \rightarrow P_{U|X} ) > 0$.} for us to obtain guarantees on the downstream tasks performance (See \autoref{prop-bayes-ext}). It avoids the non-comparability problem by evaluating \textbf{"how much information" we lose when passing from one model to the other one}.

Le Cam~\cite{le1964sufficiency} showed that, for a given task $Y$, the deficiency $\delta(P_{U|Y} \rightarrow P_{V|Y} )$ is directly related to the expected Bayes risks on the task (see \autoref{subsec:le-cam}).
We extend this result to the comparison of two embedding models $P_{U|X}$ and $P_{V|X}$ in a task-agnostic manner and build the relation to the expected Bayes risks for any classification task $Y$.  
\begin{coro}\label{prop-bayes-ext}
Given two embedding models $P_{U|X}$ and $P_{V|X}$ satisfying: 
\begin{enumerate}[noitemsep,topsep=0pt,parsep=0pt,partopsep=0pt]
\item[(i)] The deficiency $ \delta(P_{U|X} \rightarrow P_{V|X} ) \leq \gamma$.  
\item[(ii)] For any conditional  distribution $P_{Y|X}$ on finite alphabets $\mathsf{Y}$,  
\[
\inf_{\rho_U:\mathsf{U}\rightarrow \mathcal{P}(\mathsf{Y})}  \Pr\big(\hat{Y}_U\neq Y \big)  -  \varepsilon   \leq \inf_{\rho_V:\mathsf{V}\rightarrow \mathcal{P}(\mathsf{Y})}  \Pr\big(\hat{Y}_V\neq Y \big) .
\]
\end{enumerate}
Statement (ii) implies (i) provided that $\gamma \geq 2| \mathsf{Y}|\varepsilon  $ and conversely, (i) implies (ii) provided that $ \gamma\leq \varepsilon$. 
\end{coro}
The proof of this Corollary is relegated to~\autoref{Appex-prop-bayes}.
\begin{rem}
    In particular, we can infer that for any classification task $Y$, the expected Bayes risk of the embedding model $U$, denoted by   $\mathcal{R}_U$,  is upper bounded by the expected Bayes risk of the embedding model $V$, denoted by   $\mathcal{R}_V$:
    \begin{equation*}
        \mathcal{R}_U  -  \mathcal{R}_V  \leq  \delta(P_{U|X} \rightarrow P_{V|X} ),  \quad \textrm{for all conditional  distributions $P_{Y|X}$, }  %
    \end{equation*}
and similarly, $    |    \mathcal{R}_U  -  \mathcal{R}_V  | \leq   \max\big\{ \delta(P_{U|X} \rightarrow P_{V|X} ), \delta(P_{V|X} \rightarrow P_{U|X} ) \big\} $, for all conditional distributions $P_{Y|X}$. If both deficiencies are small, the resulting expected Bayes risks of the embedding models $U$ and $V$ will be close to each other for any target task $Y$.
\end{rem}

\section{Quantifying Information Sufficiency Between Embedding Models}
\label{sec:is_estimation}

We want to compare embedding models using the concept of deficiency, leveraging \autoref{prop-bayes} and \autoref{prop-bayes-ext}.
These propositions suggest that the performance on any classification task of an embedding model $U$ relative to the model $V$ is bounded by $\delta(P_{U|X} \rightarrow P_{V|X})$. However, estimating the deficiency from data samples is notably challenging~\cite{shiri2000statistical}, and while upper bounds derivation exists, they do not necessarily make it tractable.

\subsection{Estimating Information Sufficiency}

The deficiency $\delta(P_{U|X} \rightarrow P_{V|X})$ between two embedding models $P_{U|X}$ and $P_{V|X}$, measures how well $U$ can be used to simulate $V$ using a Markov kernel $M \in \mathcal{K}(\mathsf{V}|\mathsf{U})$. This section aims to build a tractable proxy for this reconstruction cost. To this end, we estimate how much we can reduce the uncertainty about $Z$ by observing $U$ by learning an appropriate Markov kernel. This corresponds to the information sufficiency~\cite{degroot1962uncertainty, arimoto1971information} and can be interpreted as the information-theoretic counterpart of the deficiency.
The information deficiency between $U$ and $V$ is then defined as:

\begin{defn}[Information sufficiency]
    The information sufficiency $ \mathcal{I}_{S}(U\rightarrow V) $, relative to parametric classes of distributions   $\Frond_\Theta(\mathsf{V})$ and  $\mathcal{K}_\Theta(\mathsf{V} |\mathsf{U} )$ (multivariate Gaussian mixtures~\cite{pichler2022differential}) is defined:
\begin{equation}
   \label{eq:info_suf}
     \mathcal{I}_{S}(U \rightarrow V )  \triangleq
   \underbrace{\inf_{f\in \Frond_\Theta(\mathsf{V})} \mathbb{E}\left[ - \log f(V) \right]}_{\text{Uncertainty of } V} - \underbrace{\mathbb{E}\left[ \inf_{M \in \mathcal{K}_\Theta(\mathsf{V} |\mathsf{U} ) } \mathbb{E}\left[ - \log M(V|U) | U \right]  \right ]}_{\text{Uncertainty when simulating } V \text{ from } U \text{ with } M}.
\end{equation}
\end{defn}

\begin{wrapfigure}[9]{L}{0.215\textwidth}
     \vspace{-0.3cm}
    \centering %
    \includegraphics[width=\linewidth]{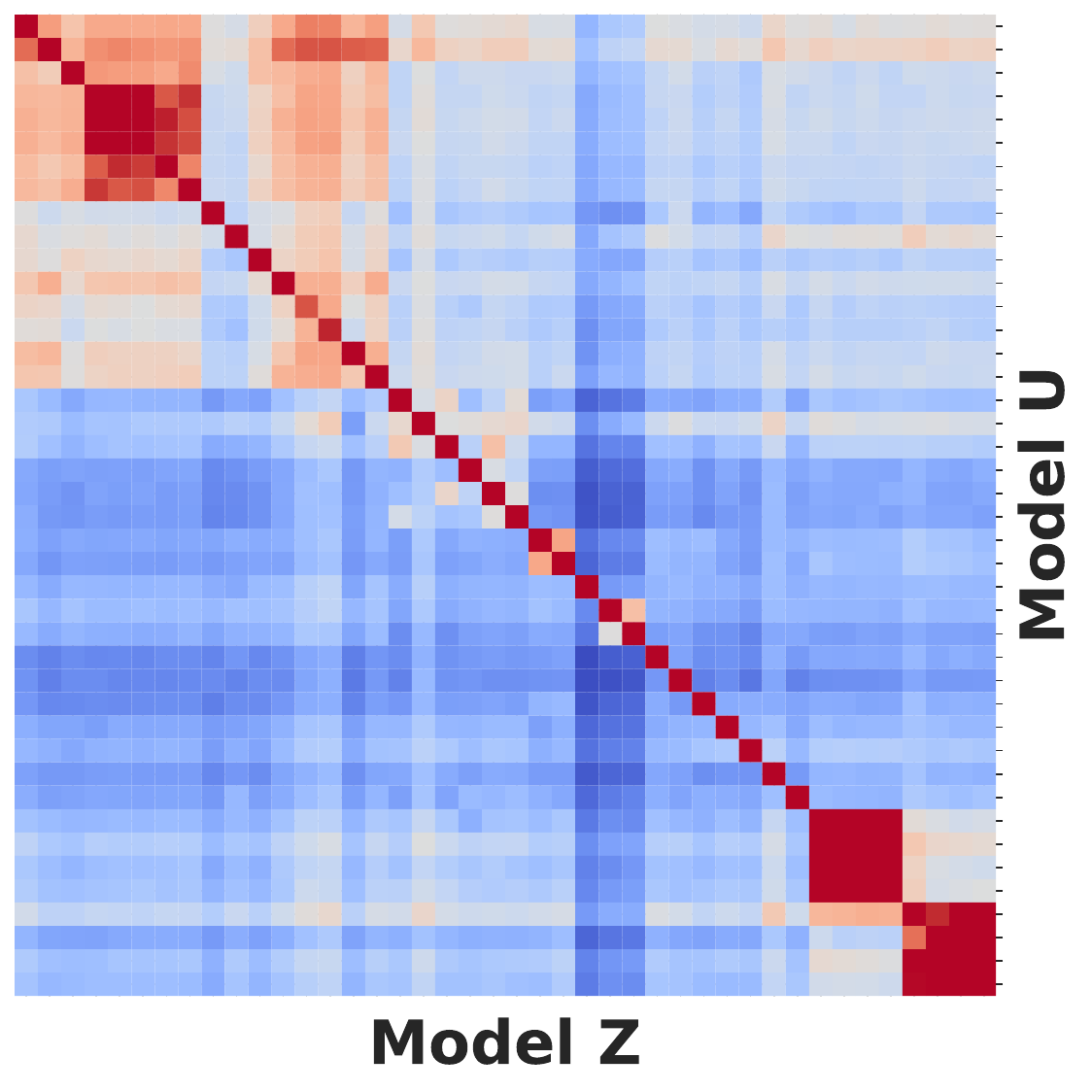}
    \caption{Pairwise \(\Irond_{S}\) for text embedders.}
    \label{fig:nlp_heatmap_informativeness}
\end{wrapfigure}

\begin{rem} When the information sufficiency $  \mathcal{I}_{S}(U \rightarrow V )$ is large, it signifies that $U$ offers a substantial amount of information to simulate $V$, a proxy for a small deficiency. Conversely, when $  \mathcal{I}_{S}(U \rightarrow V )$ is lower, it implies that the channel $P_{V|Y}$ is subject to considerable noise or randomness, leading to a greater loss of statistical information.
\end{rem}

We hence attempt to simulate $V$ from $U$ by learning a  Markov kernel $M \in \mathcal{K}_\Theta(\mathsf{V} |\mathsf{U} )$, via a mixture of multivariate Gaussians, and measure the uncertainty reduction it induces.

\paragraph{Pairwise embedder evaluation.}
For set of embbeders $(Z_k)_k$ represented by their Markov kernels $\{P_{Z_k|X}\}_k$, we compute the pairwise information sufficiency $\mathcal{I}_{S}(Z_k \rightarrow Z_l)$.
The pairwise information sufficiency matrix defines the adjacency matrix of a directed graph of embedders (\autoref{fig:nlp_heatmap_informativeness}).
\autoref{prop-bayes-ext} shows that embedders sharing high information sufficiency are expected to perform similarly on any downstream tasks, motivating the identification of communities in the graph. While the graph construction is in $\mathcal{O}(N^2)$; where $N$ is the number of embedders, it is in practice tractable for a reasonable number of embedders (refer to  \autoref{sec:computational_ressources}) for more details).

\paragraph{Practical embedding evaluation.} We construct the set of all information sufficiency using $Z_k$: $\mathcal{S}_{\mathcal{I_S}}\left(k\right) = \left\{\mathcal{I}_{S}(Z_k \rightarrow Z_l) \right\}_{l\not= k}$. We build our information sufficiency score ({\sys} score) by taking the median of $\mathcal{S}_{\mathcal{I_S}}\left(k\right)$.
Details on the {\sys} score's estimation can be found in~\autoref{sec:is-actual-estim}.

\section{Experimental Setup}
\label{sec:common_setting}

We aim to evaluate the practical utility of the {\sys} score to rank and select the best embedders for a given data distribution. We compare this ranking to those obtained on various downstream tasks. Our experimental protocol is divided into three main steps:
\begin{enumerate}[noitemsep,topsep=0pt,parsep=0pt,partopsep=0pt]
    \item We evaluate the {\sys} score of the models by identifying a large and diverse dataset that is supposed to be representative of the data distribution of interest.
    \item We train a small feedforward neural network ($\rho_{Z_k}$) per embedder $P_{Z_k|X}$ to perform each downstream task and record its performances ($R^2$ score for regression, AUROC/accuracy for binary/multiclass classification).
    \item We compare the models' performances on the downstream tasks and the {\sys} score by measuring three types of correlations: the Pearson correlation, the Spearman correlation, and the Kendall-Tau coefficient.\footnote{For the experiments in molecular modeling, in each subset regression and classification tasks are mixed. Hence, we do not compute the Pearson correlation to avoid mixing scores obtained for different metrics.}(See ~\autoref{sec:naive_baselines} for additional baselines).
\end{enumerate}

\begin{figure}[htbp]
    \centering
    \begin{subtable}{.4\linewidth}
        \centering
        \resizebox{\linewidth}{!}{\begin{tabular}{lrrr}
\toprule
 & $\varrho_p$ & $\varrho_s$ & $\tau$ \\
\midrule
Retrieval (15 datasets) & 0.89 & 0.89 & 0.69 \\
Classification (12 datasets) & 0.92 & 0.88 & 0.73 \\
Clustering (11 datasets) & 0.86 & 0.85 & 0.66 \\
STS (10 datasets) & 0.92 & 0.83 & 0.63 \\
Reranking (4 datasets) & 0.84 & 0.78 & 0.64 \\
\midrule\midrule
Average (56 datasets) & \bfseries 0.94 & \bfseries 0.90 & \bfseries 0.73 \\
\midrule\midrule
Additional Classif (8 datasets) & 0.89 & 0.84 & 0.66 \\
\bottomrule
\end{tabular}
}
        \caption{NLP}
        \label{tab:common_table_correlation_nlp}
    \end{subtable}\hspace{1.5cm}%
    \begin{subtable}{.37\linewidth}
        \centering
        \resizebox{\linewidth}{!}{\begin{tabular}{lrrr}
\toprule
 & $\varrho_p$ & $\varrho_s$ & $\tau$ \\
\midrule
\textbf{A}bsorption (8 datasets) & - & 0.89 & 0.70 \\
\textbf{D}istribution (3 datasets) & - & 0.89 & 0.70 \\
\textbf{M}etabolism (8 datasets) & - & 0.94 & 0.79 \\
\textbf{E}xcretion (3 datasets) & - & 0.77 & 0.60 \\
\textbf{T}oxicity (9 datasets) & - & 0.92 & 0.75 \\
\midrule\midrule
\textbf{ADMET} (31 datasets) & - & \textbf{0.94} & \textbf{0.80} \\
\midrule\midrule

\textit{\small{DTI (1496 tasks) see~\autoref{sec:DTI}}} & - & \textit{0.88} & \textit{0.70} \\
\bottomrule
\end{tabular}
}
        \caption{Molecular Modelling}
        \label{tab:common_table_correlation_mol}
    \end{subtable}
    \begin{subfigure}{0.43\textwidth}
        \centering
        \includegraphics[width=\linewidth]{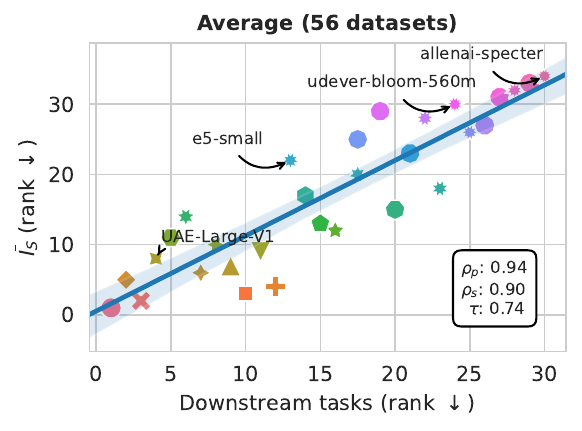}
        \caption{NLP}
        \label{fig:nlp_common_average_mteb}
    \end{subfigure}\hspace{1cm}%
    \begin{subfigure}{0.43\textwidth}
        \centering
        \includegraphics[width=\linewidth]{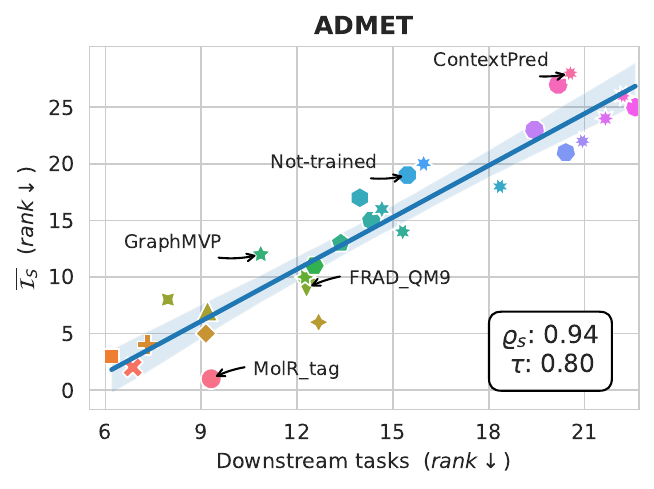}
        \caption{Molecular Modeling}
        \label{fig:mol_common_average}
    \end{subfigure}
    \caption{Correlation between {\sys} scores and downstream task performances in (a) NLP and (b) Molecular Modelling. $\varrho_p$ is the Pearson correlation, $\varrho_s$ the spearman correlation, and $\tau$ is the Kendall-Tau coefficient. 
    See \autoref{sec:full_mteb_results} for unaggregated results in NLP and \autoref{sec:complementary_resultsADMET} in molecular modeling.
    }
    \label{fig:overal_scatterplot_correlations}
\end{figure}

\FloatBarrier

\section{Text Embeddings Evaluation}
\label{Experimental-NLP}
\subsection{Experimental setting}

\paragraph{Embedders \& Datasets.} We compared $34$ models with different training objectives, training datasets, and architectures. We included embedders derived from modern LLM such as LLaMA~\cite{touvron2023llama}, Mistral~\cite{jiang2023mistral}, Gemma~\cite{gemmateam2024gemma}, Croissant~\cite{faysse2024croissantllm} and T5 encoders~\cite{ni2021sentencet5}; common embedders derived from BERT architectures~\cite{devlin2019bert, feng2022languageagnostic, reimers2019sentencebert} or RobERTa~\cite{gao2022simcse} and embedders trained on specific embeddings objectives such Angle~\cite{li2023angleoptimized}, Stella\footnote{\url{https://huggingface.co/infgrad/stella-base-en-v2}}, E5 models~\cite{wang2022text}, LaBSE~\cite{feng2022languageagnostic}. A comprehensive list of the models can be found in \autoref{sec:nlp_model_dataset_statistics}, \autoref{tab:nlp_metadata_table} with their main characteristics and links to the Huggingface Hub for reproducibility. We used them to extract embeddings for many different datasets from the MTEB benchmark such as Banking77~\cite{Casanueva2020}, Sickr~\cite{Zhang_2019}, Amazon polarity~\cite{10.1145/2507157.2507163}, SNLI~\cite{young-etal-2014-image} and IMDB~\cite{maas-EtAl:2011:ACL-HLT2011}. We provide the datasets statistics in \autoref{sec:nlp_model_dataset_statistics}, \autoref{tab:nlp_datasets}.

\paragraph{Downstream tasks evaluation.} We rely on the results released on the MTEB leaderboard\footnote{\url{https://huggingface.co/spaces/mteb/leaderboard}} and compare our rankings to the rankings and scores obtained by the different models on the different tasks. We evaluate additional tasks that are not included in the MTEB benchmark, such as tweet\_eval~\cite{barbieri2018semeval, mohammad2018semeval, barbieri2020tweeteval, van2018semeval, basile-etal-2019-semeval}, DAIR Emotion~\cite{saravia-etal-2018-carer}, agnews topic classification~\cite{Zhang2015CharacterlevelCN}, Clinc intent detection~\cite{larson-etal-2019-evaluation} PAWS-X~\cite{pawsx2019emnlp} and Rotten Tomatoes~\cite{PangLee:05a}.

\subsection{Model's Information Sufficiency analysis}
\label{sec:nlp_model_informativeness}

\begin{figure}[H]
    \centering
    \begin{minipage}[b]{0.4\textwidth}
        \begin{subfigure}{\textwidth}
            \centering
            \includegraphics[width=\linewidth]{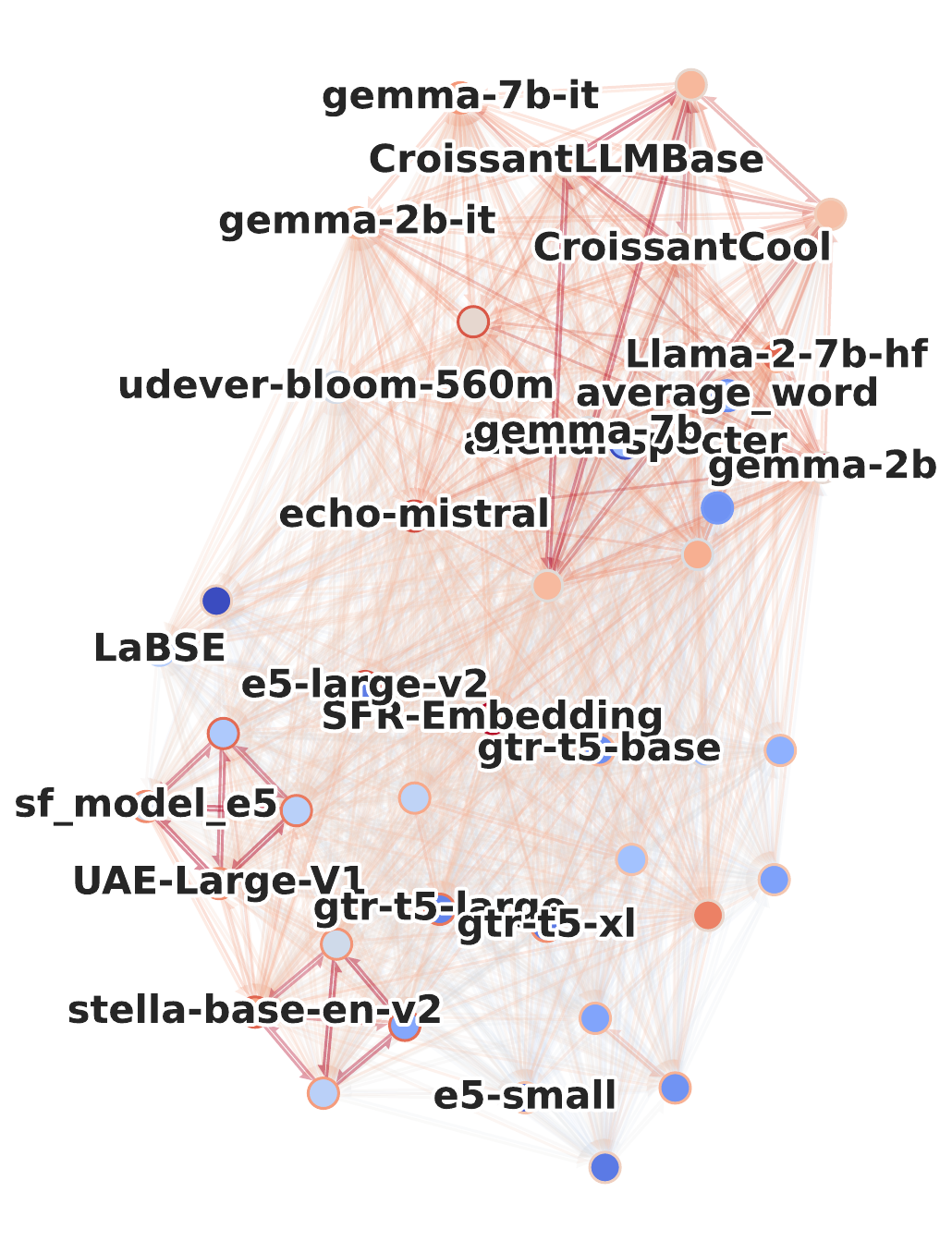}
            \caption{}
            \label{fig:predictive_mi_graph}
        \end{subfigure}
    \end{minipage}\begin{minipage}[b]{0.4\textwidth}
    {\begin{subfigure}{\textwidth}
         \centering
         \includegraphics[width=\textwidth]{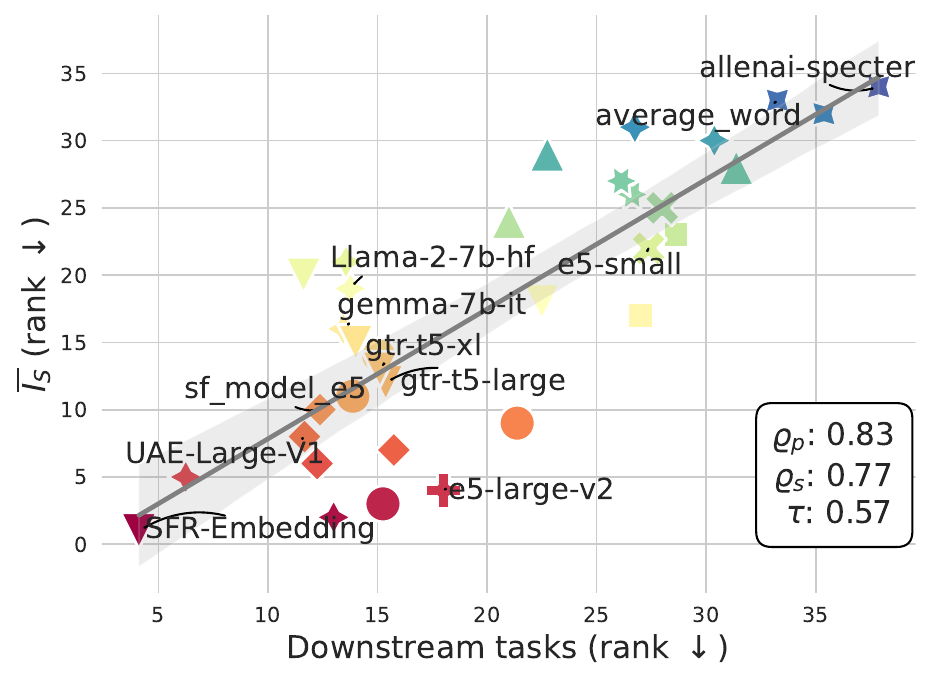}
         \caption{}
         \label{fig:my_tasks_rankings_scatter}
    \end{subfigure}
        \begin{subfigure}{\textwidth}
            \includegraphics[width=\linewidth]{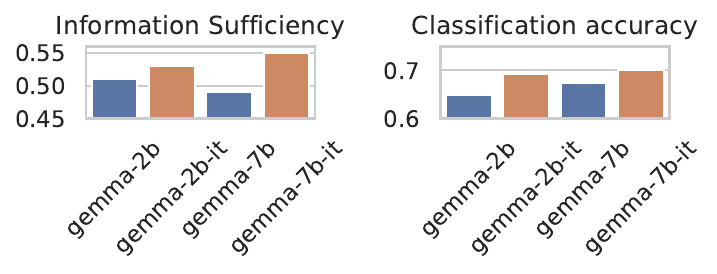}
            \caption{}
            \label{fig:nlp_instruct_finetuning}
        \end{subfigure}
    }
    \end{minipage}
    \caption{\autoref{fig:predictive_mi_graph}, presents the information sufficiency directed graph and the induced communities. \autoref{fig:my_tasks_rankings_scatter} displays the performance on additional downstream tasks and models not evaluated in the MTEB leaderboard. \autoref{fig:nlp_instruct_finetuning} shows that instruction finetuning positively impacts the models' performance on the downstream tasks and that this improvement is captured by {\sys}.}
    \label{fig:nlp_graph_it_my_tasks}
\end{figure}

\label{sec:nlp_predicting_performance}
\label{sec:nlp_mteb_benchmark}

\paragraph{Correlation with downstream tasks performance.} The MTEB Benchmark offers a natural starting point to compare models' ranking according to their performance on downstream tasks and their {\sys} score. In \autoref{fig:nlp_common_average_mteb}, we show that the {\sys} score of an embedder correlates positively with its performance on a wide range of downstream tasks, from classification and similarity tasks to retrieval and clustering tasks. Overall, our {\sys} score correlates strongly with MTEB's average score (Spearman correlation of $0.90$ and a Pearson correlation of $0.94$, see \autoref{fig:nlp_common_average_mteb}) and with the subtask performance~\autoref{tab:common_table_correlation_nlp}). We extended our experiments to a more extensive set of models not included in the MTEB benchmark and observed a similar trend (\autoref{fig:my_tasks_rankings_scatter}). Per-datasets results are reported in~\autoref{sec:full_mteb_results} and ablations in \autoref{sec:ablation_nlp}. All our results show that our estimation of the information sufficiency between models is a good proxy for the performance of the models on a wide range of tasks.

\paragraph{Embedder communities.}  The pair-wise information sufficiency evaluation between the models can be used to cluster them into communities~\cite{Blondel_2008}(\autoref{fig:predictive_mi_graph}, \autoref{fig:nlp_heatmap_informativeness})\footnote{We rely on the Louvain community detection implementation from networkx\cite{hagberg2008exploring}}. We observe that the extracted clusters group together models that are similar in their training objectives and architectures. LLM-based models such as LLaMA, Mistral, Gemma, and Croissant are clustered together, while BERT-based models share another cluster. Similarly, models trained specifically for embedding purposes, such as UAE-Large-V1 and ember-v1, are grouped together. This suggests that the ordering induced by information sufficiency is meaningful and can be used to identify models with similar properties and behaviors. Consistently with \autoref{prop-bayes-ext}, we observe that the performance of the models on the downstream tasks is similar within the same cluster (\autoref{sec:nlp_community_performance}). In addition, we found that it captures improvements by both steps of pretraining and instruction fine-tuning (\autoref{fig:nlp_instruct_finetuning}, \autoref{sec:nlp_instruction_finetuning_app})

\FloatBarrier

\section{Molecular Modeling}
\label{sec:drugdiscovery}

\subsection{Experimental setting}
\label{subsec:drugdiscovery-exp}

\noindent\textbf{Embedders.}
To process molecular data, embedders can leverage different representations of the molecules, providing an interesting benchmark to evaluate the {\sys} score.
We evaluated models derived from the molecular representation learning literature, summed up in ~\autoref{sec:models_mol}.
We considered various input modalities such as string representations (SMILES~\cite{weininger_smiles_1988}, SELFIES~\cite{Krenn_2020}), 2D-graphs by using graph neural networks (GNNs), and 3D-representations (using the TorchMD-net architecture~\cite{pelaez2024torchmdnet}).
We added a randomly initialized baseline GNN model that was not trained on any dataset.

\noindent\textbf{Datasets.}
To evaluate the information sufficiency between embedders, we compared the models on the ZINC 250k dataset\cite{irwin_zinc_2005}, designed to gather compounds that could be relevant to a wide range of therapeutic projects.
This dataset contains 250k commercially available compounds meant to be used in diverse therapeutic projects.

\noindent\textbf{Downstream tasks.}
We evaluated the embedders on 31 downstream tasks extracted from the Therapeutic Data Commons~\cite{Huang2021tdc} platform.
This section focuses on ADMET tasks (Absorption, Distribution, Metabolism, Excretion, and Toxicity).
Results on Drug-Target interaction tasks can be found in~\autoref{sec:DTI}.
Datasets collected are split into a training, validation, and test set, following the scaffold-split strategy, further described in see~\autoref{sec:complementary_resultsADMET}.

\subsection{Model's Information Sufficiency analysis}
\label{subsec:drugdiscovery-inf}

\begin{figure}
  \centering
  \begin{subfigure}{.42\textwidth}
    \centering
    \includegraphics[width=1\linewidth]{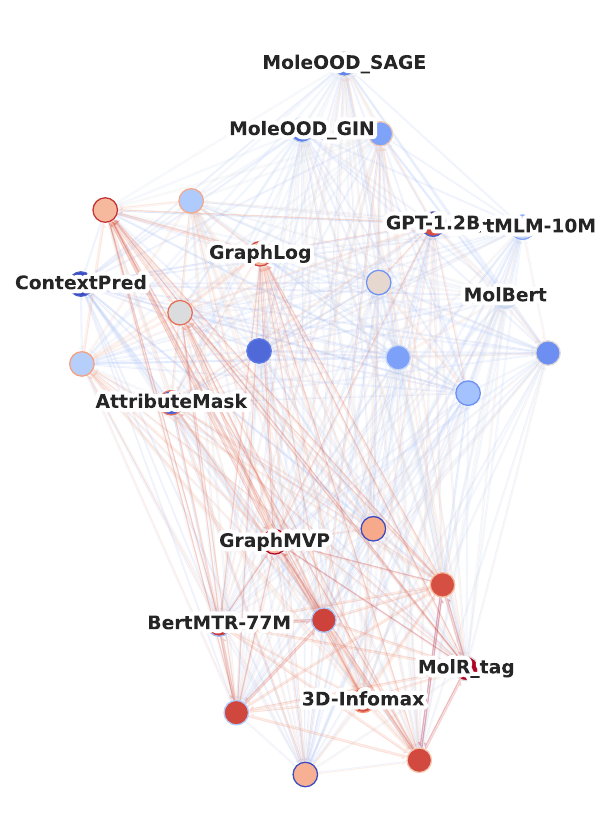}
    \caption{}
    \label{fig:mol_clustermap}
    \end{subfigure}
  \begin{subfigure}{.46\textwidth}
    \centering
    \includegraphics[width=1\linewidth]{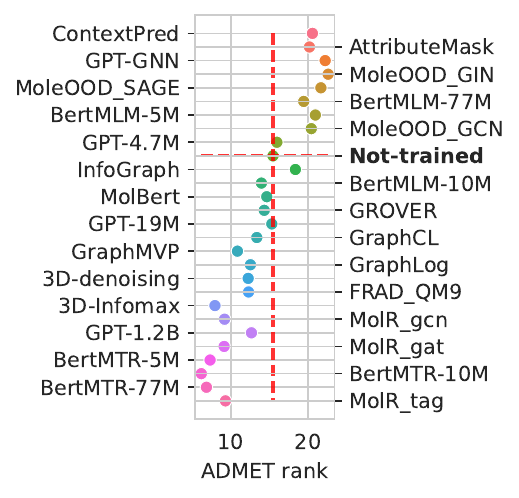}
    \vspace{0.0cm}
    \caption{}
    \label{fig:meanrank_models_group}
    \end{subfigure}
  \caption{
    (a) Pairwise information sufficiency graph between the embedders. The center color represents the ability to simulate other models, while the surrounding colors represent the ability to be simulated by other models.
  Red indicates a high ability to simulate or be simulated, while blue indicates a low ability.
  (b) Mean rank of the models (ordered by {\sys} score) on downstream tasks.
  }
\end{figure}

\textbf{Global results.}
The {\sys} score ranking is consistent with the results of the embedders on the ADMET downstream tasks, achieving a Spearman correlation of 0.95 and a Kendall-tau coefficient of 0.80, as reported in~\autoref{fig:mol_common_average}.
Detailed results for each of the 31 tasks are available in~\autoref{sec:complementary_resultsADMET} in~\autoref{tab:admet}.
Table~\ref{tab:common_table_correlation_mol} shows the correlation between the {\sys} score rankings and the performances obtained on the ADMET tasks within each category.
High correlations are achieved within most task categories, especially when large tasks are available (containing an important number of molecules).
On excretion tasks, the correlation is lower (below $0.8$), which can be explained by the fact that these tasks are the most challenging regression tasks available, where the fine-tuned models reach the lowest $R^2$ scores between $0$ and $0.2$ (see~\autoref{sec:complementary_resultsADMET}).

\textbf{Most / Least promising models.}
We observe in~\autoref{fig:meanrank_models_group} that the most promising models are the \textit{\tiny{(Chem)}}Bert-MTR models\cite{ahmad2022chemberta2}\footnote{BertMTR-$X$M stands for a \textit{\tiny{(Chem)}}Bert-MTR model trained on $X$M molecules.} and MolR\cite{wang2022chemicalreactionaware}, the former trained on SMILES representations to predict a variety of computationally available molecular properties, and the latter trained on 2D graphs to preserve equivalence of molecules w.r.t chemical reactions. Surprisingly, these models share high predictive mutual information (being assigned to the same Louvain community in \autoref{fig:mol_clustermap}), suggesting that they capture similar information despite significant differences in their training methods.
These models also appear to be the most competitive on the ADMET tasks.
 On the other hand, and consistently with Sun \textit{et al.}~\cite{sun2022does}'s observation, training methods for 2D-GNNs such as following attribute masking and context prediction objective are deemed as the least informative according to the {\sys} score.
This is explained by the simplicity of these pretraining objectives for this data modality.
These methods are also among the least competitive methods on the ADMET downstream tasks.

\noindent\textbf{NLP-inspired models.} \textit{\tiny{(Chem)}}Bert-MLM~\cite{ahmad2022chemberta2}, MolBert~\cite{fabian2020molecular} and \textit{\tiny{(Chem)}}GPT\cite{frey_neural_2023} leverage masked language model objective applied to string representations (SMILES and SELFIES).
Unsurprisingly, as seen in~\autoref{fig:mol_clustermap}, these models are clustered, suggesting they capture similar information.
However, they fail to simulate other models in the pool, resulting in low {\sys} scores, a result consistent with the known limitations of these pretraining objectives~\cite{chithrananda2020chemberta, torrisi2023do}.
A noticeable exception is \textit{\tiny{(Chem)}}GPT-1.2B (the biggest model of the pool by far), which displays a significantly higher {\sys} score.

\textbf{"Not-trained" GNN.}
\autoref{fig:meanrank_models_group} helps visualize the performances of the different models relative to our baseline "Not-trained" GNN.
Surprisingly, some models are ranked less promising than this baseline by the {\sys} score.
However, all of these less promising models obtain poorer performances on the downstream tasks.
Similarly, except for InfoGraph~\cite{sun2019infograph}, every model ranked more promising than the "Not-trained" GNN baseline and obtained better results on ADMET tasks.
This surprising result validates evaluation of the {\sys} score w.r.t this baseline.

\section{Limitations and Conclusions}
\label{sec:limitations}
\label{sec:conclusions}

We proposed a principled approach to embedding model evaluation by framing model ranking as a variation of comparing statistical experiments. Utilizing concepts of sufficiency, informativeness, and deficiency, we developed mathematically grounded metrics for pairwise comparisons between embedders without relying on labeled data in downstream tasks. Our tractable relaxation, termed information sufficiency, demonstrated strong correlations with rankings based on downstream task performance in extensive experiments. Although successful, our method still has at least two primary limitations. First, its effectiveness depends on the number and diversity of available embedders (see~\autoref{subsec:impact_nb_models}). Future work could explore using randomly initialized embedders (random projections) instead of pre-trained ones. Second, we can enhance our proxy for predicting the deficiency between models by exploring better methods (e.g., estimating the $f$-divergence) to directly learn the Markov kernel that minimizes the total variation distance, which we leave for future research.

\section*{Acknowledgments}
    This work was granted access to the HPC resources of IDRIS under the allocation 2023-AD011013290R2 made by GENCI, and enabled by support provided by Calcul Quebec and the Digital Research Alliance of Canada. We warmly thank Heitor Rapela, Banafsheh Karimian, and Eric Aubinais for their advice and comments about our work. We also owe a special highlight to Loïc Fosse for the many discussions and hindsights he provided and for the subsequent follow-up projects.

\newpage
\bibliographystyle{plain}
\bibliography{biblio}

\newpage
\appendix
\addcontentsline{toc}{section}{Appendix} %

\part{Appendix} %
\parttoc %
\newpage

\section{Background and Notation}\label{appendix_def}

We consider all alphabets to be standard Borel~\cite{crauel2002random} (i.e., isomorphic to a Borel subspace of a Polish space), encompassing virtually all practical scenarios. Each such space $\mathsf{Y}$ is equipped with its Borel $\sigma$-algebra $\mathcal{B}(\mathsf{Y})$. The set of all probability measures on $\mathsf{Y}$ is denoted by $\mathcal{P}(\mathsf{Y})$. The total variation distance between $P$ and $Q$ in $\mathcal{P}(\mathsf{Y})$ is defined as
\begin{equation}
\|P - Q\|_{\textrm{TV}} = \sup_{\mathcal{A} \in \mathcal{B}(\mathsf{Y})} |P(\mathcal{A}) - Q(\mathcal{A})|, \label{eq-def-TV}
\end{equation}
and the Kullback–Leibler divergence is defined by 
\begin{equation}
 \textrm{KL}\big( P\| Q  \big) =\left \{ \displaystyle  \begin{array}{cc}
    \displaystyle \int_{\mathsf{X}} \log  \frac{\mathsf{d} P }{\mathsf{d} Q} \mathsf{d}P &  \textrm{if }\, P \ll Q\\
      +\infty  & \textrm{otherwise.}
 \end{array}  \right.  
\end{equation}
Given a joint probability measure $P_{XY}$ in $\mathcal{P}(\mathsf{X}\times \mathsf{Y})$   induced by two random variables $X\in \mathsf{X}$ and $Y\in\mathsf{Y}$ with product measures  $P_{X} P_{Y}$, the Mutual Information  is defined as  $I(X;Y) = \textrm{KL}\big( P_{XY}\| P_{X} P_{Y}  \big) $. If $P_X \in \mathcal{P}(\mathsf{X})$ is a probability measure induced by $X\in \mathsf{X}$, the Differential Entropy is defined by 
\begin{equation}
h(X) = - \int_{\mathsf{X}} \log \frac{\mathsf{d} P_X}{\mathsf{d} \mu} \, dP_X, 
\end{equation}
where $\mu$ denotes the Lebesgue measure. Similarly, it is possible to define the conditional entropy of $Y$ given $X$ which is denoted by $h(Y|X)$. The mutual information satisfies the identities $I(X;Y) = h(Y) - h(Y|X) = h(X) - h(X|Y) $ and $h(Y|X)\leq h(Y)$ (see~\cite{Cover91} for further details).  

A Markov (or transition probability) kernel between $\mathsf{X}$ and $\mathsf{Y}$ is a mapping $T : \mathcal{B}(\mathsf{Y}) \times \mathsf{X} \to [0,1]$, satisfying $T(\cdot|x) \in \mathcal{P}(\mathsf{Y})$  for all $x \in \mathsf{X}$ and $T(\mathcal{B}|\cdot)$ being a measurable function on $\mathsf{X}$ for any $B \in \mathcal{B}(\mathsf{Y})$. The space of all such $T$ is denoted by $\mathcal{K}(\mathsf{Y}|\mathsf{X})$. In cases where both $\mathsf{Y}$ and $\mathsf{X}$ are finite, any $\mathcal{K}(\mathsf{Y}|\mathsf{X})$ is represented as a stochastic matrix with elements $T(y|x)$, $(x,y) \in \mathsf{X} \times \mathsf{Y}$.
Every $T \in \mathcal{K}(\mathsf{Y}|\mathsf{X})$ induces a mapping $\mathcal{P}(\mathsf{X}) \longrightarrow  \mathcal{P}(\mathsf{Y})$, denoted by $T$, mapping any $P \in \mathcal{P}(\mathsf{X}) $ to $Q = T\!\circ\!P \in \mathcal{P}(\mathsf{Y}) $, where
\begin{equation}
Q(B) = (T\!\circ\!P)(B) \triangleq \int_{\mathsf{X}} T(B|x)P(\mathsf{d}x), \quad \forall B \in \mathcal{B}(\mathsf{Y}).
\end{equation}
We denote the composition of Markov kernels by juxtaposition: for $M\in \mathcal{K}(\mathsf{Z}|\mathsf{Y})$ and $T \in \mathcal{K}(\mathsf{Y}|\mathsf{X})$, their composition $M\!\circ\!T \in \mathcal{T}(\mathsf{Z}|\mathsf{X})$ is defined by 
\begin{equation}
(M\!\circ\! T)(Z|x) \triangleq \int_{\mathsf{Y}}  M(Z|y) T(\mathsf{d}y|x) , \quad \forall x\in \mathsf{X}, \, Z \in \mathcal{B}(\mathsf{Z}).
\end{equation}
 We define the average of the total variation distance between two  Markov kernels $T, T^\prime \in \mathcal{K}(\mathsf{Y}|\mathsf{X})$ as follows: 
\begin{equation}
 \mathbb{E} \|T - T^\prime \|_{\textrm{TV}} \triangleq  \mathbb{E} \|T(\cdot|X) - T^\prime(\cdot |X)\|_{\textrm{TV}}. \label{eq-def-TV-Kernel}
\end{equation}
A statistical model is a triple $\mathcal{M}_U \equiv \big(\mathsf{U}, \mathcal{B}(\mathsf{U}), ( P_{U|Y}(\cdot|y):y\in \mathsf{Y} )\big) $, where $(\mathsf{U}, \mathcal{B}(\mathsf{U}) ) $  is a sample space;  $\mathsf{Y}$ is a concept space, and $P_{U|Y}: \mathcal{B}(\mathsf{U}) \times \mathsf{Y} \to [0,1]$  is a Markov kernel (or transition probability).

\section{Proofs Theoretical Results}
\label{sec:proof}
\begin{prop*}[Relationships of sufficiency and information] The following relationships hold:
\begin{itemize}
    \item[(i)] \textbf{Sufficiency $\Rightarrow$ informativeness.} If the embedding model $P_{U|X}$ is  sufficient for the embedding model $P_{V|X}$, i.e. $U \succcurlyeq_S V$, then $U \succcurlyeq_I V$. However, \textbf{Informativeness $\nRightarrow$ sufficiency.} %
    \item[(ii)]  \textbf{Informativeness $\Rightarrow$ higher capacity  to distinguish  concepts.} If the embedding model $P_{U|X}$ is more informative  than embedding model $P_{Z|X}$, i.e. $U  \succcurlyeq_I V$, then   
    \[
    \textrm{KL}\big( P_{U|Y}(\cdot|y_0)\| P_{U|Y} (\cdot|y_1) \big) \geq     \textrm{KL}\big( P_{V|Y} (\cdot|y_0)\| P_{V|Y} (\cdot|y_1) \big),
    \]
for any pair of concepts $(y_0,y_1)\in \mathsf{Y}\times \mathsf{Y}$ and all probability distributions  $ P_{YX}$. 
\end{itemize} 
\label{Prop1_app}
\end{prop*}
\subsection{Proof Proposition~\ref{Prop1}}

\begin{proof}
It is immediate to check that the data-processing inequality and the Markov chain $ Y \leftrightarrow X \leftrightarrow U \leftrightarrow V $  implies the relation in claim (i).  On the other hand, the non-equivalence is proved by means of an explicit counterexample~\cite{Korner1977}. Given any $0<p<1/2$ with $\bar{p}=1-p$. For some $\epsilon, \delta >0$, consider two discrete embedding models defined by the following matrices:
\begin{equation}
P_{V|X} = 1/2\left( \begin{array}{cc}
    1 +\epsilon & 1 - \epsilon \\
     1 & 1  
\end{array} \right) P_{U|X}  \quad \textrm{with} \quad 
P_{U|X}  = \left( \begin{array}{cc}
    p & \bar{p}  \\
     p+\delta  & \bar{p} - \delta   
\end{array} \right). 
\end{equation} 
By taking $\epsilon, \delta >0$ small enough, both $P_{U|X} $ and $P_{V|X} $ are stochastic matrices. It follows that $P_{V|X} $ is not a degraded version of $P_{U|X} $ but provided that $\epsilon, \delta$ are sufficient small, the embedding model $P_{U|X} $ is more informative than $P_{V|X} $, which proves the claim. 

In order to show (ii), let $P_{U|Y}$ and $P_{V|Y}$ be the corresponding probability measures induced by $ P_{X|U}(\cdot | u ) $ via the embedding models:
\begin{equation}
P_{U|Y}(U|y) = \int_{\mathsf{X}}  P_{U|X} (U|x) P_{X|Y}(\mathsf{d}x|y) , \quad \forall y\in \mathsf{Y}, \, U \in \mathcal{B}(\mathsf{U}),    \label{eq-KL-1}
\end{equation}
and 
 \begin{equation}
   P_{V|Y}(V|y) = \int_{\mathsf{X}}  P_{V|X} (V|x) P_{X|Y}(\mathsf{d}x|y) , \quad \forall y\in \mathsf{Y}, \, V \in \mathcal{B}(\mathsf{V}),  \label{eq-KL-2}   
\end{equation}
for any $y\in \mathsf{Y}=\{y_0,y_1\}$. For a $0\leq \lambda \leq 1$, let $P_{X|Y}(\cdot |y_0)\in \mathcal{P}(\mathsf{X})$ and $P_{X|Y}(\cdot |y_1)\in \mathcal{P}(\mathsf{X})$ be two arbitrary probability measures on $\mathsf{X}$. Let $P_{X|Y}(X |y)$ be defined by 
 \[
  P_{X|Y}(X|y) = \mathbbm{1} [y=y_0] P_{X|Y}(X |y_0)  + \mathbbm{1} [y=y_1] P_{X|Y}(X |y_1). 
 \]
By replacing it into equations  \eqref{eq-KL-1} and \eqref{eq-KL-2}, we obtain  
\begin{eqnarray}
  P_{U|Y}(U|y) = \mathbbm{1} [y=y_0] P_{U|Y}(U |y_0)  + \mathbbm{1} [y=y_1] P_{U|Y}(U |y_1)   \\
    P_{V|Y}(V|y) = \mathbbm{1} [y=y_0] P_{V|Y}(V |y_0)  + \mathbbm{1} [y=y_1] P_{V|Y}(V |y_1) 
\end{eqnarray}
and let $P_Y(y_0) = \lambda $ and $P_Y(y_1) = 1- \lambda $. The above probability measures correspond to a quadruple of random variables:  $(Y_\lambda,X_\lambda,U_\lambda,V_\lambda) \in \mathcal{P}(\mathsf{Y}\times \mathsf{X} \times \mathsf{U} \times \mathsf{V})$. Consider the function $f(\lambda)$ defined by 
\[
f(\lambda) = I(Y_\lambda ; U_\lambda) - I(Y_\lambda ; V_\lambda). 
\]
It is not difficult to check that $f(\lambda)\geq 0$ for all $0\leq \lambda \leq 1$, and $f(0)=0$ which requires that $f^\prime (0)\geq 0$. By taking the differentiation, we obtain 
\[
   f^\prime (0)= \textrm{KL}\big( P_{U|Y}(\cdot|y_0)\| P_{U|Y} (\cdot|y_1) \big) -    \textrm{KL}\big( P_{V|Y} (\cdot|y_0)\| P_{V|Y} (\cdot|y_1) \big)\geq 0,
    \]
which implies the claim (ii). This concludes the proof.
\end{proof}

\subsection{Comments about capacity to distinguish concepts}\label{sec_distinguish}

Notice that  the KL divergence between the induced  distributions of the resulting embedding is not less for the embedding model $U$ than $Z$. Indeed, consider the case  of binary classification  $\mathsf{Y}=\{y_0,y_1\}$ with uniformly distributed concepts. Pinsker's inequality~\cite{Tsybakov:1315296} together with claim (iii) imply 
\[
\textrm{KL}\big( P_{U|Y}(\cdot|y_0)\| P_{U|Y} (\cdot|y_1) \big) \geq 2 \| P_{V|Y} (\cdot|y_0) - P_{V|Y} (\cdot|y_1) \|_{\textrm{TV}}^2.  
\]
From which, it is easy to verify that the accuracy of the expected Bayes accuracy  of the optimal classifier  based on $V$ is upper bounded by~\cite[Lemma 2.1]{Tsybakov:1315296}:
\[
\sup_{\psi} \Pr(\psi(V) = Y) \leq  1- \frac{1}{2}\exp\left( - \textrm{KL}\big( P_{U|Y}(\cdot|y_0)\| P_{U|Y} (\cdot|y_1) \big)\right),
\]
where the exponent in the upper bound is subject to the discriminating capacity  through the KL divergence of the embedding model $U$ on $\mathsf{U}$.  

\subsection{Proof of Proposition~\ref{prop-bayes} and Corollary~\ref{prop-bayes-ext}} \label{Appex-prop-bayes}

We begin with the proof of Proposition~\ref{prop-bayes}. 

\begin{proof}  Clearly, the assumption (i) implies the statement (ii) by Data-Processing. Conversely, let us assume point (ii) holds. This means that, for every probbaility distribution $P_X$ and all conditional  probability distributions $P_{Y|X}$, there exists $\rho_U:\mathsf{U} \rightarrow\mathcal{P}(\mathsf{Y})$ such that 
\begin{eqnarray}
& \displaystyle    \sum_{(y,x) \in \mathsf{Y}\times  \mathsf{X}}  P_{YX}(y,x)   \int_{\mathsf{U}} \rho_U(y|u)     P_{U|X}(\mathsf{d}u|x)     \geq\nonumber   \\
& \quad\quad\quad\quad\quad\sup\limits_{\rho_V} \displaystyle  \sum_{(y,x) \in \mathsf{Y}\times  \mathsf{X}}  P_{YX}(y,x)   \int_{\mathsf{V}}   \rho_V(y|v)   P_{V|X}(\mathsf{d}v|x)   , \label{eq-sup_inequality}
\end{eqnarray}
where $\rho_V:\mathsf{V} \rightarrow \mathcal{P}(\mathsf{Y})$ is a (possibly randomized) inference procedure is transition probabilities, which the learner can optimize to maximize the guessing probability.

Let the decision rule  $\rho_V(y|v) = \mathbb{1}[v\in {A}_y]$ for any partition $\{{A}_y\}_{y\in\mathsf{Y}}$ of $\mathsf{V}$ with ${A}_y \in \mathcal{B}(\mathsf{V})$. Then, for any $P_{Y|X}$, expression~\eqref{eq-sup_inequality} implies the existence  there exists $\rho_U(y|u) $ such that
\begin{eqnarray}
& \displaystyle\sum\limits_{(y,x)\in \mathsf{Y}\in \mathsf{X}} P_{YX}(y,x)  \left[\displaystyle   \int_{\mathsf{V}}  P_{V|X}(\mathsf{d} v|x)\mathbb{1}[v\in {A}_y] -  \int_{\mathsf{U}}   \rho_U(y|u) P_{U|X}(\mathsf{d}u|x)   \right]\\
 & = \quad \quad \displaystyle \sum\limits_{(y,x) \in \mathsf{Y}\times \mathsf{X}} P_{YX}(y,x)  \left[\displaystyle   \int_{ {A}_y}  P_{V|X}(\mathsf{d} v|x) -   \int_{\mathsf{U}}   \rho_U(y|u) P_{U|X}(\mathsf{d}u|x)   \right]\leq 0.
\end{eqnarray}
However, we can rewrite the last expression as: 
\begin{eqnarray}
 \sup_{P_{Y|X}} \inf_{\rho_U}  \displaystyle \sum\limits_{(y,x)\in \mathsf{Y}\times \mathsf{X}} P_{YX}(y,x)  \left[\displaystyle   \int_{\mathcal{A}_y}  P_{V|X}(\mathsf{d} v|x) -   \int_{\mathsf{U}}   \rho_U(y|u) P_{U|X}(\mathsf{d}u|x)   \right]\leq 0.
\end{eqnarray}
By applying the  minimax theorem~\cite{rockafellar-1970a}, it is possible to  exchange the order of the inf and the sup, which yields:
\begin{eqnarray}
&&\inf_{\rho_U}\sup_{\{A_y\}} \, \mathbb{E}\left[\sup_{P_{Y|X}}\sum\limits_{y\in \mathsf{Y} } P_{Y|X}(y|X) \Gamma\big((y,X),\rho_U\big)\right] \leq 0\\
&& \Gamma\big((y,x),\rho_U\big) \triangleq 
\left[\displaystyle   \int_{\mathcal{A}_y}  P_{V|X}(\mathsf{d} v|x) -   \int_{\mathsf{U}}   \rho_U(y|u) P_{U|X}(\mathsf{d}u|x)   \right].
\end{eqnarray}
We observe that
\begin{equation}
\sum\limits_{y\in \mathsf{Y} } \Gamma\big((y,x),\rho_U\big) = 0, 
\end{equation}
for each $x\in \mathsf{X}$ and thus, 
\begin{equation}
\max\limits_{y\in \mathsf{Y} }  \Gamma\big((y,x),\rho_U\big)\geq  0, 
\end{equation}
where the equality holds if and only if $\Gamma\big((y,x),\rho_U\big) = 0$ for all $y\in \mathsf{Y}$, for each $x\in \mathsf{X}$, since by contradiction otherwise 
\begin{equation}
\sum\limits_{y\in \mathsf{Y}}  \Gamma\big((y,x),\rho_U\big) < 0.  
\end{equation}
Therefore, 
\begin{eqnarray}
&\inf\limits_{\rho_U} 
\sup\limits_{\{A_y\}} \mathbb{E}\left[\sup\limits_{P_{Y|X}} \sum\limits_{y\in \mathsf{Y}} P_{Y|X}(y|X) \left(\displaystyle   \int_{\mathcal{A}_y}  P_{V|X}(\mathsf{d} v|X) -   \int_{\mathsf{U}}   \rho_U(y|u) P_{U|X}(\mathsf{d}u|X)   \right) \right]  \nonumber\\
& =  \inf\limits_{\rho_U}\sup\limits_{\{A_y\}} \mathbb{E}\left[\max\limits_{y\in \mathsf{Y}} \Gamma\big((y,X),\rho_U\big)\right],
\end{eqnarray}
 which means the maximum is achieved by degenerate random variables $Y=f(X)$ achieving the maximum for each $x\in \mathsf{X}$. Consequently, we have that 
\begin{eqnarray}
\inf\limits_{\rho_U} \mathbb{E} \| P_{V|X} -  \rho_U \!\circ\! P_{U|X} \|_{\textrm{TV}}
&=  &\inf\limits_{\rho_U} \sup\limits_{\{A_y\}} \,  \mathbb{E}\left[\max\limits_{y\in \mathsf{Y}} \Gamma\big((y,X),\rho_U\big)\right] = 0, 
\end{eqnarray}
hence $\inf\limits_{\rho_U}  \mathbb{E} \| P_{V|X} -  \rho_U \!\circ\! P_{U|X} \|_{\textrm{TV}} = 0 $, and so the  existence of the transition probability $\rho_U$ such that   $P_{U|X}$ is sufficient for $P_{V|X}$. This concludes the proof of the Proposition~\ref{prop-bayes}. 
\end{proof}

We now show the proof of  Corollary~\ref{prop-bayes-ext}. 

\begin{proof} 

Continuing from the proof of Proposition~\ref{prop-bayes}, which remains unchanged, until one demonstrates that for $\varepsilon > 0$,
\begin{eqnarray}\label{eq-rhs}
\inf\limits_{\rho_U} \sup\limits_{\{A_y\}} \,  \mathbb{E}\left[\max\limits_{y\in \mathsf{Y}} \Gamma\big((y,X),\rho_U\big)\right]  \leq   \varepsilon  . 
\end{eqnarray}
To proceed from this point, let us now examine the following quantity:
\[
 \sum\limits_{y\in \mathsf{Y} }  |\Gamma\big((y,x),\rho_U\big) |.
\]

The above quantity is the induced $\ell_1$-norm distance between $  \int_{\mathcal{A}_y}  P_{V|X}(\mathsf{d} v|x)$ and $\int_{\mathsf{U}}   \rho_U(y|u) P_{U|X}(\mathsf{d}u|x) $. Since, for all $x\in \mathsf{X}$, 
$$
\sum\limits_{y\in \mathsf{Y} } \Gamma\big((y,x),\rho_U\big) =0, 
$$
we have that
\[
\sum\limits_{y\in \mathsf{Y} } | \Gamma\big((y,x),\rho_U\big)| = 2 \sum\limits_{y\in \mathsf{Y} : \Gamma((y,x),\rho_U)\geq 0} \Gamma\big((y,x),\rho_U\big),\quad \textrm{for all $x\in \mathsf{X} $}
\]
which implies that, for the strategy $\rho_U$ achieving the left-hand side of Eq.~\eqref{eq-rhs},
\[
\sup\limits_{\{A_y\}}\mathbb{E}  \sum\limits_{y\in \mathsf{Y} } | \Gamma\big((y,X),\rho_U\big)| \leq 2 |\mathsf{Y} | \sup\limits_{\{A_y\}} \mathbb{E}  \left[\max\limits_{y\in \mathsf{Y}} \Gamma\big((y,X),\rho_U\big)\right]\leq  2| \mathsf{Y}| \varepsilon  ,\quad \textrm{for all $x\in \mathsf{X} $}
\]
Hence,
\[
 \sup\limits_{\{A_y\}} \, \mathbb{E}  \sum\limits_{y\in \mathsf{Y} } \left|\displaystyle   \int_{\mathcal{A}_y}  P_{V|X}(\mathsf{d} v|X) -   \int_{\mathsf{U}}   \rho_U(y|u) P_{U|X}(\mathsf{d}u|X)   \right| \leq   2| \mathsf{Y}| \varepsilon .
\]
This concludes the proof of the Corollary~\ref{prop-bayes-ext}. 
\end{proof}

\subsection{Example of comparisons of statistical experiments}\label{examples}

\begin{example} [Statistical experiments with Gaussian embedding models] Let $U|x$ and $V|x$ be independently normally distributed as $\mathcal{N}(x, \sigma^2) $  and $\mathcal{N}(x, \epsilon^2\sigma^2) $ , respectively, with $0 < \epsilon  < 1 $.
\begin{itemize}
    \item Case of $\sigma^2=\sigma^2_0$ known. Here $U \succcurlyeq_{S} V$ since $V+\nu|x$ has the same distribution as $U|x$ when $\nu \sim\mathcal{N}(0, (1-\epsilon^2)\sigma_0^2) $. That is $V$ is strictly more informative than $U$. However, $U$ is strictly more informative than $V$. 
    \item  Case of $ x = 0$ known. One can observe that $U \approx V$ since the variables $V/ \epsilon$ have the same distribution as the $U$, and the variables $\epsilon U$ have the same distribution as the $V$.
    \item Case of $ x $  and $ \sigma $   unknown. Surprisingly, in this case  $U$ and $V$ are not comparable.
\end{itemize}    
\end{example}

\subsection[Sufficiency and Inference Procedures with Embedding Models]{Sufficiency and Inference Procedures with Embedding Models~\cite{blackwell1951comparison, le1964sufficiency}}\label{appendix_propo_one-task}

\begin{prop}[Sufficiency and risks of a given task on embedding models~\cite{blackwell1951comparison, le1964sufficiency}]
An embedding model $P_{U|Y}\in \mathcal{K}(\mathsf{U}|\mathsf{Y})$  is deemed to be sufficient for another one $P_{V|Y}\in \mathcal{K}(\mathsf{V}|\mathsf{Y})$ if and only if, for any bounded loss function $\ell$ where $\|\ell\|_{\infty} \leq 1$, and for any inference procedure $\rho_V:\mathsf{V} \rightarrow \mathsf{Y}$, there exists a inference procedure (possibly randomized) $\rho_U:\mathsf{U} \rightarrow\mathcal{P}(\mathsf{Y})$  such that the resulting statistical risks satisfy 
\begin{equation}\label{eq-domination}
    \mathcal{R}_{y}(P_{U|Y}, \rho_U, \ell)\leq \mathcal{R}_{y}(P_{V|Y}, \rho_V, \ell) , \quad \textrm{ for all $y \in \mathsf{Y}$}.
\end{equation}
Here we denote by $\mathcal{R}_{y}(P_{U|Y}, \rho_Y, \ell)$ and $\mathcal{R}_{y}(P_{V|Y}, \rho_V, \ell)$ the statistical risks for the corresponding inference frameworks, respectively. 
\end{prop}
\begin{rem}
The restriction $\|\ell\|_{\infty} \leq 1$ is irrelevant here. However, we opt for simplicity and limit our focus to situations where one encounters dominated statistical models with Polish sample spaces. In essence, various extensions do not significantly alter the conceptual aspects of the underlying statistical problem (see~\cite{Torgersen_1991} for further details). Rather, they primarily reflect the complexity of its measure-theoretic formulation.     
\end{rem}

\subsection[Deficiency and Expected Risk]{Deficiency and Expected Risk~\cite{le1964sufficiency}}
\label{subsec:le-cam}

In 1964, Le Cam~\cite{le1964sufficiency} clarified the relationship between the sufficiency of an embedding model on a given task and its expected risk on this task. The following theorem provides a formal statement of this relationship.

\begin{thm}[Le Cam~\cite{le1964sufficiency}]
    Let $\varepsilon > 0$ be fixed. Then, $\delta(P_{U|Y} \rightarrow P_{V|Y} ) < \varepsilon$ if and only if, for any bounded loss function $\ell$ where $\|\ell\|_{\infty} \leq 1$, and for any inference procedure $\rho_V$ using the embedding model $P_{V|Y}$, there exists a inference procedure (possibly randomized) $\rho_U$ based the embedding model $P_{U|Y}$ such that the risks satisfy
    $
    \mathcal{R}_{y}(P_{U|Y}, \rho_U, \ell)-\varepsilon  \leq \mathcal{R}_{y}(P_{V|Y}, \rho_V, \ell) , \quad \textrm{ for all $y \in \mathsf{Y}$}.
$
\end{thm}

\section{NLP Experiment Details}
\label{sec:nlp_experiment_details}

In this section, we provide all the necessary experimental details to reproduce the experiments in NLP. For the {\sys} score estimation, please see \autoref{sec:emir_estimation}. First, we detail the models and datasets used in the experiments. We provide the training details of the downstream tasks, and finally, we present the comprehensive results of the NLP experiments.

\subsection{Models and Datasets statistics}
\label{sec:nlp_model_dataset_statistics}

In \autoref{tab:nlp_metadata_table}, we provide the metadata of the models used in the NLP experiments and their scores on the MTEB benchmark when they exist. We provide in \autoref{tab:nlp_datasets} the statistics of the datasets used to evaluate the {\sys} score.

\begin{table}
\caption{Metadata of the evaluated models and their information sufficiency.}
\label{tab:nlp_metadata_table}
\centering \resizebox{0.8\textwidth}{!}{ \begin{tabular}{lrrr}
\toprule
 & Dim. & Max Tokens & $\bar{I_{S}}$ \\
Model &  &  &  \\
\midrule
\href{https://huggingface.co/SFR-Embedding-Mistral}{SFR-Embedding-Mistral} & 4096 & 32768 & \bfseries 0.59 \\
\href{https://huggingface.co/echo-mistral-7b-instruct-lasttoken}{echo-mistral-7b-instruct-lasttoken} & 4096 & 32768 & \bfseries 0.58 \\
\href{https://huggingface.co/stella-base-en-v2}{stella-base-en-v2} & 768 & 512 & \bfseries 0.57 \\
\href{https://huggingface.co/e5-large-v2}{e5-large-v2} & 1024 & 512 & \bfseries 0.57 \\
\href{https://huggingface.co/GritLM-7B}{GritLM-7B} & 4096 & 32768 & \bfseries 0.56 \\
\href{https://huggingface.co/ember-v1}{ember-v1} & 1024 & 512 & \bfseries 0.56 \\
\href{https://huggingface.co/gte-large}{gte-large} & 1024 & 512 & \bfseries 0.56 \\
\href{https://huggingface.co/UAE-Large-V1}{UAE-Large-V1} & 1024 & 512 & \bfseries 0.55 \\
\href{https://huggingface.co/gte-base}{gte-base} & 768 & 512 & \bfseries 0.55 \\
\href{https://huggingface.co/sf\_model\_e5}{sf\_model\_e5} & 1024 & 512 & \bfseries 0.55 \\
\href{https://huggingface.co/GIST-Embedding-v0}{GIST-Embedding-v0} & 768 & 512 & \bfseries 0.55 \\
\href{https://huggingface.co/gtr-t5-large}{gtr-t5-large} & 768 & 512 & \bfseries 0.55 \\
\href{https://huggingface.co/gtr-t5-xl}{gtr-t5-xl} & 768 & 512 & \bfseries 0.55 \\
\href{https://huggingface.co/bge-base-en-v1.5}{bge-base-en-v1.5} & 768 & 512 & \bfseries 0.54 \\
\href{https://huggingface.co/sentence-t5-large}{sentence-t5-large} & 768 & 512 & \bfseries 0.54 \\
\href{https://huggingface.co/gemma-7b-it}{gemma-7b-it} & N/A & N/A & \bfseries 0.54 \\
\href{https://huggingface.co/gte-tiny}{gte-tiny} & 384 & 512 & \bfseries 0.53 \\
\href{https://huggingface.co/gtr-t5-base}{gtr-t5-base} & 768 & 512 & \bfseries 0.53 \\
\href{https://huggingface.co/Llama-2-7b-hf}{Llama-2-7b-hf} & N/A & N/A & \bfseries 0.53 \\
\href{https://huggingface.co/sentence-t5-xl}{sentence-t5-xl} & 768 & 512 & \bfseries 0.53 \\
\href{https://huggingface.co/gemma-2b-it}{gemma-2b-it} & N/A & N/A & \bfseries 0.52 \\
\href{https://huggingface.co/e5-small}{e5-small} & 384 & 512 & \bfseries 0.52 \\
\href{https://huggingface.co/bge-micro-v2}{bge-micro-v2} & 384 & 512 & \bfseries 0.51 \\
\href{https://huggingface.co/all-distilroberta-v1}{all-distilroberta-v1} & N/A & N/A & \bfseries 0.51 \\
\href{https://huggingface.co/multilingual-e5-small}{multilingual-e5-small} & 384 & 512 & \bfseries 0.51 \\
\href{https://huggingface.co/msmarco-bert-co-condensor}{msmarco-bert-co-condensor} & 768 & 512 & \bfseries 0.51 \\
\href{https://huggingface.co/sup-simcse-bert-base-uncased}{sup-simcse-bert-base-uncased} & 768 & 512 & \bfseries 0.50 \\
\href{https://huggingface.co/all-MiniLM-L6-v2}{all-MiniLM-L6-v2} & 384 & 512 & \bfseries 0.50 \\
\href{https://huggingface.co/all-mpnet-base-v2}{all-mpnet-base-v2} & 768 & 514 & \bfseries 0.49 \\
\href{https://huggingface.co/udever-bloom-560m}{udever-bloom-560m} & 1024 & 2048 & \bfseries 0.49 \\
\href{https://huggingface.co/LaBSE}{LaBSE} & 768 & 512 & \bfseries 0.47 \\
\href{https://huggingface.co/average\_word\_embeddings\_komninos}{average\_word\_embeddings\_komninos} & 300 & N/A & \bfseries 0.42 \\
\href{https://huggingface.co/average\_word\_embeddings\_glove.6B.300d}{average\_word\_embeddings\_glove.6B.300d} & 300 & N/A & \bfseries 0.41 \\
\href{https://huggingface.co/allenai-specter}{allenai-specter} & 768 & 512 & \bfseries 0.38 \\
\bottomrule
\end{tabular}
}
\end{table}

\begin{table}\centering
\caption{Statistics of the datasets used as umbrella datasets for {\sys} informativeness evaluation.}
\label{tab:nlp_datasets}
\begin{tabular}{llr}
\toprule
 &  & Size \\
Dataset & Split &  \\
\midrule
\multirow[c]{2}{*}{ag\_news} & test & 7600 \\
 & train & 120000 \\
\cline{1-3}
amazon\_polarity & test & 100000 \\
\cline{1-3}
banking77 & test & 3080 \\
\cline{1-3}
biosses-sts & test & 182 \\
\cline{1-3}
\multirow[c]{3}{*}{paws-x;en} & test & 2000 \\
 & train & 49401 \\
 & validation & 2000 \\
\cline{1-3}
\multirow[c]{3}{*}{rotten\_tomatoes} & test & 1066 \\
 & train & 8530 \\
 & validation & 1066 \\
\cline{1-3}
sickr-sts & test & 6077 \\
\cline{1-3}
\multirow[c]{2}{*}{snli} & test & 13132 \\
 & validation & 13134 \\
\cline{1-3}
\multirow[c]{3}{*}{sst2} & test & 1821 \\
 & train & 67349 \\
 & validation & 872 \\
\cline{1-3}
sts12-sts & test & 4946 \\
\cline{1-3}
sts13-sts & test & 2638 \\
\cline{1-3}
sts14-sts & test & 6351 \\
\cline{1-3}
sts15-sts & test & 5170 \\
\cline{1-3}
\multirow[c]{2}{*}{stsbenchmark-sts} & test & 2552 \\
 & validation & 2910 \\
\cline{1-3}
\multirow[c]{3}{*}{tweet\_eval;emoji} & test & 50000 \\
 & train & 45000 \\
 & validation & 5000 \\
\cline{1-3}
\multirow[c]{3}{*}{tweet\_eval;emotion} & test & 1421 \\
 & train & 3257 \\
 & validation & 374 \\
\cline{1-3}
\multirow[c]{3}{*}{tweet\_eval;sentiment} & test & 12284 \\
 & train & 45615 \\
 & validation & 2000 \\
\cline{1-3}
wiki-paragraphs & validation & 100000 \\
\cline{1-3}
\bottomrule
\end{tabular}
\end{table}

\subsection{Downstream tasks training details.}

All the downstream tasks are trained in the exact same way. We use a dense classifier with two hidden layers of dimension $256$ and train for two epochs using ADAM~\cite{adam} with a learning rate of $10^{-3}$, on the official training set and evaluated on either the validation or test set when they are available (with respect to the Huggingface datasets). We do not perform early stopping or selection using the validation set.

\subsection{NLP Comprehensive Results}

We provide in this section the unaggregated results for the main NLP experiments presented in the main text, then we provide numerous ablation studies and additional results to address different aspects of the {\sys} score of the embeddings in NLP.

\subsubsection{Full MTEB Benchmark Results}
\label{sec:full_mteb_results}

\begin{table}
\caption{Summary of the evaluated embedders with their performance on the MTEB benchmark.}
\label{tab:nlp_mteb_perfs_table}
\resizebox{\textwidth}{!}{ \begin{tabular}{lrrrrrr}
\toprule
 & Average & Classification & Clustering & Reranking & Retrieval & STS \\
Model &  &  &  &  &  &  \\
\midrule
\href{https://huggingface.co/SFR-Embedding-Mistral}{SFR-Embedding-Mistral} & 67.56 & 78.33 & 51.67 & 60.64 & 59.00 & 85.05 \\
\href{https://huggingface.co/echo-mistral-7b-instruct-lasttoken}{echo-mistral-7b-instruct-lasttoken} & 64.68 & 77.43 & 46.32 & 58.14 & 55.52 & 82.56 \\
\href{https://huggingface.co/stella-base-en-v2}{stella-base-en-v2} & 62.61 & 75.28 & 44.90 & 58.78 & 50.10 & 83.02 \\
\href{https://huggingface.co/e5-large-v2}{e5-large-v2} & 62.25 & 75.24 & 44.49 & 56.61 & 50.56 & 82.05 \\
\href{https://huggingface.co/GritLM-7B}{GritLM-7B} & 66.76 & 79.46 & 50.61 & 60.49 & 57.41 & 83.35 \\
\href{https://huggingface.co/ember-v1}{ember-v1} & 63.54 & 75.99 & 45.58 & 60.04 & 51.92 & 83.34 \\
\href{https://huggingface.co/gte-large}{gte-large} & 63.13 & 73.33 & 46.84 & 59.13 & 52.22 & 83.35 \\
\href{https://huggingface.co/UAE-Large-V1}{UAE-Large-V1} & 64.64 & 75.58 & 46.73 & 59.88 & 54.66 & 84.54 \\
\href{https://huggingface.co/gte-base}{gte-base} & 62.39 & 73.01 & 46.20 & 58.61 & 51.14 & 82.30 \\
\href{https://huggingface.co/sf\_model\_e5}{sf\_model\_e5} & 63.34 & 73.96 & 46.61 & 59.86 & 51.80 & 83.85 \\
\href{https://huggingface.co/GIST-Embedding-v0}{GIST-Embedding-v0} & 63.71 & 76.03 & 46.21 & 59.37 & 52.31 & 83.51 \\
\href{https://huggingface.co/gtr-t5-large}{gtr-t5-large} & 58.28 & 67.14 & 41.60 & 55.36 & 47.42 & 78.19 \\
\href{https://huggingface.co/gtr-t5-xl}{gtr-t5-xl} & 58.42 & 67.11 & 41.51 & 55.96 & 47.96 & 77.80 \\
\href{https://huggingface.co/bge-base-en-v1.5}{bge-base-en-v1.5} & 63.55 & 75.53 & 45.77 & 58.86 & 53.25 & 82.40 \\
\href{https://huggingface.co/sentence-t5-large}{sentence-t5-large} & 57.06 & 72.31 & 41.65 & 54.00 & 36.71 & 81.83 \\
\href{https://huggingface.co/gemma-7b-it}{gemma-7b-it} & N/A & N/A & N/A & N/A & N/A & N/A \\
\href{https://huggingface.co/gte-tiny}{gte-tiny} & 58.69 & 70.35 & 42.09 & 55.77 & 44.92 & 80.46 \\
\href{https://huggingface.co/gtr-t5-base}{gtr-t5-base} & 56.19 & 65.25 & 38.63 & 54.23 & 44.67 & 77.07 \\
\href{https://huggingface.co/Llama-2-7b-hf}{Llama-2-7b-hf} & N/A & N/A & N/A & N/A & N/A & N/A \\
\href{https://huggingface.co/sentence-t5-xl}{sentence-t5-xl} & 57.87 & 72.84 & 42.34 & 54.71 & 38.47 & 81.66 \\
\href{https://huggingface.co/gemma-2b-it}{gemma-2b-it} & N/A & N/A & N/A & N/A & N/A & N/A \\
\href{https://huggingface.co/e5-small}{e5-small} & 58.89 & 71.67 & 39.51 & 54.45 & 46.01 & 80.87 \\
\href{https://huggingface.co/bge-micro-v2}{bge-micro-v2} & 56.57 & 68.04 & 39.18 & 54.29 & 42.56 & 78.65 \\
\href{https://huggingface.co/all-distilroberta-v1}{all-distilroberta-v1} & N/A & N/A & N/A & N/A & N/A & N/A \\
\href{https://huggingface.co/multilingual-e5-small}{multilingual-e5-small} & 57.87 & 70.74 & 37.08 & 53.87 & 46.64 & 79.10 \\
\href{https://huggingface.co/msmarco-bert-co-condensor}{msmarco-bert-co-condensor} & 52.35 & 64.71 & 37.64 & 51.84 & 32.96 & 76.47 \\
\href{https://huggingface.co/sup-simcse-bert-base-uncased}{sup-simcse-bert-base-uncased} & 48.87 & 67.32 & 33.43 & 47.54 & 21.82 & 79.12 \\
\href{https://huggingface.co/all-MiniLM-L6-v2}{all-MiniLM-L6-v2} & 56.26 & 63.05 & 42.35 & 58.04 & 41.95 & 78.90 \\
\href{https://huggingface.co/all-mpnet-base-v2}{all-mpnet-base-v2} & 57.78 & 65.07 & 43.69 & 59.36 & 43.81 & 80.28 \\
\href{https://huggingface.co/udever-bloom-560m}{udever-bloom-560m} & 55.81 & 68.04 & 36.89 & 52.60 & 41.19 & 79.93 \\
\href{https://huggingface.co/LaBSE}{LaBSE} & 45.21 & 62.71 & 29.55 & 48.42 & 18.99 & 70.80 \\
\href{https://huggingface.co/average\_word\_embeddings\_komninos}{average\_word\_embeddings\_komninos} & 42.06 & 57.65 & 26.57 & 44.75 & 21.22 & 62.46 \\
\href{https://huggingface.co/average\_word\_embeddings\_glove.6B.300d}{average\_word\_embeddings\_glove.6B.300d} & 41.96 & 57.29 & 27.73 & 43.29 & 21.62 & 61.85 \\
\href{https://huggingface.co/allenai-specter}{allenai-specter} & 40.28 & 52.37 & 34.06 & 48.10 & 15.88 & 61.02 \\
\bottomrule
\end{tabular}
}
\end{table}

\begin{figure}
    \centering
    \includegraphics[width=\textwidth]{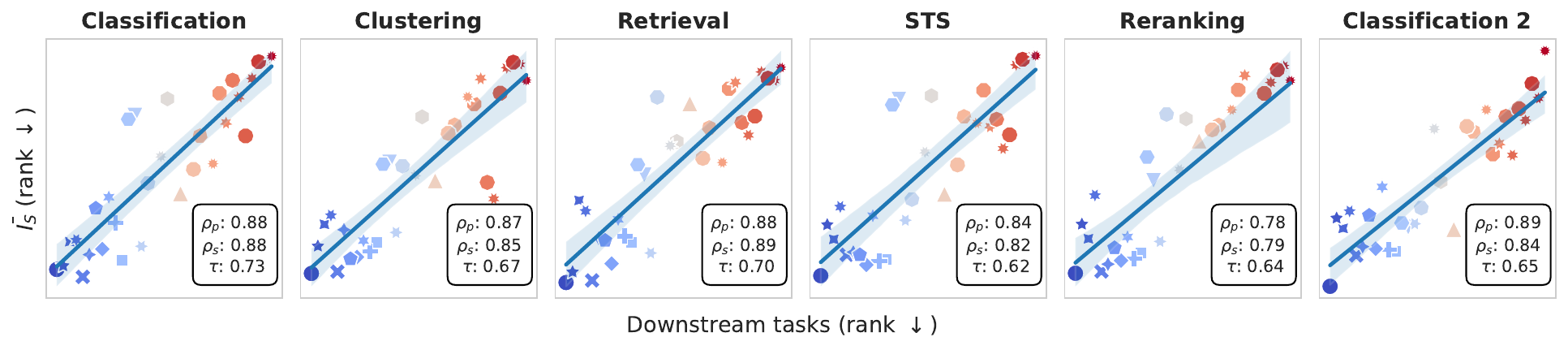}
    \caption{Correlations between rankings on different subtasks and their {\sys} score ranking.}
    \label{fig:nlp_horizontal_scatterplot_mtebs}
\end{figure}

The strength of the MTEB benchmark is that it evaluates embedders on a very large and diverse set of downstream tasks. We provide an \autoref{tab:nlp_mteb_full_results} and \autoref{fig:nlp_horizontal_scatterplot_mtebs} the full results of the MTEB benchmark (English) for the models used in the NLP experiments.

\begin{table}
    \centering
    \resizebox{0.8\textwidth}{!}{
    \begin{tabular}{llrrr}
\toprule
 &  & $\varrho_p$ & $\varrho_s$ & $\tau$ \\
Task category &  &  &  &  \\
\midrule
\multirow[c]{13}{*}{Classification} & \bfseries Average & \bfseries 0.90 & \bfseries 0.83 & \bfseries 0.68 \\
 & AmazonCounterfactualClassification (en) & 0.75 & 0.67 & 0.50 \\
 & AmazonPolarityClassification & 0.78 & 0.77 & 0.58 \\
 & AmazonReviewsClassification (en) & 0.83 & 0.82 & 0.65 \\
 & Banking77Classification & 0.92 & 0.84 & 0.65 \\
 & EmotionClassification & 0.87 & 0.76 & 0.58 \\
 & ImdbClassification & 0.77 & 0.76 & 0.57 \\
 & MassiveIntentClassification (en) & 0.93 & 0.84 & 0.68 \\
 & MassiveScenarioClassification (en) & 0.92 & 0.84 & 0.67 \\
 & MTOPDomainClassification (en) & 0.94 & 0.89 & 0.72 \\
 & MTOPIntentClassification (en) & 0.80 & 0.76 & 0.59 \\
 & ToxicConversationsClassification & 0.68 & 0.51 & 0.36 \\
 & TweetSentimentExtractionClassification & 0.70 & 0.50 & 0.35 \\
\cline{1-5}
\multirow[c]{12}{*}{Clustering} & \bfseries Average & \bfseries 0.88 & \bfseries 0.82 & \bfseries 0.63 \\
 & ArxivClusteringP2P & 0.59 & 0.64 & 0.46 \\
 & ArxivClusteringS2S & 0.66 & 0.73 & 0.54 \\
 & BiorxivClusteringP2P & 0.43 & 0.39 & 0.30 \\
 & BiorxivClusteringS2S & 0.58 & 0.64 & 0.42 \\
 & MedrxivClusteringP2P & 0.41 & 0.34 & 0.26 \\
 & MedrxivClusteringS2S & 0.58 & 0.48 & 0.34 \\
 & RedditClustering & 0.92 & 0.84 & 0.67 \\
 & RedditClusteringP2P & 0.86 & 0.88 & 0.72 \\
 & StackExchangeClustering & 0.93 & 0.88 & 0.71 \\
 & StackExchangeClusteringP2P & 0.66 & 0.63 & 0.46 \\
 & TwentyNewsgroupsClustering & 0.91 & 0.86 & 0.68 \\
\cline{1-5}
\multirow[c]{4}{*}{PairClassification} & \bfseries Average & \bfseries 0.91 & \bfseries 0.83 & \bfseries 0.67 \\
 & SprintDuplicateQuestions & 0.69 & 0.63 & 0.44 \\
 & TwitterSemEval2015 & 0.92 & 0.76 & 0.57 \\
 & TwitterURLCorpus & 0.85 & 0.81 & 0.64 \\
\cline{1-5}
\multirow[c]{5}{*}{Reranking} & \bfseries Average & \bfseries 0.85 & \bfseries 0.74 & \bfseries 0.58 \\
 & AskUbuntuDupQuestions & 0.85 & 0.65 & 0.53 \\
 & MindSmallReranking & 0.86 & 0.84 & 0.65 \\
 & SciDocsRR & 0.63 & 0.50 & 0.35 \\
 & StackOverflowDupQuestions & 0.89 & 0.77 & 0.58 \\
\cline{1-5}
\multirow[c]{16}{*}{Retrieval} & \bfseries Average & \bfseries 0.89 & \bfseries 0.87 & \bfseries 0.67 \\
 & ArguAna & 0.80 & 0.77 & 0.59 \\
 & ClimateFEVER & 0.75 & 0.78 & 0.59 \\
 & CQADupstackRetrieval & 0.85 & 0.74 & 0.59 \\
 & DBPedia & 0.94 & 0.90 & 0.74 \\
 & FEVER & 0.87 & 0.86 & 0.67 \\
 & FiQA2018 & 0.85 & 0.74 & 0.57 \\
 & HotpotQA & 0.88 & 0.87 & 0.69 \\
 & MSMARCO & 0.84 & 0.78 & 0.60 \\
 & NFCorpus & 0.91 & 0.86 & 0.68 \\
 & NQ & 0.88 & 0.78 & 0.61 \\
 & QuoraRetrieval & 0.90 & 0.80 & 0.63 \\
 & SCIDOCS & 0.79 & 0.64 & 0.49 \\
 & SciFact & 0.79 & 0.78 & 0.57 \\
 & Touche2020 & 0.71 & 0.58 & 0.41 \\
 & TRECCOVID & 0.73 & 0.65 & 0.51 \\
\cline{1-5}
\multirow[c]{11}{*}{STS} & \bfseries Average & \bfseries 0.90 & \bfseries 0.75 & \bfseries 0.57 \\
 & STSBenchmark & 0.85 & 0.70 & 0.53 \\
 & BIOSSES & 0.80 & 0.72 & 0.52 \\
 & SICK-R & 0.83 & 0.64 & 0.46 \\
 & STS12 & 0.80 & 0.63 & 0.44 \\
 & STS13 & 0.86 & 0.66 & 0.46 \\
 & STS14 & 0.89 & 0.68 & 0.47 \\
 & STS15 & 0.91 & 0.79 & 0.62 \\
 & STS16 & 0.91 & 0.76 & 0.59 \\
 & STS17 (en-en) & 0.81 & 0.45 & 0.32 \\
 & STS22 (en) & 0.84 & 0.61 & 0.45 \\
\cline{1-5}
Summarization & SummEval & 0.46 & 0.31 & 0.21 \\
\cline{1-5}
\bottomrule
\end{tabular}

    }
    \caption{Detailed correlations between the {\sys} score of the models and their performance on the MTEB benchmark.}
    \label{tab:nlp_detailed_correlations_part}
    \label{tab:nlp_mteb_full_results}
\end{table}

We obtain significant positive correlations in all categories of downstream tasks. We noticed that we obtained significantly poorer results on STS, Clustering, and Reranking tasks than on classification tasks. We believe this behavior is due to the nature of these tasks. Indeed, they do not rely on training an additional model on top of the embeddings but rather directly use the embeddings as is in dot products or similarity measures. An embedder could produce very informative embeddings, \textit{i.e.}, it is possible to extract the useful information using a small model, and at the same time, these embeddings not be adequate for dot-product-based similarity measures. We believe further investigation is needed to understand the behavior of the models on these tasks. Especially to see if training a small model on top of the embeddings can improve the performance of these tasks.

\subsubsection{Ablation studies}
\label{sec:ablation_nlp}

Many factors can impact the estimation of the {\sys} score of the models, such as the dimensions of the different embeddings and the number of available embedders to evaluate. the {\sys} score can capture many different aspects of the embeddings, such as the quality of the embeddings. We provide in this section a comprehensive set of ablation studies to evaluate the impact of these different factors on the {\sys} score of the embeddings.

\label{sec:nlp_instruction_finetuning_app}
\paragraph{Impact of instruction finetuning.} Instruction finetuning is now a common practice to improve the alignment of the base of models and expand the models' reasoning capabilities. In \autoref{fig:nlp_instruct_finetuning}, show that instruction fine-tuning positively impacts the models' performance on the downstream tasks and that the {\sys} score captures this improvement. In addition to studying the impact of instruction finetuning, we evaluated models at different checkpoints during their initial pretraining in \autoref{sec:nlp_training_steps_impact} using the CroissantLLM checkpoints~\cite{faysse2024croissantllm}.

\subsubsection{Impact of training steps}
\label{sec:nlp_training_steps_impact}

\begin{figure}
    \centering
    \includegraphics[width=0.8\textwidth]{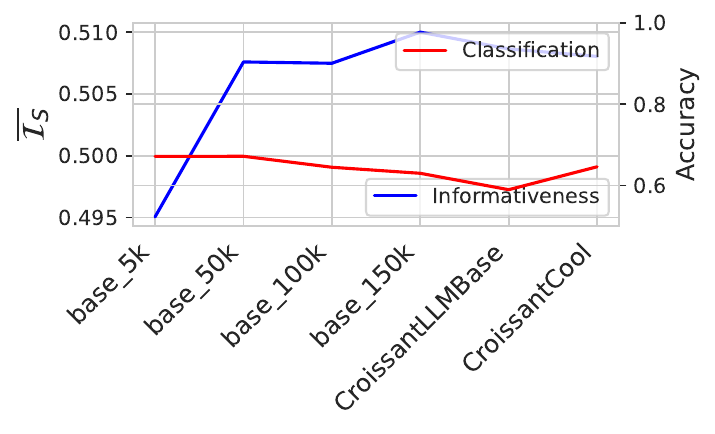}
    \caption{Impact of the number of training steps on the performance of the models on the downstream tasks and their {\sys} score. }inforatmation sufficiency
    \label{fig:steps_vs_perf_info}
\end{figure}

Surprisingly, we found that the number of training steps does not significantly impact the models' performance on the downstream tasks nor on the {\sys} score of the embeddings. The {\sys} score correctly captures this behavior as shown in \autoref{fig:steps_vs_perf_info}. We hypothesize that the {\sys} score in terms of embeddings is, in this case, determined by a few numbers of training steps (the first $5000$) and the overall architecture of the model.
Training the model further even leads to a decrease in performance on the downstream tasks, which is not captured by the {\sys} score of the embeddings; this could be due to the very small variation in the performance.

\subsubsection{Importance of embedding size normalization}
\label{sec:nlp_embedding_size_normalization}

We found that considering the amount of information packed by an embedding per coordinate is crucial to obtain a good ranking of the models. In \autoref{fig:mteb_correlation_no_norm}, we show the correlation between the performance of the models on the MTEB benchmark and their {\sys} score, not normalized by embedding size. While positive significative correlation is still present, the correlation is much weaker than when the dimension of the embeddings normalizes the information sufficiency.

\subsubsection{Community and cluster performance}

\begin{figure}[H]
   \centering
   \begin{minipage}[b]{0.5\textwidth}
       \begin{subfigure}{\textwidth}
           \centering
           \includegraphics[width=\linewidth]{fig/nlp/mteb_ds_mis_graph_community_IX_1toX_2_d_2}
           \caption{}
           \label{fig:predictive_mi_graph_app}
       \end{subfigure}
   \end{minipage}\begin{minipage}[b]{0.4\textwidth}
   {\begin{subfigure}{\textwidth}
        \centering
        \includegraphics[width=\textwidth]{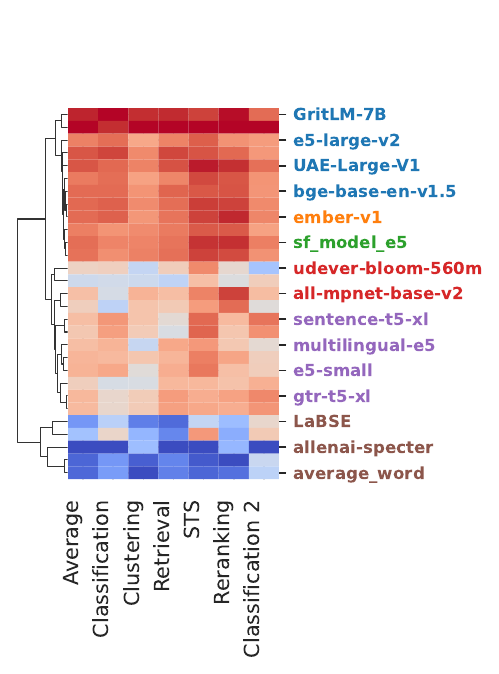}
        \caption{}
        \label{fig:nlp_community_perfs}
   \end{subfigure}
   }
   \end{minipage}
   \caption{We present different interesting properties of {\sys}. In \autoref{fig:predictive_mi_graph}, we show that it can be used to cluster models \autoref{fig:nlp_community_perfs}, reports the performance of the models on the different task categories. They are grouped by similar behaviors on these tasks (dendrograms) and colored by the communities discovered in the information sufficiency graph. (Only models evaluated as part of the MTEB benchmark are shown).}
   \label{fig:rankings_informativeness_classification}
\end{figure}

\begin{figure}
    \centering
    \includegraphics[width=0.8\textwidth]{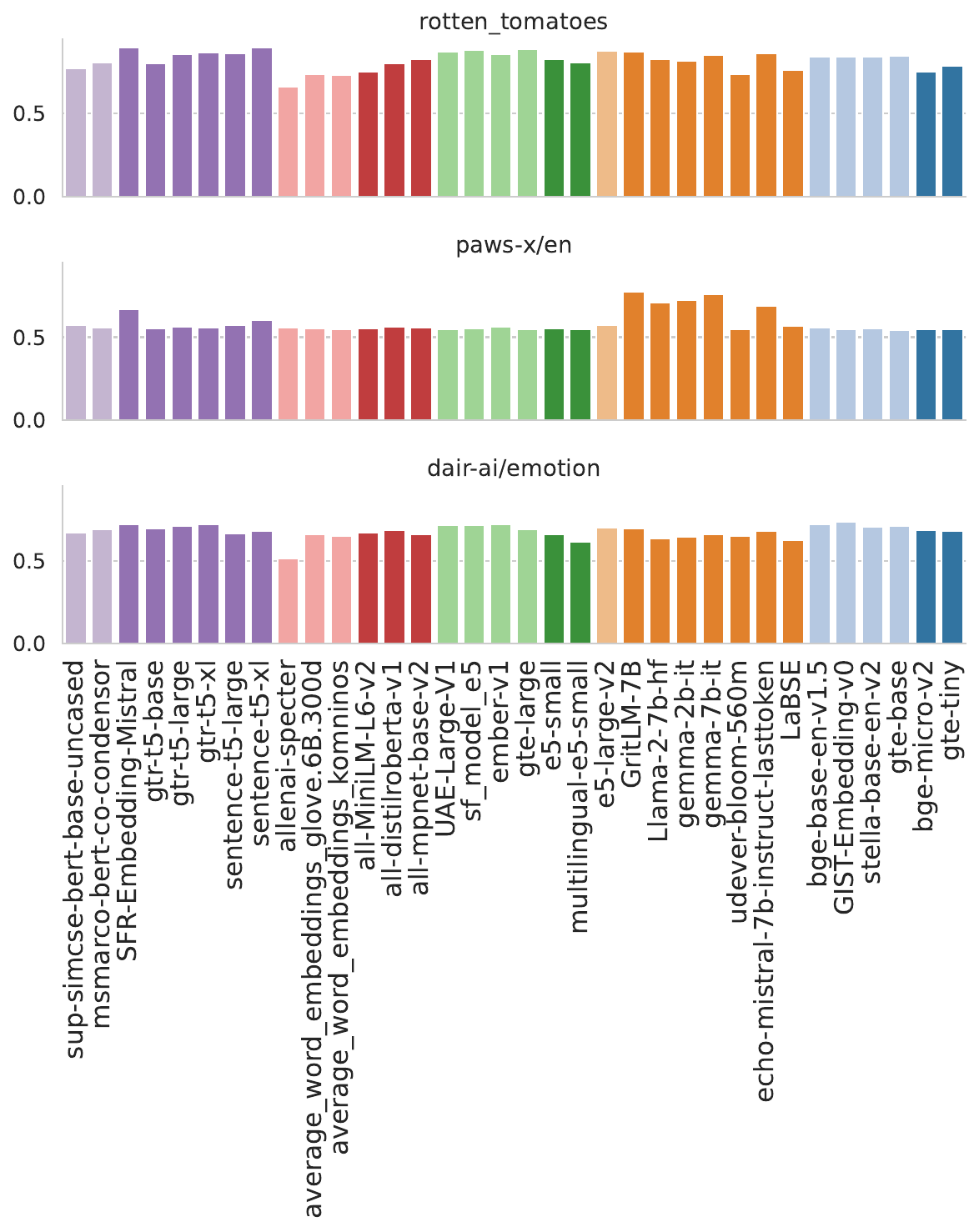}
    \caption{Performance of the models on the downstream tasks grouped by clusters discovered by the directed {\sys}.}
    \label{fig:nlp_community_perfs_barplot_app}
\end{figure}

\label{sec:nlp_community_performance}
\label{subsec:nlp-clustcomm}
We postulate that models clustered together by information sufficiency are likely to behave similarly on the downstream tasks. We evaluate this hypothesis by grouping the models by clusters discovered using the information sufficiency and reporting their performance on the downstream tasks. In \autoref{fig:nlp_community_perfs} and \autoref{fig:nlp_community_perfs_barplot_app}, we observe that models within the same cluster tend to have similar behaviors on the downstream tasks.

\subsubsection{Evaluating information sufficiency on different datasets}

\begin{figure}
    \centering
    \includegraphics[width=\textwidth]{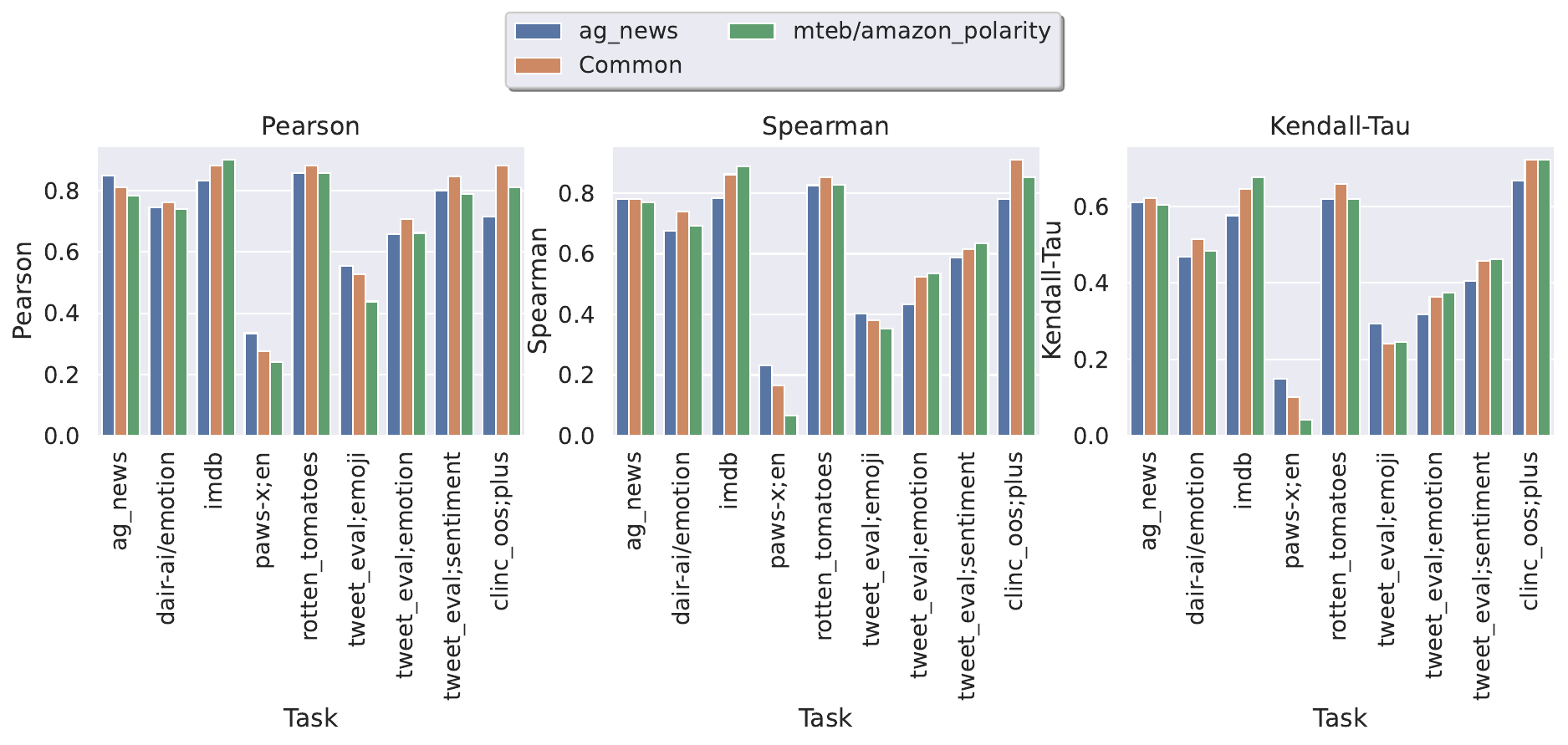}
    \caption{Correlation between {\sys} scores computed on different datasets and the cross-performance on different tasks.}
    \label{fig:emir_dataset_impact}
\end{figure}

The {\sys} score is evaluated with a fixed dataset supposed to represent the data distribution of interest (either a very diverse set or a subset following a distribution specific to a subfield like medical or legal texts). We cross-evaluated the {\sys} of the models on different datasets and the performance of the models on the downstream tasks in \autoref{fig:emir_dataset_impact}. We find that closer datasets in terms of the data distribution lead to a higher correlation between the {\sys} score of the models and their performance on the downstream tasks. It is especially highlighted when comparing the correlations we get when evaluating {\sys} on the AG News and Amazon polarity datasets. The first one corresponds to news articles, and the task is to guess the topic, whereas Amazon Polarity corresponds to product reviews, which is a sentiment analysis task. We find that the {\sys} score evaluated on Amazon Polarity tends to yield way better correlation with the performance on the sentiment analysis downstream tasks such as tweet\_eval/sentiment, tweet\_eval/emotion, IMDB or Rotten Tomatoes or to a lesser extent dair/emotion. Interestingly, the difference is less significant on the tweet\_eval/emoji subtask.

\section{Molecular Experiment Details}
\label{sec:molecular_experiment_details}

\subsection{Embedders considered}
\label{sec:models_mol}

\begin{table}[H]
    \caption{Models evaluated on the ZINC dataset.}
    \centering
    \resizebox{\textwidth}{!}{
        \begin{tabular}{c| c | c |  c | c c c}
            \toprule
            Model name & \small SMILES & \small 2D-GNN &  \small 3D-GNN & Architecture & Out size & Dataset (size)\\
            \midrule
            Not-trained & & \checkmark & & \small GIN  &  300 &  - \\
            \midrule
            AttributeMask\cite{sun2022does} &  & \checkmark & & \small GIN  & 300 & GEOM~\cite{axelrod2022geom} \small (50k)\\
            ContextPred\cite{sun2022does} &  & \checkmark & & \small GIN  & 300 & GEOM~\cite{axelrod2022geom} \small (50k)\\
            GPT-GNN\cite{gptgnn} &  &  \checkmark & & \small GIN  & 300 & GEOM~\cite{axelrod2022geom} \small (50k)\\
            InfoGraph\cite{sun2019infograph} & & \checkmark & & \small GIN  & 300 & GEOM~\cite{axelrod2022geom} \small (50k)\\
            GraphCL\cite{You2020GraphCL} & & \checkmark & & \small GIN  & 300 & GEOM~\cite{axelrod2022geom} \small (50k)\\
            GROVER\cite{rong2020self} & & \checkmark & & \small GIN  & 300 & GEOM~\cite{axelrod2022geom} \small (50k)\\
            GraphLog\cite{xu2021selfsupervised} & & \checkmark & & \small GIN  & 300 & GEOM~\cite{axelrod2022geom} \small (50k)\\
            GraphMVP\cite{liu2022pretraining}\footnotemark[1] & & \checkmark & & \small GIN  & 300 & GEOM~\cite{axelrod2022geom} \small (50k)\\
            3D-infomax\cite{stark2021_3dinfomax}\footnotemark[1] & & \checkmark & & \small PNA & 800 & QMugs~\cite{isert2021qmugs} \small (620k)\\
            MolR\cite{wang2022chemicalreactionaware} & & \checkmark & & \small GCN, GAT, TAG & 1024 & USPTO\cite{wang2022chemicalreactionaware} \small (~1.5M)\\
            MoleOOD\cite{yang2022learning} & & \checkmark & & \small GIN , SAGE, GCN & 256 & BACE~\cite{hu2020ogb} \small (400k)\\
            \midrule
            ChemBERT \small MLM\cite{ahmad2022chemberta2} & \checkmark &  & & \small RoBERTa & 600 & PubChem~\cite{pubchem} \small (5M, 10M, 77M)\\
            ChemBERT \small MTR\cite{ahmad2022chemberta2}  & \checkmark &  & & \small RoBERTa & 384 & PubChem~\cite{pubchem} \small (5M, 10M, 77M)\\
            MolBert\cite{fabian2020molecular} & \checkmark &  & & \small BERT & 256 & GuacaMol~\cite{Brown_2019} \small (1.6M)\\
            ChemGPT\cite{frey_neural_2023} & \checkmark &  & & \small GPT & 128, 256, 2048 &  PubChem~\cite{pubchem} \small (10M)\\
            \midrule
            3D-denosing\cite{zaidi2023pretraining} & & & \checkmark & TorchMD-net & 256 & PCQM4Mv2\cite{hu2021ogblsc} \small (3.7M)\\
            3D-fractional\cite{pmlr-v202-feng23c} & & & \checkmark & TorchMD-net & 256 & PCQM4Mv2\cite{hu2021ogblsc} \small (3.7M)\\
            \bottomrule
        \end{tabular}
    }
    \vspace{0.2cm}
    \label{tab:molmodels}
\end{table}

We considered $28$ models for the molecular experiments, summed up in~\autoref{tab:molmodels}.
Some models were used in different versions (architectures, number of parameters, pretraining dataset's size), such as the ChemBert models, followed by the size of their datasets, or ChemGPT, followed by their number of parameters.

Most 2D-GNNs were trained on the GEOM~\cite{axelrod2022geom} dataset and were gathered from the repository of GraphMVP~\cite{liu2022pretraining} model. 
Note that the MoleOOD~\cite{yang2022learning} model was trained on the BACE~\cite{hu2020ogb} dataset, with a supervised task specific to the $\beta$-secretase enzyme.
As a result, this model can be seen as "already specialized", explaining its poor performance in our evaluation.

We used the RD-Kit and Datamol tool-kits\cite{landrum_rdkitrdkit_2020,mary_datamol-iodatamol_2024} to pre-process the molecules and to generate three-dimensional conformers for 3D-models.
To run the models using 3D views of the molecules, we generated five conformers (possible 3D configuration of the molecule) for each SMILES and kept the conformer with the lowest energy.
Note that this methodology is imperfect, as the 3D coordinates might be noisy; however, we followed the same procedure to pre-process the ZINC dataset, to evaluate the information sufficiency, and on the datasets corresponding to each downstream task.

Finally we considered a variety of models architecture for 2D-GNNs notably graph isomorphism network (GIN)~\cite{xu2018how}, principal neighbor aggregation networks (PNA)~\cite{corso2020pna}, graph convolutional network (GCN)~\cite{duvenaud2015convolutional}, graph attention network (GAT)~\cite{veličković2018graph}, topology adaptive graph convolutional networks (TAG)~\cite{du2018topology} and GraphSAGE~\cite{hamilton2018inductive}.
For SMILES-based models, backbones are inspired by BERT~\cite{devlin2019bert}, RoBERTa~\cite{liu2019roberta}, and GPT~\cite{frey_neural_2023}.
Finally, both our 3D models use TorchMD-net\cite{pelaez2024torchmdnet} as a backbone.

\subsection{Details on the information sufficiency estimation}
\label{subsec:pmiestdet}

\begin{figure}
    \begin{subfigure}{0.54\textwidth}
        \centering
        \includegraphics[width=\linewidth]{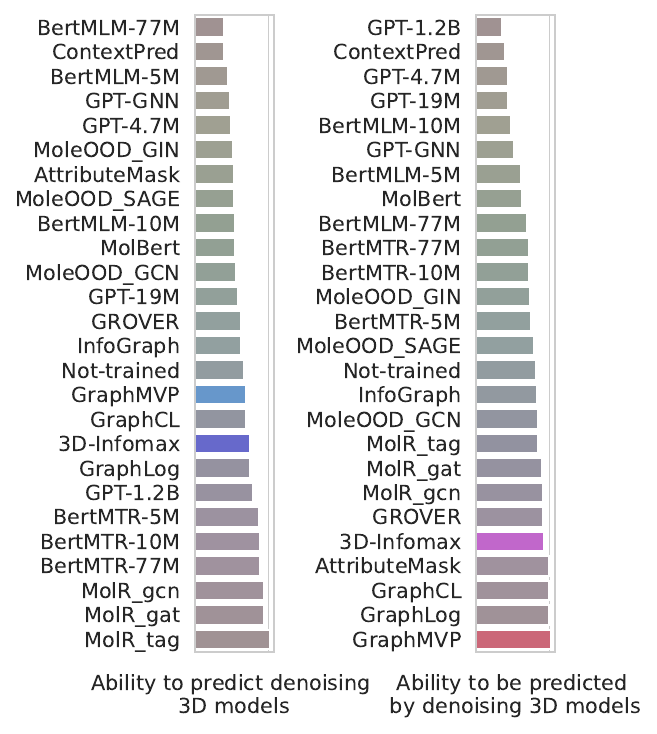}
        \caption{}
        \label{fig:denoising_3D_MI}
    \end{subfigure}\hfill
    \begin{subfigure}{0.45\textwidth}
        \centering
        \includegraphics[width=\linewidth]{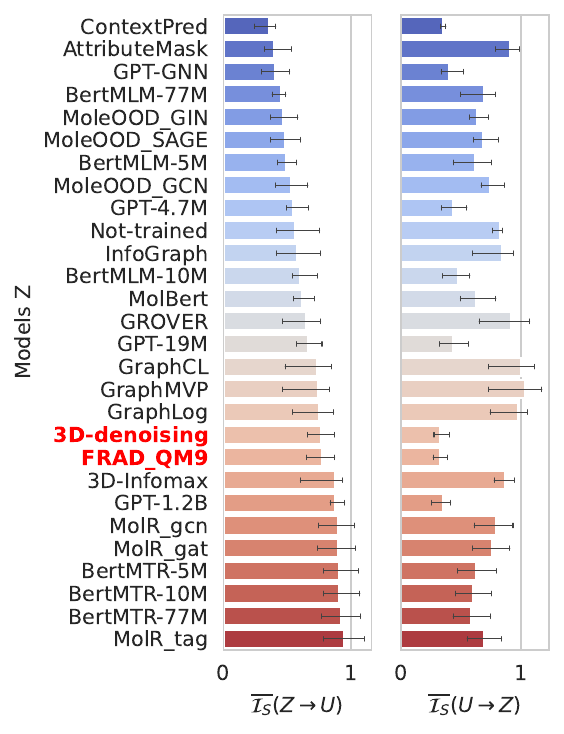}
        \caption{}
        \label{fig:barplot_MI3d}
    \end{subfigure}\hfill
  \caption{
      (a) Information sufficiency of embedders over 3D-Denoising models (left) and of 3D-Denoising models over the other embedders (right).
      (b) Information sufficiency of embedders in both directions. We see the 3D denoising models are among the least predicted models.
   }
  \label{fig:fig_annx_MImol}
\end{figure}

\begin{figure}
    \centering
    \includegraphics[width=.8\linewidth]{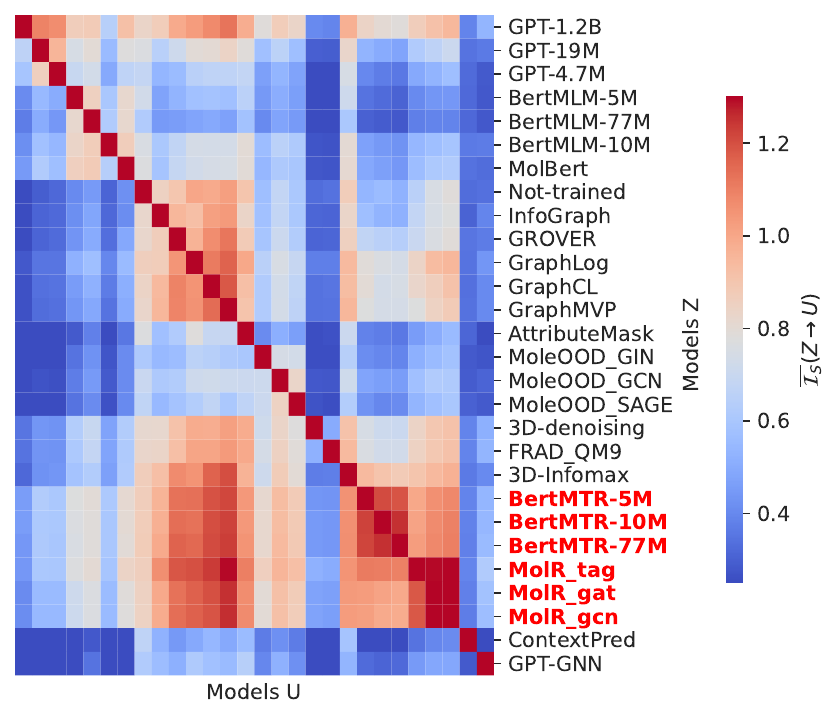}
    \caption{
        Pairwise information sufficiency between molecular embedders.
    }
    \label{fig:clustermap_MI}
\end{figure}

\textbf{3D models.}
The two 3D models considered (FRAD~\cite{pmlr-v202-feng23c} and Denoising~\cite{zaidi2023pretraining}) obtain high {\sys} scores while being among the least predictable (\autoref{fig:barplot_MI3d} and \autoref{fig:clustermap_MI}).
This suggests that these models capture 3D-specific features inaccessible from other modalities while maintaining sufficient overlap to predict them.

\noindent\textbf{2D-3D models} Some 2D-GNNs we considered (GraphMVP and 3D-infomax) are trained to maximize the mutual information between their embeddings and 3D representations of the molecule.
Hence, we expect these models to be related to the 3D-denoising models we considered. 
However, we observe in~\autoref{fig:denoising_3D_MI} that these models do not achieve particularly high information sufficiency scores over 3D-denoising models.
On the other hand, the 3D models achieve high information sufficiency scores over them, which might suggest that these 2D models and 3D-denoising models share information that is easier to access from the 3D models.
However, we want to point out that GraphMVP and 3D-infomax are both among the most predicted models; that is to say, among the other models in our pool, they achieve the highest information sufficiency scores.

\subsection{Complementary results on ADMET tasks}
\label{sec:complementary_resultsADMET}

\definecolor{r2}{rgb}{0.1, 0.0, 0.5}
\definecolor{aur}{rgb}{0.1, 0.3, 0.0}

\begin{table}
    \caption{
        ADMET tasks extracted from the Therapeutic Data Commons platform~\cite{Huang2021tdc} considered in our experiments.
        We report the correlation between the informativeness score and the performances of the embedders on the downstream tasks in terms of Pearson correlation $\varrho_p$, Spearman correlation $\varrho_s$ and Kendall-Tau $\tau$.
        We also report the average metric of the models on each task across the grid search runs, in terms of \textbf{\textcolor{r2}{\boldmath$R^2$ for regression tasks}} and \textbf{\textcolor{aur}{AUROC for classification tasks}}.
        The tasks are ordered within each category by the correlation with the informativness score (in terms of Spearman correlation).
    }
    \label{tab:molecul_tasks}
    \centering
    \resizebox{\textwidth}{!}{
        \begin{tabular}{c c c | c c | c c c | c}
            \toprule
            Category & Model & Task & cls & reg & \multicolumn{3}{c|}{Correlation} & Avg. metric\\
            &name&size & & & $\varrho_p$ & $\varrho_s$ & $\tau$ & in the grid\\
            \midrule
            \multirow{8}{*}{Absorption}
            & P-glycoprotein Inhibition & 1212 & \checkmark & & 0.92 & 0.93 & 0.76 & \textcolor{aur}{$0.88 \pm 0.03$} \\
            & AqSolDB & 9982 & & \checkmark & 0.91 & 0.91 & 0.75 & \textcolor{r2}{$0.52 \pm 0.09$} \\
            & Lipophilicity & 4200 & & \checkmark & 0.88 & 0.89 & 0.71 & \textcolor{r2}{$0.29 \pm 0.07$} \\
             & Caco-2 Permeability & 906 & & \checkmark & 0.77 & 0.80 & 0.61 & \textcolor{r2}{$0.32 \pm 0.08$} \\
            & Human Intestinal Absorption & 578 & \checkmark & &  0.79 & 0.77 & 0.54 &  \textcolor{aur}{$0.67 \pm 0.02$} \\
            & FreeSolv & 642 & & \checkmark & 0.65 & 0.73 & 0.53 & \textcolor{r2}{$0.36 \pm 0.12$} \\
            & PAMPA Permeability & 2035 & \checkmark & & 0.61 & 0.63 & 0.44 & \textcolor{aur}{$0.67 \pm 0.02$} \\
            & Oral Bioavailability & 640 & \checkmark & & 0.50 & 0.45 & 0.33 & \textcolor{aur}{$0.68 \pm 0.02$} \\

            \midrule

            \multirow{3}{*}{Distribution}
            & Plasma-Protein BDR & 1614 &  & \checkmark & 0.85 & 0.84 & 0.68 & \textcolor{r2}{$0.18 \pm 0.04$} \\
            & Blood-Brain barrier & 1975 & \checkmark & & 0.79 & 0.81 & 0.60 & \textcolor{aur}{$0.32 \pm 0.08$} \\
            & VDss & 1130 & & \checkmark & 0.71 & 0.73 & 0.53 & \textcolor{r2}{$0.16 \pm 0.05$} \\

            \midrule

            \multirow{8}{*}{Metabolism}
            & CYPP450 3A4 Inhib. & 12328 & \checkmark & & 0.96 & 0.96 & 0.85 & \textcolor{aur}{$0.80 \pm 0.04$} \\
            & CYPP450 1A2 Inhib. & 12579 & \checkmark & & 0.94 & 0.95 & 0.81 & \textcolor{aur}{$0.87 \pm 0.03$} \\
            & CYPP450 2C19 Inhib. & 12665 & \checkmark & & 0.94 & 0.95 & 0.75 & \textcolor{aur}{$0.82 \pm 0.03$} \\
            & CYPP450 2C9 Inhib. & 12092 & \checkmark & & 0.92 & 0.92 & 0.79 & \textcolor{aur}{$0.83 \pm 0.03$} \\
            & CYPP450 2D6 Inhib. & 13130 & \checkmark & & 0.94 & 0.91 & 0.74 & \textcolor{aur}{$0.78 \pm 0.04$} \\
            & CYPP450 2D6 Substrate & 664 & \checkmark & & 0.75 & 0.74 & 0.57 & \textcolor{aur}{$0.75 \pm 0.03$} \\
            & CYPP450 3A4 Substrate & 667 & \checkmark & & 0.50 & 0.53 & 0.35 & \textcolor{aur}{$0.63 \pm 0.02$} \\
            & CYPP450 2C9 Substrate & 666 & \checkmark & & 0.20 & 0.13 & 0.09 & \textcolor{aur}{$0.65 \pm 0.02$} \\

            \midrule

            \multirow{3}{*}{Excretion}
            & Clearance hepatocyte & 1020 & & \checkmark & 0.78 & 0.79 & 0.57 & \textcolor{r2}{$0.11 \pm 0.03$} \\
            & Half Life & 667 & & \checkmark & 0.76 & 0.78 & 0.58 & \textcolor{r2}{$-0.06 \pm 0.11$} \\
            & Clearance microsome & 1102 & & \checkmark & 0.72 & 0.72 & 0.54 & \textcolor{r2}{$0.08 \pm 0.03$} \\

            \midrule

            \multirow{9}{*}{Toxicity}
                    & Tox21 & 7831 & \checkmark & & 0.93 & 0.93 & 0.78 & \textcolor{aur}{$0.75 \pm 0.03$} \\
            & \multirow{2}{*}{hERG} & 13445 & \checkmark & & 0.91 & 0.90 & 0.75 & \textcolor{aur}{$0.76 \pm 0.04$} \\
            &                        & 648 & \checkmark & & 0.81 & 0.84 & 0.63 & \textcolor{aur}{$0.75 \pm 0.03$} \\
            & Acute Toxicity LD50 & 7385 & & \checkmark & 0.82 & 0.82 & 0.63 & \textcolor{r2}{$0.16 \pm 0.04$} \\
            & Ames Mutagenicity & 7255 & \checkmark & & 0.79 & 0.78 & 0.60 & \textcolor{aur}{$0.79 \pm 0.03$} \\
            & ClinTox & 1484 & \checkmark & & 0.69 & 0.69 & 0.49 & \textcolor{aur}{$0.71 \pm 0.03$} \\
            & Carcinogens & 278 & \checkmark & & 0.47 & 0.49 & 0.35 & \textcolor{aur}{$0.76 \pm 0.08$} \\
            & Drug Induced Liver Injury & 475 & \checkmark & & 0.39 & 0.36 & 0.25 & \textcolor{aur}{$0.83 \pm 0.03$} \\
            & Skin Reaction & 404 & \checkmark & & 0.01 & 0.07 & 0.06 & \textcolor{aur}{$0.64 \pm 0.03$} \\

            \bottomrule

        \end{tabular}
    }
    \vspace{0.5cm}

    \label{tab:admet}
\end{table}

The datasets chosen for the molecular experiments are extracted from the Therapeutics Data Commons (TDC)~\cite{Huang2021tdc} platform.
We focused our experiments on ADMET tasks, crucial for drug discovery and development, which results in a total of $31$ tasks, described in~\autoref{tab:admet}.

\begin{wraptable}{r}{0.6\textwidth}
    \centering
    \resizebox{0.6\textwidth}{!}{
        \begin{tabular}{c c c c}
        \toprule
        Dropout rate & hidden dimension & network depth & n-epochs\\
        \midrule
        $0$, $0.2$ & $4$, $8$, $16$, $32$, $64$, $128$ & $1$, $2$, $3$ & min$(200,200 * \frac{5000}{\text{task size}})$\\
        \bottomrule
        \label{tab:grid}
        \end{tabular}
    }
    \caption{Hyperparameters tuned for the evaluation of embedders on ADMET downstream tasks}
\end{wraptable}

Each dataset is split into a training set, a validation set, and a test set following a scaffold split.
This splitting strategy ensures that molecules sharing a similar scaffold will be part of the same split in the task. This corresponds to a more realistic scenario, where practitioners only have access to molecules belonging to the same chemical series.
Classifiers are then trained on the training set of each task, where the best hyperparameters and checkpoints are selected on the validation set.
The final performance is finally measured on the test set, and we run each experiment $10$ times with different random seeds.

A grid search is performed on each dataset individually to maximize the average AUROC or $R^2$ score across all models for binary classification and regression.
We chose a maximum number of epochs depending on the task size to ensure all models have time to converge, limiting this amount to grow to at most $200$ epochs.

\autoref{tab:admet} also displays the variation of the correlation coefficient between the ranking obtained on the {\sys} score and the performances obtained on the downstream tasks regarding Spearman and Kendall correlations.
We can see that the {\sys} score correlates well with the performance of downstream tasks when the amount of data available is large.

\begin{figure}
    \centering
    \includegraphics[width=.9\linewidth]{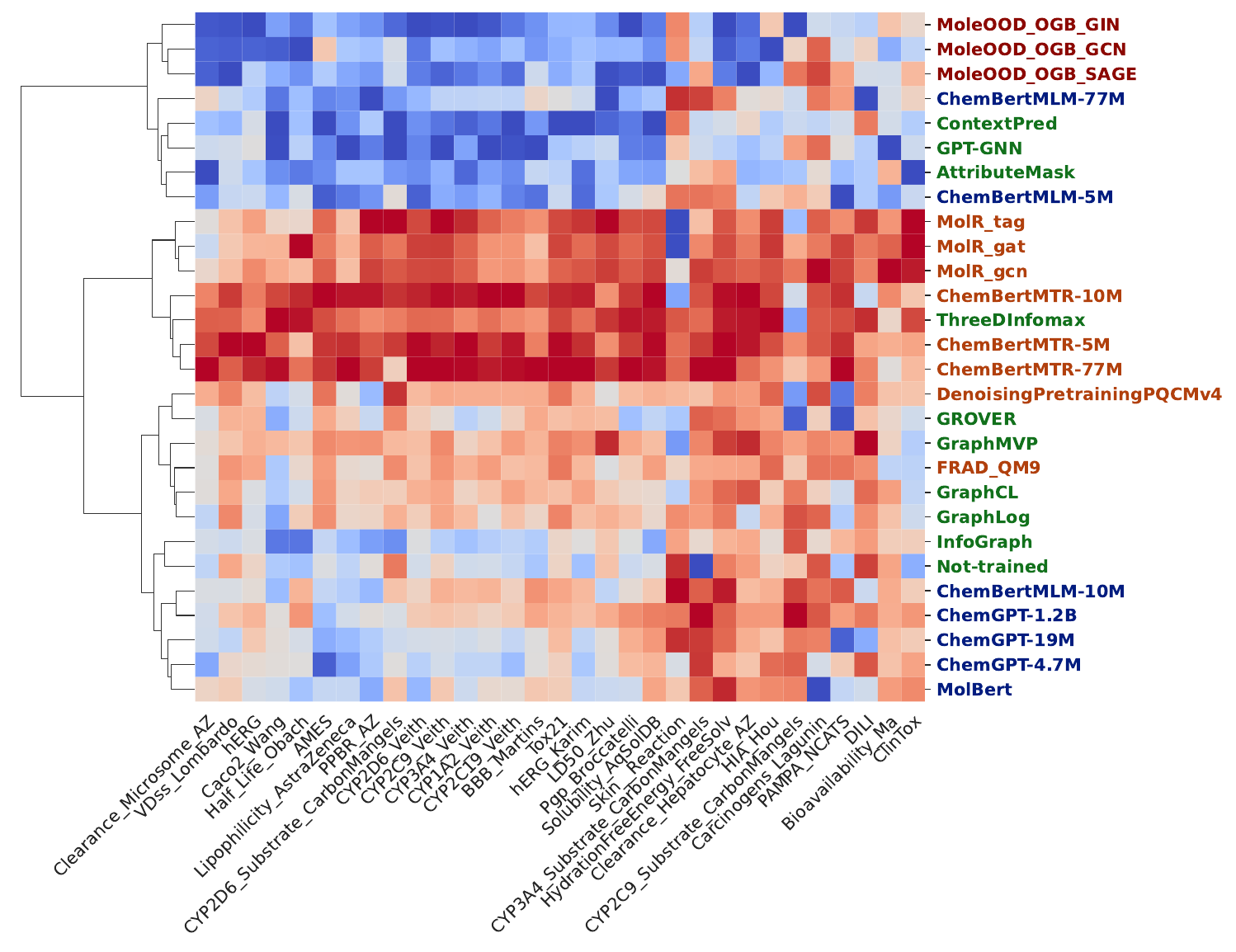}
    \caption{
        Heatmap representation of the performances of the different models on the downstream tasks, where embedders behaving similarly on the various tasks are clustered, and the embedders are colored based on their community computed in~\autoref{subsec:drugdiscovery-inf} based on the {\sys} score.
    }
    \label{fig:communities_task_ward}
\end{figure}

Finally, we can see in~\autoref{fig:communities_task_ward} that by grouping models based on their performances on these tasks, we obtain a similar clustering to the one obtained on the {\sys} score in~\autoref{subsec:drugdiscovery-inf}, with NLP-inspired models grouped. Similarly, the \textit{tiny{Chem}}Bert-MTR and MolR models are also grouped.

\subsection{Drug target Interaction prediction}
\label{sec:DTI}

\begin{figure}[b]
  \centering
  \includegraphics[width=.7\linewidth]{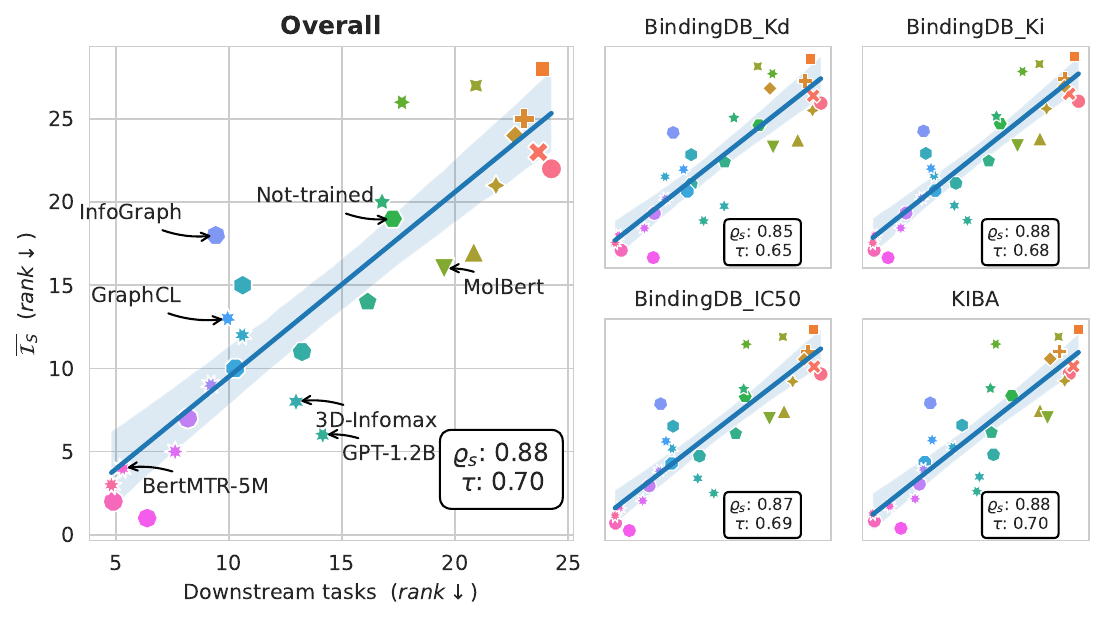}
  \caption{
      Correlations of the {\sys} score with the performances on the DTI tasks, in terms of Spearman and Kendall coefficients.
  }
  \label{fig:mearnak_detailed_global_DTI_KIBA}
\end{figure}

We propose further evaluating the embedders on yet another type of downstream task: Drug-Target Interaction.
This task aims to predict the binding affinity between a given pair (drug, target).
Since none of our models can process protein sequences, we decompose each dataset into multiple regression tasks on a single target by querying all molecules associated with a label for this target.
Each task is then formulated as a set of molecules: $\mathcal{X} = \{x_i\}_{i\in \{0,\ldots, N\}}$, and their labels $\mathcal{Y} = \{y_i\}_{i\in \{0,\ldots, N\}}$

However, such tasks can be small, making it hard to build proper models from the embeddings. In contrast, the number of tasks is very large, making it computationally expensive to proceed to an adequate hyperparameter selection.
To bypass these limitations, we propose to estimate the embedded space's clustering quality for each model by measuring how close the labels of a molecule are compared to its nearest neighbors for each task.
In other words, we measure:
\begin{equation}
    \begin{aligned}
        \Tilde{\mathcal{L}}_{n_{\textrm{neigh}}}(\mathcal{X},\mathcal{Y}) &=\frac{1}{N} \sum_{i=0}^{N} \mathcal{L}_{n_{\textrm{neigh}}}(i, \mathcal{X}, \mathcal{Y}),\\
        \text{with} \quad\mathcal{L}_{n_{\textrm{neigh}}}(i, \mathcal{X}, \mathcal{Y}) &=\frac{1}{n_{\textrm{neigh}}} \sum_{j\in \mathcal{N}_{i ,n_{\textrm{neigh}}}(\mathcal{X})} \lVert y_i - y_j \rVert^2,
    \end{aligned}
\end{equation}
and $\mathcal{N}_{i,n_{\textrm{neigh}}}(\mathcal{X})$ is the set containing the $n_{\textrm{neigh}}$ closest neighbors of $x_i\in\mathcal{X}$, which would be the performances of a K-nearest neighbors regressor on the task when using one data sample.

This quantity can be interpreted as a proxy of the embedding's capability to perform a similarity search, a classic chemo-informatic method using the similarity between different molecular projections to perform predictions.
This training-free and computationally inexpensive approach allows us to evaluate the models on many tasks/targets.

We focused on 4 DTI datasets: KIBA~\cite{kiba}, BindingDB-Kd, BindingDB-Ki, and BindingDB-IC50~\cite{bindingdb}, with a total of $1496$ tasks.
We removed all tasks containing less than 128 molecules to ensure minimum data for the clustering evaluation.

\begin{table}[H]
    \centering
    \resizebox{\textwidth}{!}{
        \begin{tabular}{l | cccc | cccc | cccc | cccc}
        \toprule
        &\multicolumn{4}{c }{KIBA (175 targets)} & \multicolumn{4}{c }{BindingDB-Kd (22 targets)} & \multicolumn{4}{c }{BindingDB-Ki (372 targets)} & \multicolumn{4}{c }{BindingDB-IC50 (927 targets)} \\
        \midrule
        $n_{neighb}$ & 1 & 2 & 4 & 8 & 1 & 2 & 4 & 8 & 1 & 2 & 4 & 8 & 1 & 2 & 4 & 8 \\
        \midrule
        $\varrho_p$ & 0.85 & 0.84 & 0.84 & 0.83 & 0.90 & 0.90 & 0.87 & 0.84 & 0.88 & 0.88 & 0.88 & 0.87 & 0.87 & 0.86 & 0.86 & 0.86\\
        $\varrho_s$ & 0.88 & 0.88 & 0.87 & 0.86 & 0.87 & 0.86 & 0.85 & 0.82 & 0.87 & 0.87 & 0.87 & 0.88 & 0.86 & 0.87 & 0.87 & 0.86\\
        $\tau$ & 0.72 & 0.71 & 0.69 & 0.67 & 0.68 & 0.66 & 0.65 & 0.62 & 0.68 & 0.67 & 0.69 & 0.69 & 0.66 & 0.68 & 0.69 & 0.68\\

        \bottomrule
        \end{tabular}   
    }
    \vspace{0.2cm}
    \caption{
        Correlation between \sys's informativness score and our clustering evaluation score $\Tilde{\mathcal{L}}_{n_{\textrm{neighb}}}$ on the four DTI datasets considered.
    }
    \label{tab:dtitab}
\end{table}

We obtain similar results as in~\autoref{subsec:drugdiscovery-exp}, our metric correlating with the performances on the different tasks considered.
\autoref{fig:mearnak_detailed_global_DTI_KIBA} sums up all results by establishing a ranking across models and the number of neighbors, where we can see that MolR and ChemBerta-MTR appear as both the most promising models according to their {\sys} score, and the best models evaluated.
Furthermore, the outliers observed in\autoref{subsec:drugdiscovery-exp} show different behaviors in this setting.
For instance, while 3D-Infomax seemed under-estimated and InfoGraph over-estimated by the {\sys} score after seeing the results on the ADMET tasks, Infograph appears under-estimated in this setup, and 3D-infomax over-estimated.

\section{Information Sufficiency Estimation}
\label{sec:emir_estimation}

\subsection{Estimation method}
\label{sec:is-actual-estim}
As stated in \autoref{sec:theory}, the deficiency $\delta(P_{U|X} \rightarrow P_{Z|X})$ is an intractable object measuring the cost of the reconstruction of $Z$ from $U$.
Due to this intractability, we propose to estimate the information sufficiency $\mathcal{I}_{S}(U \rightarrow Z)$, which is a tractable proxy for the deficiency.

\subsubsection{KNIFE Estimator}
\label{sec:knife-estimator}

We recall the definition of the information sufficiency:
\begin{equation*}
    \mathcal{I}_{S}(U \rightarrow Z )  \triangleq
    \underbrace{\inf_{f\in \Frond_\Theta(\mathsf{Z})} \mathbb{E}\left[ - \log f(Z) \right]}_{\text{Uncertainty of } Z} - \underbrace{\mathbb{E}\left[ \inf_{M \in \mathcal{K}_\Theta(\mathsf{Z} |\mathsf{U} ) } \mathbb{E}\left[ - \log M(Z|U) | U \right]  \right ]}_{\text{Uncertainty when simulating } Z \text{ from } U \text{ with } M}.
\end{equation*}

We denote $C$ the number of modes chosen for the Gaussian mixture distributions and $\operatorname{GM}_{\bm{\mu, \Sigma, w}}$ the Gaussian mixture distribution with $C$ components, parametrized by $\bm{\mu, \Sigma, w}$, with $\mathbf{1}^T \bm{w}=1$, such that

$\operatorname{GM}_{\bm{\mu, \Sigma, w}}(z) = \sum_{c=1}^C \bm{w}_c \mathcal{N}(z | \bm{\mu}_c, \bm{\Sigma_}c)$, where $ \mathbf{w}_c$ is the weight of the $c$-th component, and $\mathcal{N}(z | \bm{\mu}_c, \bm{\Sigma}_c)$ is the density of a multivariate Gaussian distribution with mean $\bm{\mu}_c$ and covariance $\bm{\Sigma}_c$ the c-th mean and covariance matrix.

To estimate the information sufficiency between the two embedders $U$ and $Z$, we follow the procedure described in KNIFE~\cite{pichler2022differential}.

$\Frond_\Theta$ is hence the class of multivariate Gaussian mixtures with $C$ components, it is parametrized by $\bm{\mu, \Sigma, w}$. 
These learnable parameters are optimized to maximize the log-likelihood of $Z$.\footnote{In practice since this operation does not depend on $U$, we store the weights of this distribution, and use it for all pairs $\mathcal{I}_{S}(. \rightarrow Z )$.}

The class of Markov kernels $\mathcal{K}_\Theta(\mathsf{Z} |\mathsf{U} )$ is also composed of multivariate Gaussian mixtures whose parameters are estimated using a small feedforward neural network. Such that for each $u \in \mathsf{U}$, the parameters of the Gaussian mixture are $\bm{\mu}(u), \bm{\Sigma}(u), \bm{w}(u)$.

In practice we considered the covariance matrix to be diagonal to avoid the number of parameters of the Gaussian mixtures to grow too large with the dimension of the embeddings $d$.
The number of parameters to be estimated for the Gaussian mixtures are hence: $C \times (2d + 1)$, with the number of parameters of the feedforward networks for the Markov kernels.
\begin{algorithm}
    \caption{Estimation of $\mathcal{I}_S(U \rightarrow Z)$, $\operatorname{GM}_{\mathbf{\mu}, \mathbf{\Sigma}, \mathbf{w}}$ denotes the Gaussian Mixture model with means $\mathbf{\mu}$, covariances $\mathbf{\Sigma}$ and weights $\mathbf{w}$.}
    \label{algo:is_estimation}
    \begin{algorithmic}
        \REQUIRE {Pairs of corresponding embeddings $(z_i, u_i)_N$}
        \ENSURE {Information sufficiency $\mathcal{I}_S(U \rightarrow Z)$}

        \STATE $\mathbf{\mu_Z}, \mathbf{\Sigma_Z}, \mathbf{w_Z} \gets \arg \min_{\mathbf{\mu}, \mathbf{\Sigma}, \mathbf{w}} - \sum_{i=1}^N \log \operatorname{GM}_{\mathbf{\mu}, \mathbf{\Sigma}, \mathbf{w}}(z_i)$
        \STATE $\mathbf{\mu_{Z|U}}, \mathbf{\Sigma_{Z|U}}, \mathbf{w_{Z|U}}  \gets \arg \min_{\mathbf{\mu}, \mathbf{\Sigma}, \mathbf{w}}- \sum_{i=1}^N\ log \operatorname{GM}_{\mathbf{\mu}(u_i), \mathbf{\Sigma}(u_i), \mathbf{w}(u_i)}(z_i)  $
        \STATE Uncertainty of $Z$: $H(Z) \gets \frac{1}{N} \sum_{i=1}^N \log \operatorname{GM}_{\mathbf{\mu_Z}, \mathbf{\Sigma_Z}, \mathbf{w_Z}}(z_i)$
        \STATE Uncertainty of $Z$ given $U$: $H(Z|U) \gets \frac{1}{N} \sum_{i=1}^N \log \operatorname{GM}_{\mathbf{\mu_{Z|U}}(u_i), \mathbf{\Sigma_{Z|U}}(u_i), \mathbf{w_{Z|U}}(u_i)}(z_i)$
        \STATE Return $\mathcal{I}_S(U \rightarrow Z) \gets H(Z) - H(Z|U)$
    \end{algorithmic}
\end{algorithm}

Both the parameters of the marginal distribution of $Z$ and the parameters of the conditional distribution of $Z$ given $U$ are estimated through likelihood maximization. The uncertainty of $Z$ is estimated by the negative log-likelihood of the data under the marginal distribution of $Z$. The uncertainty of $Z$ given $U$ is estimated by the negative log-likelihood of the data under the conditional distribution of $Z$ given $U$.

\subsubsection{Embedding dimension normalization.} 
\begin{figure}
    \centering
    \includegraphics[width=0.9\linewidth]{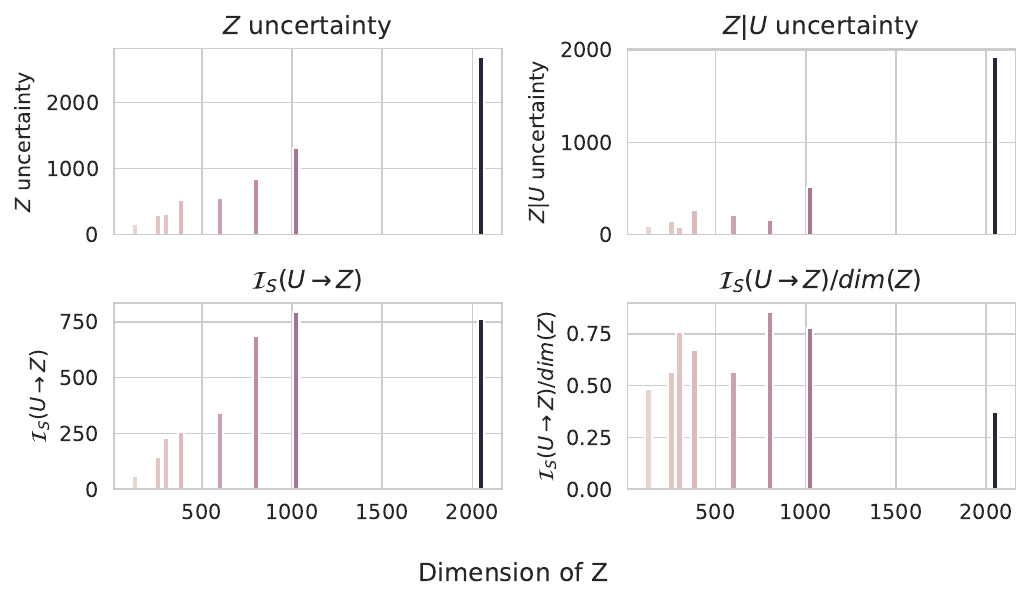}
    \caption{
        Relationship between the dimension of $Z$'s latent space and the quantities estimated to compute the information sufficiency in molecular modeling.
    }
    \label{fig:MI_est_analysis}
\end{figure}

However, this approach faces one major drawback: it favors models that generate embeddings of high dimensionality.
To evaluate the information sufficiency between the models, we estimate the uncertainty of $Z$ and the uncertainty of $Z$ given $U$.
As seen in~\autoref{fig:MI_est_analysis}, the estimated information sufficiency is highly correlated to the dimension of the latent space of $Z$, favoring models with high-dimensional latent spaces.
This can be explained by the fact that these embedders yield larger marginal uncertainties.
The resulting difference in the uncertainties $\mathcal{I}_{S}(U \rightarrow Z)$ is hence larger in absolute values.

\begin{figure}[H]
    \begin{subfigure}[v]{0.44\textwidth}
        \centering
        \includegraphics[width=\linewidth]{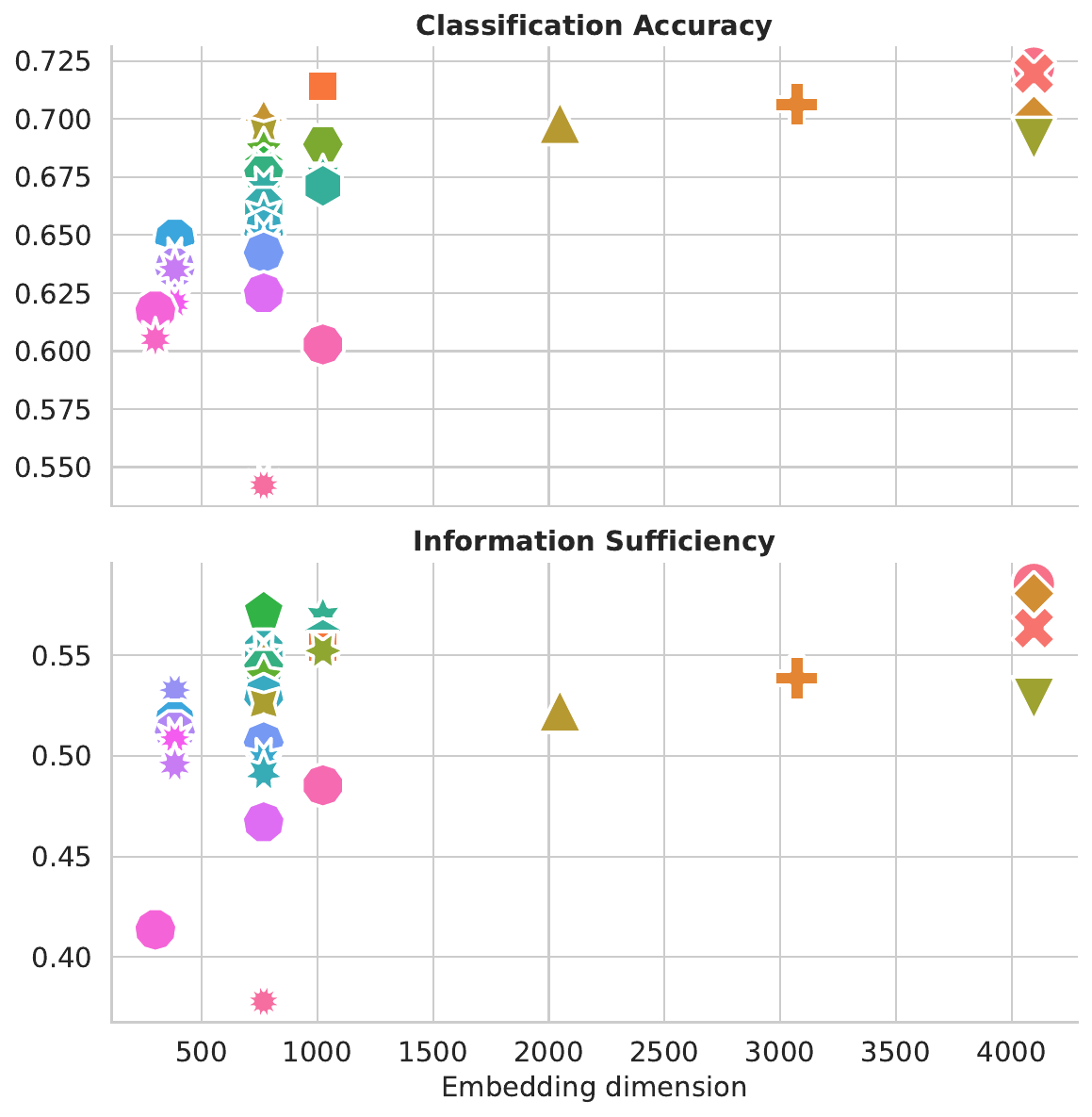}
        \caption{Impact of the embedding size on the quality of the embeddings for the models measured as actual performance on downstream tasks and {\sys} score.}
        \label{fig:size_vs_perf_info}
    \end{subfigure} \hfill
    \begin{subfigure}[v]{0.54\textwidth}
        \centering
        \includegraphics[width=\linewidth]{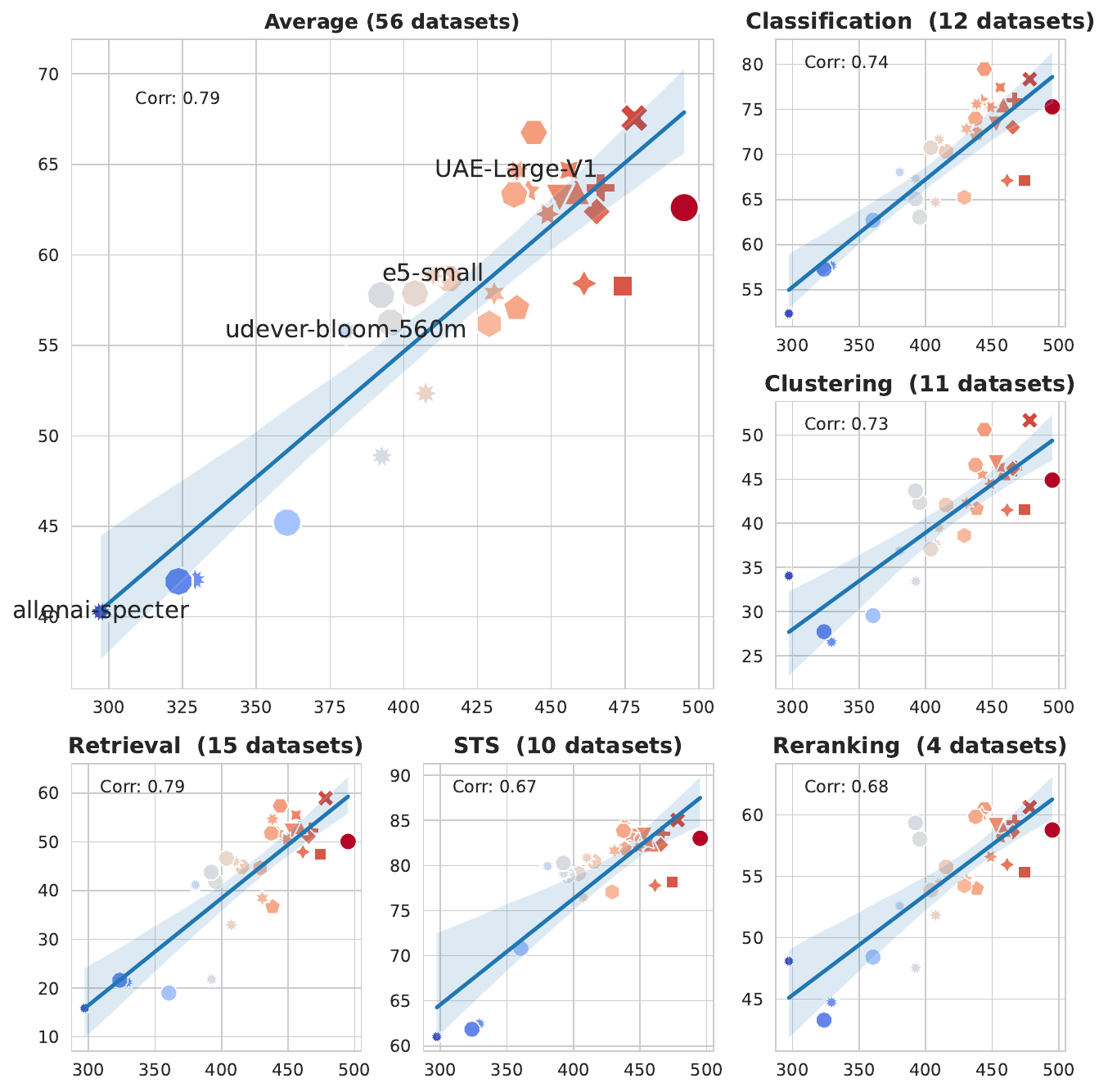}
        \caption{Correlations on the MTEB benchmark when the embeddings are not normalized by their size.}
        \label{fig:mteb_correlation_no_norm}
    \end{subfigure}
    \caption{}
\end{figure}

We can see in~\autoref{fig:MI_est_analysis} that the dimension of the latent space of $Z$ and the uncertainty of $Z$ are evolving linearly.
We thereby divide the information sufficiency by $dim(Z)$, which can be seen as an approximation of the normalization by the uncertainty of $Z$. We report in \autoref{fig:mteb_correlation_no_norm} results without this normalization. It still correlates significantly with the downstream tasks performance, but the correlation is stronger when the normalization is applied.
Hence, we focus on the relative variation of the uncertainty of $Z$ explained by $U$.
Note that the uncertainty of $Z$ can be negative, as it can be assimilated to a differential entropy.
As a result, the ``true`` relative variation of the uncertainty would not be suitable for comparing different models (as it would not guarantee any ordering). While there is a general trend where larger models do have larger embeddings and perform better, well-trained smaller embeddings are competitive with larger embeddings, and the {\sys} score captures this behavior (\autoref{fig:size_vs_perf_info}).

We build our proxy by measuring the median values of the set $\mathcal{S}_{\mathcal{I_S}}\left(k\right) = \left\{\mathcal{I}_{S}(Z_k \rightarrow Z_l) \right\}_{l\not= k}$ for an embedder $Z_k$ in our pool of models.

\label{sec:nlp_embedding_size_impact}

\subsubsection{Median instead of mean.} 
\begin{figure}[htbp]
    \centering
    \begin{subtable}{.5\linewidth}
        \centering
        \resizebox{\linewidth}{!}{\begin{tabular}{lrrr}
\toprule
 & $\varrho_p$ & $\varrho_s$ & $\tau$ \\
\midrule
STS (10 datasets) & 0.90 & 0.85 & 0.66 \\
Retrieval (15 datasets) & 0.89 & 0.85 & 0.62 \\
Classification (12 datasets) & 0.92 & 0.84 & 0.65 \\
Clustering (11 datasets) & 0.86 & 0.83 & 0.61 \\
Reranking (4 datasets) & 0.86 & 0.79 & 0.65 \\
\midrule\midrule
Average (56 datasets) & \bfseries 0.93 & \bfseries 0.86 & \bfseries 0.66 \\
\midrule\midrule
Additional Classif (8 datasets) & 0.84 & 0.81 & 0.61 \\
\bottomrule
\end{tabular}
}
        \caption{NLP}
        \label{tab:common_table_correlation_nlp_mean}
    \end{subtable}\begin{subfigure}{0.5\textwidth}
        \centering
        \includegraphics[width=\linewidth]{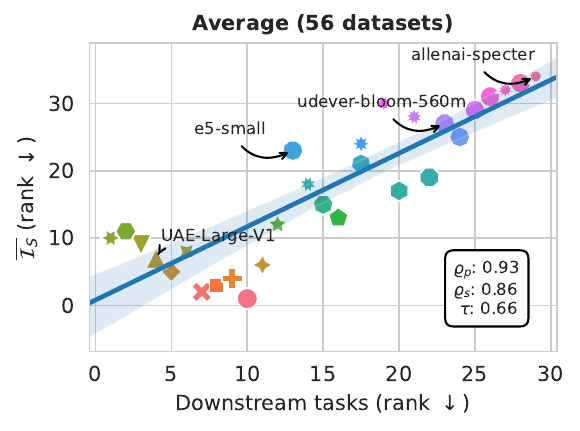}
        \caption{NLP}
        \label{fig:nlp_common_average_mteb_mean}
    \end{subfigure}\hspace{1cm}%

    \caption{Correlation between {\sys} score computed as the mean information sufficiency and the downstream task performances in NLP. $\varrho_p$ is the Pearson correlation, $\varrho_s$ is the spearman correlation}
    \label{fig:mteb_correlation_mean}
\end{figure}

We use the median instead of the mean to compute the {\sys} score. The median is more robust to outliers and the distribution of available embedders. For example, if many models are very similar, the mean would be biased by these models, while the median would not. Thus, we chose the median. While this change has a minor impact when there is enough diversity in the models, it can have a significant impact when the models are very similar, for example, when including different checkpoints of the same model~\autoref{fig:mteb_correlation_mean}.

\subsection{Hyperparameter selection}

\begin{figure}
    \centering
        \centering
        \includegraphics[width=\linewidth]{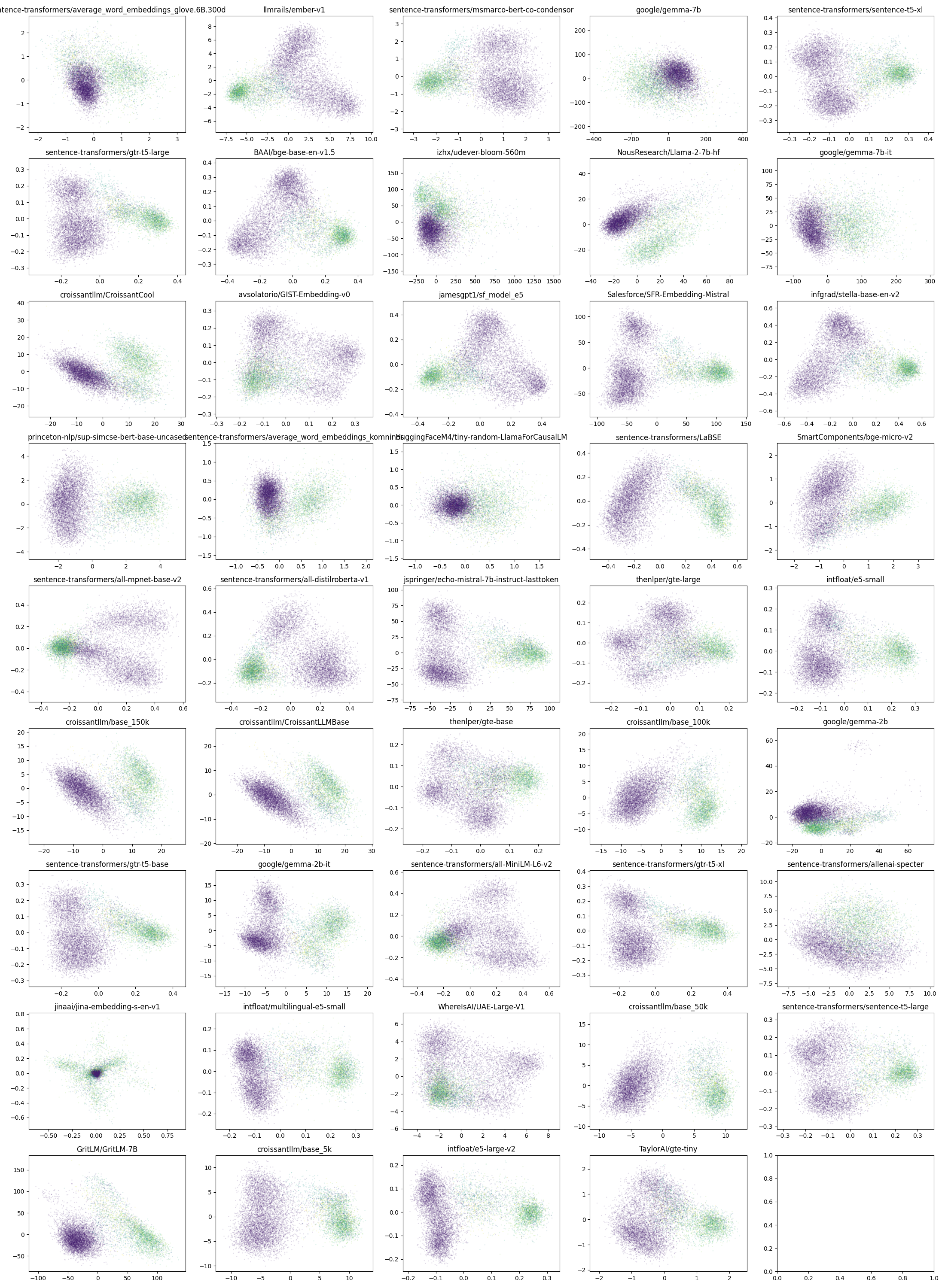}
        \caption{2D Projection of the embeddings of the models in the first two principal components colored by datasets in NLP}
        \label{fig:embeddings_vis}
\end{figure}

\begin{figure}
    \centering
        \centering
        \includegraphics[width=\linewidth]{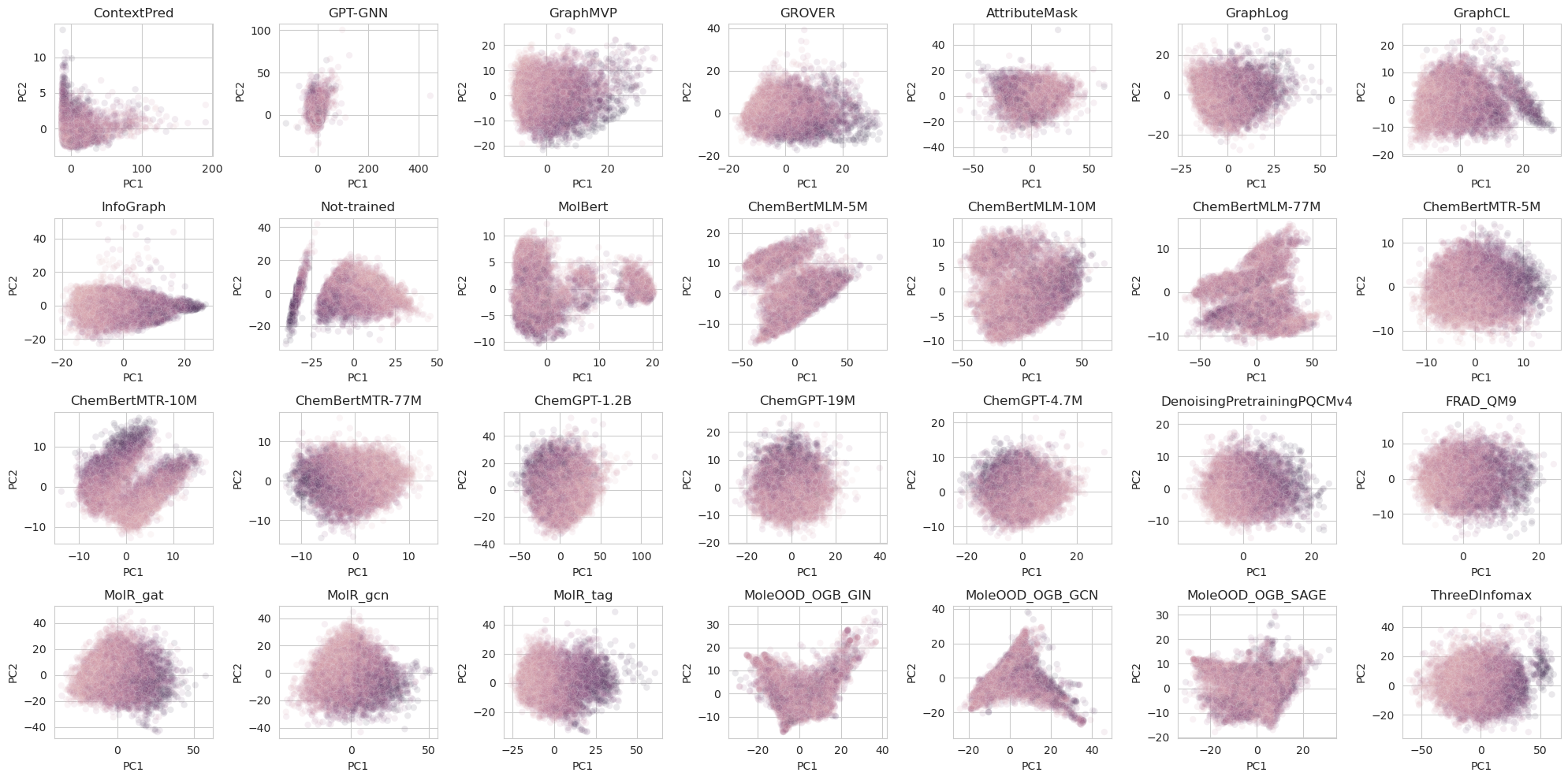}
        \caption{2D Projection of the embeddings of the models in the first two principal components colored by datasets in molecular modeling. Hue corresponds to the synthetic accessibility score~\cite{sas}.}
        \label{fig:embeddings_vis_mol}
\end{figure}

\begin{figure}[H]
    \centering
    \begin{subfigure}[v]{0.33\textwidth}
        \centering
        \includegraphics[width=\linewidth]{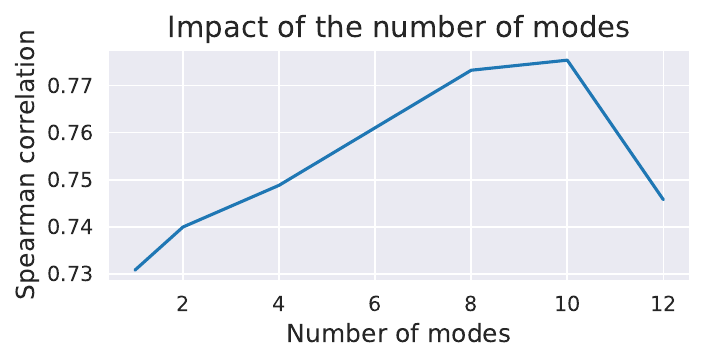}
        \caption{Spearman}
        \label{fig:modes_vs_spearman}
    \end{subfigure} \hfill\begin{subfigure}[v]{0.33\textwidth}
        \centering
        \includegraphics[width=\linewidth]{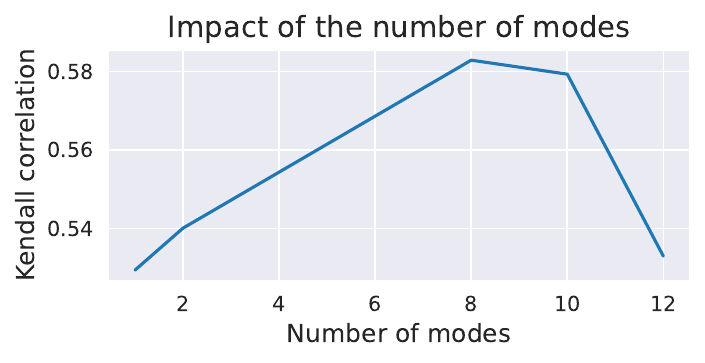}
        \caption{Kendall-Tau.}
        \label{fig:modes_vs_kendall}
    \end{subfigure}\begin{subfigure}[v]{0.33\textwidth}
        \centering
        \includegraphics[width=\linewidth]{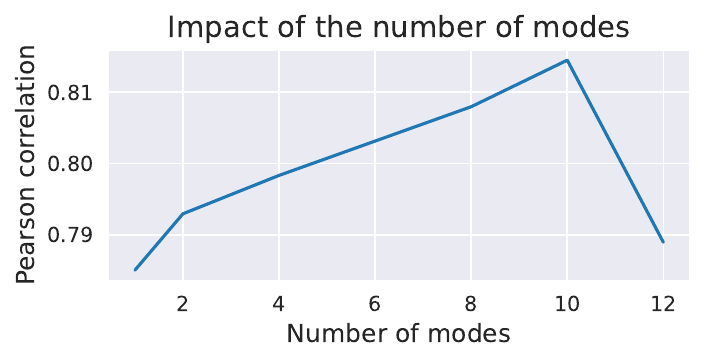}
        \caption{Pearson.}
        \label{fig:modes_vs_Pearson}
    \end{subfigure} \hfill
  \caption{Impact of the number of modes used to estimate the {\sys} score and its correlation with the downstream tasks performance in NLP. We chose to use $8$ modes in practice.}
    \label{fig:modes_vs_correlation}
\end{figure}

We use parametric classes composed of multivariate Gaussian mixture distributions for $\Frond_\Theta$ and $ \mathcal{K}_\Theta(\mathsf{Z} |\mathsf{U} )$ in the definition of the information sufficiency (\autoref{eq:info_suf}), the number of components in the mixture is a crucial parameter that needs to be selected concerning the data distribution of interest. We ran ablation studies to evaluate the impact of the number of components in the mixture on the information sufficiency estimation and the correlation between the {\sys} score and the downstream tasks performance (\autoref{fig:modes_vs_correlation}). We found that the ideal number of components in NLP is $8$, and in molecular modeling, it is $4$. \autoref{fig:embeddings_vis} and \autoref{fig:embeddings_vis_mol} show the embeddings of the models in the first two principal components.

\subsection{Impact of the task size}
\label{subsec:task_size_impact}

\begin{figure}[H]
    \centering
    \includegraphics[width=\linewidth]{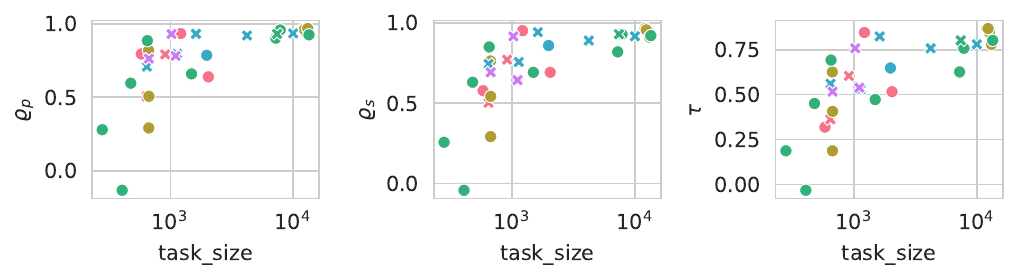}
    \caption{
        Impact of the task size on the {\sys} score's ranking's correlation with the downstream tasks performances in molecular modeling in terms of Pearson correlation $\varrho_p$, Spearman correlation $\varrho_s$ and Kendall-Tau $\tau$ coefficient.
    }
    \label{fig:task_size_impact}
\end{figure}

Our study focused on finding the most promising model to be competitive on any downstream tasks.
However, if the downstream task is not learnable, the most promising model could appear as not competitive on this specific task.
In particular, if the amount of data available in the downstream task is insufficient, the differences between different embedder's representations might not be easily leveraged.
This phenomenon can be seen in~\autoref{fig:task_size_impact}, highlighting how when fewer than 1000 data points are available, the correlation between the {\sys} score and the downstream performance becomes weaker.

\subsection{Impact of the number of models}
\label{subsec:impact_nb_models}

\begin{figure}[H]
    \centering
    \includegraphics[width=\textwidth]{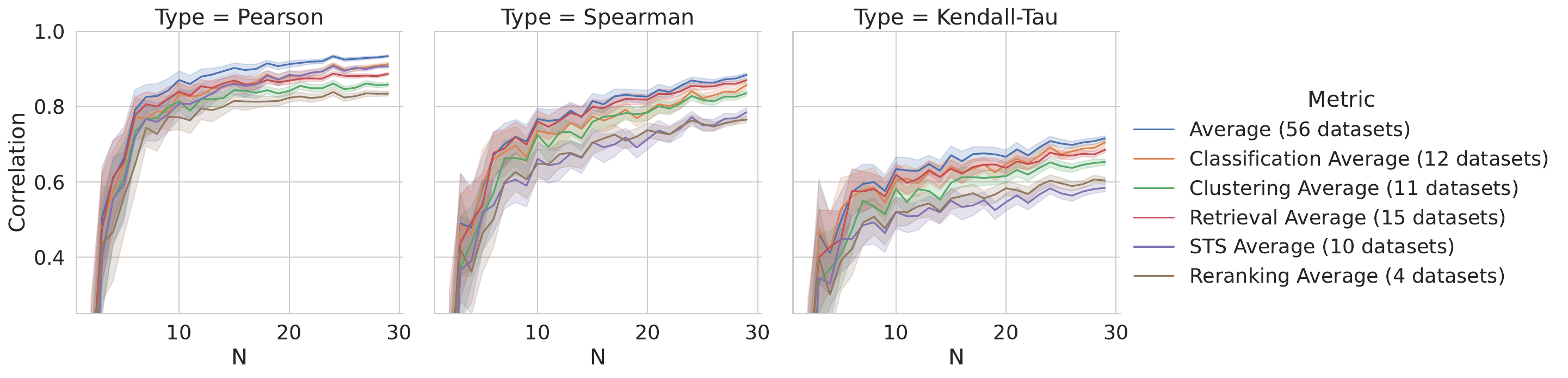}
    \caption{Impact of the number of models used to compute the {\sys} score in NLP.}
    \label{fig:nb_models_vs_informativeness}
\end{figure}

\begin{wrapfigure}{R}{0.5\linewidth}
    \centering
    \includegraphics[width=\linewidth]{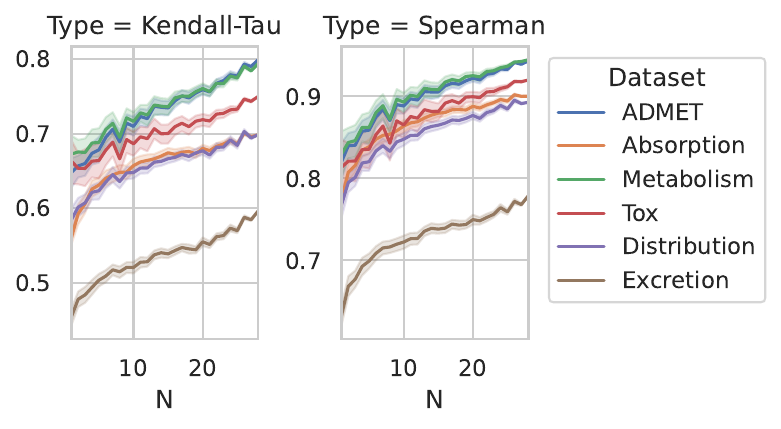}
    \caption{Impact of the number of models used to compute the {\sys} score in molecular modeling.}
    \label{fig:nb_models_vs_informativeness_mol}
\end{wrapfigure}

\label{sec:nlp_nb_models_impact}
We evaluate the strength or {\sys} score of an embedder with respect to all the others by relaxing the "for all" conditions with the median {\sys} score.
Thus, the number of available embedders might impact the performance of our method.
Indeed, if too few embedders are available, it is likely that our evaluation would be biased by favoring models similar to the few available ones.
We evaluate the impact of the number of available models by sampling subsets of our global model pool to compute the {\sys} score.
In \autoref{fig:nb_models_vs_informativeness}, we found that when fewer models are available, the rankings obtained using the {\sys} score correlate less with the downstream tasks' performance.
However, as the number of models increases, the {\sys} score becomes a good proxy for the performance of the models on the downstream tasks.

However, the evolution of this correlation is different in the two studied domains.
Even with very few models, the {\sys} score of molecular models already achieves a Spearman correlation close to $0.8$ with the downstream tasks' performance.
The correlation is much lower in NLP and only reaches $0.8$ when about $15$ models are available.
This result can be expected, seeing how sparse the pairwise comparison matrix is in NLP compared to molecular modeling.
This matrix is less sparse on molecular data, the graph induced by this adjacency matrix is more connected, and having access to a few nodes is enough to obtain measurements on the whole graph.
On the contrary, by being much sparser, having access to a few nodes in this graph in NLP might only give information about the local structure of the graph in the community of the few nodes available.
This might explain why the {\sys} score's correlation with the downstream tasks performance stabilizes in NLP at about ten models, which is equal to the number of communities identified in the graph (in~\autoref{sec:nlp_model_informativeness}).

\subsection{Comparison with other metrics}
\label{sec:naive_baselines}

We evaluate other ways to measure the informativeness (\autoref{tab:naive_baselines_nlp} and \autoref{tab:naive_baselines_mol}) of the embedders. We considered the size of the model, their embeding dimensions and the simple reconstruction error when we train a cross-encoder to go from one embedding to another.  Overall we found that our information sufficiency score significantly outperformed these naive metrics in both modalities.

In particular, training a cross-encoder with the $l_2$ reconstruction error proves to correlate with the ability to enable downstream performance, however, this correlation remains weaker compared to using the information sufficiency.

\begin{table}[h]
    \caption{Comparison with Baselines: Size of the Embedder, Dimension of the embedding output ($d$) and the $\ell_2$ reconstruction error of the embeddings for NLP datasets.}
    \centering
    \resizebox{\textwidth}{!}{
        \begin{tabular}{l|ccc|ccc|ccc|ccc}
\toprule
 & \multicolumn{3}{c}{Size} & \multicolumn{3}{c}{d} & \multicolumn{3}{c}{$\bar{I}_{\mathbf{S}}$} & \multicolumn{3}{c}{$\bar{\ell}_{2}$} \\
 & $\varrho_p$ & $\varrho_s$ & $\tau$ & $\varrho_p$ & $\varrho_s$ & $\tau$ & $\varrho_p$ & $\varrho_s$ & $\tau$ & $\varrho_p$ & $\varrho_s$ & $\tau$ \\
\midrule
Classification (12 datasets) & 0.46 & 0.42 & 0.32 & 0.52 & 0.66 & 0.55 & \bfseries 0.92 & \bfseries 0.88 & \bfseries 0.73 & -0.79 & -0.85 & -0.66 \\
Retrieval (15 datasets) & 0.40 & 0.39 & 0.29 & 0.46 & 0.63 & 0.52 & \bfseries 0.89 & \bfseries 0.89 & \bfseries 0.70 & -0.71 & -0.84 & -0.65 \\
Clustering (11 datasets) & 0.45 & 0.38 & 0.26 & 0.54 & 0.67 & 0.55 & \bfseries 0.86 & \bfseries 0.85 & \bfseries 0.67 & -0.80 & -0.84 & -0.66 \\
STS (10 datasets) & 0.27 & 0.35 & 0.25 & 0.34 & 0.66 & 0.52 & \bfseries 0.92 & \bfseries 0.82 & \bfseries 0.62 & -0.70 & -0.83 & -0.64 \\
Reranking (4 datasets) & 0.33 & 0.33 & 0.26 & 0.41 & 0.61 & 0.50 & \bfseries 0.84 & \bfseries 0.79 & \bfseries 0.64 & -0.71 & -0.78 & -0.59 \\
\midrule\midrule
Average (56 datasets) & 0.41 & 0.41 & 0.31 & 0.48 & 0.62 & 0.50 & \bfseries 0.94 & \bfseries 0.90 & \bfseries 0.74 & -0.77 & -0.84 & -0.65 \\
\midrule\midrule
Additional Classif (8 datasets) & 0.41 & 0.62 & 0.47 & 0.43 & 0.64 & 0.55 & \bfseries 0.89 & \bfseries 0.84 & \bfseries 0.66 & -0.65 & -0.72 & -0.55 \\
\bottomrule
\end{tabular}

    }

    \label{tab:naive_baselines_nlp}
\end{table}

\begin{table}[h]
    \caption{Comparison with Baselines: Size of the Embedder, Dimension of the embedding output ($d$) and the $\ell_2$ reconstruction error of the embeddings for Molecular Modeling datasets.}
    \centering
    \resizebox{1\textwidth}{!}{
        \begin{tabular}{l | rrr | rrr | rrr | rrr}
\toprule
 & & Size & & & d & & & $\bar{I}_{\mathbf{S}}$ & & &  $\bar{\ell}_2$ \\
 & $\varrho_p$ & $\varrho_s$ & $\tau$ & $\varrho_p$ & $\varrho_s$ & $\tau$ & $\varrho_p$ & $\varrho_s$ & $\tau$ & $\varrho_p$ & $\varrho_s$ & $\tau$ \\
\midrule
\textbf{A}bsorption (8 datasets) & - & -0.21 & -0.16 & - & -0.43 & -0.29& - & \textbf{0.89} & \textbf{0.70}& - & \textbf{-0.89} & \textbf{-0.70}\\
\textbf{D}istribution (3 datasets) & - & -0.07 & -0.03 & - & -0.46 & -0.31& - & \textbf{0.89} & \textbf{0.70} & - & -0.86 & -0.66\\
\textbf{M}etabolism (8 datasets) & - & 0.06 & 0.03 & - & -0.46 & -0.34& - & \textbf{0.94} & \textbf{0.79} & - & -0.90 & -0.71\\
\textbf{E}xcretion (3 datasets) & - & -0.17 & -0.11 & - & -0.24 & -0.18& - & \textbf{0.77} & \textbf{0.60} & - & \textbf{-0.77} & -0.56\\
\textbf{T}oxicity (9 datasets) & - & 0.09 & 0.06 & - & -0.49 & -0.35& - & \textbf{0.92} & \textbf{0.75} & - & -0.86 & -0.67\\
\midrule\midrule
\textbf{ADMET} (31 datasets) & - & -0.01 & 0.01 & - & -0.47 & -0.32& - & \textbf{0.94} & \textbf{0.80} & - & -0.90 & -0.72\\
\bottomrule
\end{tabular}

    }
    \label{tab:naive_baselines_mol}
\end{table}

\subsection{Computational ressources}
\label{sec:computational_ressources}

Evaluating the {\sys} score of the models is computationally inexpensive. Evaluating the {\sys} score requires only a single (small) GPU. All our experiments were conducted on NVIDIA V100 and NVIDIA A6000 GPUs.

Our method's main (computational) shortcoming stems from the need to compute the information sufficiency between all pairs of models. This is a quadratic operation in the number of models. However, in practice, optimizing and estimating the information sufficiency presented in \autoref{sec:is_estimation} is cheap. The complete evaluation of the $45$ NLP models can be done in less than $6$ hours on a single GPU.

\newpage

\section*{NeurIPS Paper Checklist}

\begin{enumerate}

    \item {\bf Claims}
    \item[] Question: Do the main claims made in the abstract and introduction accurately reflect the paper's contributions and scope?
    \item[] Answer: \answerYes{} %
    \item[] Justification: We setup the problem of evaluating embedders as a communication problem and leverage information theoretic tools to derive a principled evaluation method.

    \item {\bf Limitations}
    \item[] Question: Does the paper discuss the limitations of the work performed by the authors?
    \item[] Answer: \answerYes{} %
    \item[] Justification: We provide a separate section "Conclusion and Limitations" (\autoref{sec:limitations}) in the main paper and different ablations addressing possible shortcomings of our methods in appendices.

    \item {\bf Theory Assumptions and Proofs}
    \item[] Question: For each theoretical result, does the paper provide the full set of assumptions and a complete (and correct) proof?
    \item[] Answer: \answerYes{} %
    \item[] Justification: We provide a complete description of our setting and assumptions used to derive theoretical proofs as well, as the relaxation of the problem to make it tractable (\autoref{sec:theory}, \autoref{sec:proof}).

    \item {\bf Experimental Result Reproducibility}
    \item[] Question: Does the paper fully disclose all the information needed to reproduce the main experimental results of the paper to the extent that it affects the main claims and/or conclusions of the paper (regardless of whether the code and data are provided or not)?
    \item[] Answer: \answerYes{} %
    \item[] Justification: We provide a description of our global experimental protocol in the main paper (\autoref{sec:common_setting}) and detailed descriptions for the different field in their respective sections (\autoref{sec:nlp_experiment_details}, \autoref{sec:molecular_experiment_details}). All the datasets and models used are publicly available and the code to reproduce our results is attached to the submission.

    \item {\bf Open access to data and code}
    \item[] Question: Does the paper provide open access to the data and code, with sufficient instructions to faithfully reproduce the main experimental results, as described in supplemental material?
    \item[] Answer: \answerYes{} %
    \item[] Justification: We attached the code to the submission and provide instructions to reproduce the results in the main paper and supplemental material. All datasets and models are publicly available and chosen to be easily accessible for reproducibility.

    \item {\bf Experimental Setting/Details}
    \item[] Question: Does the paper specify all the training and test details (e.g., data splits, hyperparameters, how they were chosen, type of optimizer, etc.) necessary to understand the results?
    \item[] Answer: \answerYes{} %
    \item[] Justification: We provide in Appendix \autoref{sec:nlp_experiment_details} and \autoref{sec:molecular_experiment_details} the details of the training and test settings for the different fields. We also provide a global description of the experimental protocol in the main paper (\autoref{sec:common_setting}).

    \item {\bf Experiment Statistical Significance}
    \item[] Question: Does the paper report error bars suitably and correctly defined or other appropriate information about the statistical significance of the experiments?
    \item[] Answer: \answerYes{} %
    \item[] Justification: We provide error bars in the figures and tables of the main paper. In addition we provide robustness checks and ablations in the appendices.

    \item {\bf Experiments Compute Resources}
    \item[] Question: For each experiment, does the paper provide sufficient information on the computer resources (type of compute workers, memory, time of execution) needed to reproduce the experiments?
    \item[] Answer: \answerYes{} %
    \item[] Justification: We provide the details of the compute resources used in the main paper and appendices in \autoref{sec:computational_ressources}.

    \item {\bf Code Of Ethics}
    \item[] Question: Does the research conducted in the paper conform, in every respect, with the NeurIPS Code of Ethics \url{https://neurips.cc/public/EthicsGuidelines}?
    \item[] Answer: \answerYes{} %
    \item[] Justification: This paper does not involve major ethical concerns and the research conducted in the paper conforms with the NeurIPS Code of Ethics.

    \item {\bf Broader Impacts}
    \item[] Question: Does the paper discuss both potential positive societal impacts and negative societal impacts of the work performed?
    \item[] Answer: \answerNA{} %
    \item[] Justification: We only propose model selection method for embedders. The main concern, that is addressed in the paper, is that while our method is theoretically supported and empirically validated, it might not be applicable to all downstream tasks and should be used with caution.

    \item {\bf Safeguards}
    \item[] Question: Does the paper describe safeguards that have been put in place for responsible release of data or models that have a high risk for misuse (e.g., pretrained language models, image generators, or scraped datasets)?
    \item[] Answer: \answerNA{} %
    \item[] Justification: N/A

    \item {\bf Licenses for existing assets}
    \item[] Question: Are the creators or original owners of assets (e.g., code, data, models), used in the paper, properly credited and are the license and terms of use explicitly mentioned and properly respected?
    \item[] Answer: \answerYes{} %
    \item[] Justification: We provide the citations and direct references of all the datasets and models used in our work. To ensure reproducibility, all the datasets and models used are publicly available and under open licenses.

    \item {\bf New Assets}
    \item[] Question: Are new assets introduced in the paper well documented and is the documentation provided alongside the assets?
    \item[] Answer: \answerYes{} %
    \item[] Justification: We only created the library that implements our methods (it will be released publicly on github) and the code to reproduce our experiments. Both are documented as part of the code submission as supplementary material.

    \item {\bf Crowdsourcing and Research with Human Subjects}
    \item[] Question: For crowdsourcing experiments and research with human subjects, does the paper include the full text of instructions given to participants and screenshots, if applicable, as well as details about compensation (if any)?
    \item[] Answer: \answerNA{} %
    \item[] Justification: N/A

    \item {\bf Institutional Review Board (IRB) Approvals or Equivalent for Research with Human Subjects}
    \item[] Question: Does the paper describe potential risks incurred by study participants, whether such risks were disclosed to the subjects, and whether Institutional Review Board (IRB) approvals (or an equivalent approval/review based on the requirements of your country or institution) were obtained?
    \item[] Answer: \answerNA{} %
    \item[] Justification: N/A

\end{enumerate}

\end{document}